\documentclass{article}

\usepackage{arxiv}

\usepackage[utf8]{inputenc} 
\usepackage[T1]{fontenc}    
\usepackage{hyperref}       
\usepackage{url}            
\usepackage{booktabs}       
\usepackage{amsfonts}       
\usepackage{nicefrac}       
\usepackage{microtype}      
\usepackage{lipsum}
\usepackage{graphicx}
\usepackage{amsmath,amsfonts}
\usepackage{algorithmic}
\usepackage{algorithm}
\usepackage{array}
\usepackage{makecell,rotating,multirow,diagbox}
\usepackage{subfloat}
\usepackage{hyperref}
\usepackage{textcomp}
\usepackage{stfloats}
\usepackage{caption}
\usepackage{amsthm}
\usepackage{subfig}
\usepackage{url}
\usepackage{verbatim}
\usepackage{graphicx}
\usepackage{cite}
\usepackage{subfloat}
\graphicspath{ {./images/} }

\newtheorem{theorem}{Theorem}[section]

\newtheorem{lemma}{Lemma}[section]

\newtheorem{assumption}{Assumption}[section]
\newtheorem{remark}{Remark}[section]
\newtheorem{defn}{Definition}[section]

\title{A Hybrid Registration and Fusion Method for Hyperspectral Super-resolution}

\author{
 Kunjing Yang \\
  School of Mathematics\\
  Hunan University\\
  Changsha, Hunan 410082, P. R. China \\
  \texttt{kunjing-yang@hnu.edu.cn} \\
   \And
 Minru Bai \\
  School of Mathematics\\
  Hunan University\\
  Changsha, Hunan 410082, P. R. China \\
  \texttt{minru-bai@hnu.edu.cn} \\
  \And
 Ting Lu \\
  College of Electrical and Information Engineering\\
  Hunan University\\
  Changsha, Hunan 410082, P. R. China \\
  \texttt{tingluhnu@gmail.com} \\
}

\begin{document}
\maketitle
\date{}
\begin{abstract}
Fusing hyperspectral images (HSIs) with multispectral images (MSIs) has become a mainstream approach to enhance the spatial resolution of HSIs. Many HSI-MSI fusion methods have achieved impressive results. Nevertheless,  certain challenges persist, including: 
		(a) A majority of current methods rely on accurate registration of HSI and MSI, which can be challenging in real-world applications.
		(b) The  obtained HSI-MSI pairs may not be fully utilized.   
		In this paper, we propose a hybrid registration and fusion constrained optimization model named RAF-NLRGS. With respect to challenge (a), the  RAF model integrates batch image alignment within the fusion process, facilitating simultaneous execution of image registration and fusion. To address issue (b), the NLRGS model incorporates a nonconvex low-rank and group-sparse structure, leveraging group sparsity to effectively harness valuable information embedded in the residual data. Moreover, the NLRGS model can further enhance fusion performance based on the RAF model.
		Subsequently, the RAF-NLRGS model is solved within the framework of Generalized Gauss-Newton (GGN) algorithm and Proximal Alternating Optimization (PAO) algorithm.  Theoretically, we establish the error bounds for the NLRGS model and the convergence analysis of corresponding algorithms is also presented. Finally, extensive numerical experiments on HSI datasets are conducted to verify the effectiveness of our  method. 
\end{abstract}


\section{Introduction}
Regular color images are limited to carrying information through  three channels – red, green, and blue, whereas HSIs can encompass dozens or hundreds of continuous spectral bands. This enables HSIs to capture a wider range of the electromagnetic spectrum and provide more detailed spectral information. Considering various objects typically exhibit distinct reflectance characteristics, HSIs enable accurate  recognition of materials, which is beneficial for many tasks, including   anomaly detection \cite{Li2023} and disease monitoring \cite{Lv2021}.
	
	However, the spatial resolution of HSIs frequently suffers from various factors, including sensor noise and constraints inherent to transmission systems \cite{Milad}. Additionally, there exist fundamental physical limitations that lead to an inevitable trade-off between the spatial and spectral resolutions in HSIs, which limits their further practical applications \cite{highcost}. In contrast, MSIs usually boast superior spatial resolution but compromises on spectral resolution. A feasible approach is to fuse the high spatial resolution MSI with a high spectral resolution HSI. The fusion operator effectively harmonizes the balance between spatial and spectral resolutions, thereby enabling the extraction of more intricate information from the scene.
	
	In real-world applications, image registration serves as a crucial preprocessing step of fusion. Image registration involves meticulously aligning images through the application of spatial transformations to eradicate geometric discrepancies and ensure correspondence within a common coordinate system. A multitude of fusion methodologies  overlook this crucial step, proceeding under the assumption that the images are perfectly aligned. Nevertheless, this assumption is difficult to realize since the registration of HSIs and MSIs presents a formidable challenge, largely due to their pronounced disparities in spatial resolution and spectral resolution \cite{Qu2022}. To address the complexities of image registration task, certain researchers endeavor to execute  registration and fusion processes concurrently. For example, Chen \textit{et al.} \cite{SIRF} proposed leveraging a dynamic gradient sparsity property in the process of registering two images within the pansharpening task. Nevertheless, this approach does not fully exploit the intrinsic properties of the image. Unni \textit{et al.} \cite{Unni2020} developed a Plug and Play framework for HSI-MSI fusion tasks, which leverages neural networks to learn image priors and address the  registration challenges. However, solving the subproblems requires pre-trained denoisers and registration networks, which entails a relatively high cost. The work in \cite{Fu2020} presented an effective approach for simultaneous HSI super-resolution and geometric alignment of the image pair, incorporating spectral dictionary learning and sparse coding methods. 
	Nonetheless, relying solely on sparse representation might not adequately capture the intrinsic features of the data.
	
	To tackle these issues, we  develop a constraint optimization model named RAF that simultaneously performs registration and fusion. This framework accounts for disparities across various bands within the HSI, incorporating batch image alignment into the fusion task, which applies distinct transformations to each band.   Firstly, the RAF model capitalizes on the inherent spectral correlations within the HSI to project the target HSI onto a reduced-dimensional subspace, effectively diminishing the model's dimensionality. Secondly, we employ a low-rank approximation approach to further explore the spatial correlations within the data. Thirdly, in recognition of the differences inherent among multi-source images due to variables like illumination changes and varying weather conditions, RAF model incorporates a linear transformation in the spectral domain of the MSI to reconcile and mitigate those discrepancies.  Considering the inherent nonlinearity of the registration problem, we employ the Generalized Gauss–Newton (GGN) algorithm, which tackles it iteratively via a process of local linear approximations. Finally, convergence analysis of corresponding algorithms is also provided.
	
	Considering the RAF model undertakes both registration and fusion tasks simultaneously, it may not consistently yield the most optimal fusion outcomes. Therefore, building upon the foundation of RAF model, we propose a novel NLRGS framework, which is specifically designed to optimize and bolster the performance of the fusion process. The RAF model furnishes NLRGS framework with registered HSI-MSI pairs and an advantageous initial point.  This strategic integration has led to our two-stage method: RAF-NLRGS. For the fusion task, most  methods concentrated on the reconstruction of primary information content, while  neglecting the potential extraction of details from the residuals. Xu \textit{et al}. \cite{IR} proposed an iterative regularization method that incorporates the residual into the original model for repeated calculation. While this methodology results in considerable enhancement, it is important to acknowledge that the residual information might not necessarily conform to the priors inherent in the original model, which may lead to a marked deceleration in computational efficiency during subsequent processing stages.  Drawing inspiration from these insights, our NLRGS model, while integrating nonconvex low-rank techniques  to reconstruct the primary part of the image, also employs a group sparsity prior to meticulously extract significant fine details from the residual data, which realizes a further enhancement in image quality.
	Then, Proximal Alternating Optimization (PAO) algorithm is applied to solve the NLRGS model. Theoretically, we establish error bounds for the NLRGS model and conduct convergence analysis of the PAO algorithm. 
	
	The main contributions  are summarized as follows:
	\begin{itemize}
		\item  We integrate batch image alignment into the HSI-MSI fusion problem and devise a constrained optimization model termed RAF. The RAF model is designed to concurrently execute both image registration and fusion tasks, while also adeptly managing spectral discrepancies across images originating from diverse sources.
		
		\item To further enhance fusion performance, we propose the NLRGS model, which features a nonconvex low-rank plus group sparse structure, enabling efficient recovery of both principal and detailed information. Simultaneously, we establish error bounds for the NLRGS model.
		
		\item The GGN  and PAO algorithms are utilized to tackle the RAF and NLRGS models, respectively. Moreover, we provide a detailed convergence analysis of corresponding algorithms. Finally, extensive numerical experiments conducted on both simulated and real-world HSI datasets substantiate the efficacy of our proposed method.
	\end{itemize}
The remaining sections of this paper are organized as follows. Related work on registration and fusion  are  introduced in section \ref{A}.  In section \ref{B}, essential notation and definitions are introduced. This section also provides  basic description of the registration and fusion problem along with corresponding assumptions.
	In Section \ref{D}, we  thoroughly introduce the RAF and NLRGS models. The corresponding theories and algorithms are also provided in this section. Then in section \ref{NN}, extensive experiments on HSI-MSI registration and fusion are presented. Finally, we  make a conclusion in section \ref{G}.
	
	\section{Related work} \label{A}
	\subsection{Image registration}
	Image registration serves as a pivotal preprocessing step in image fusion, which can fundamentally be divided into two primary categories: feature-based and intensity-based approaches. In the realm of feature-based algorithms,  Scale-Invariant Feature Transform (SIFT) \cite{SIFT} is a  distinguished technique and has made significant strides in computer vision applications. To boost  the computational efficiency of SIFT, Bay \textit{et al.} \cite{Bay2006} incorporated the Hessian matrix into the SIFT framework and developed the Speeded-Up Robust Features (SURF) algorithm. The research in \cite{FAST} involves extracting feature points using the Accelerated Segment Test (FAST) algorithm,  significantly diminishing computational time requirements.
	Radiation variation Insensitive Feature Transform (RIFT) \cite{RIFT} capitalized on the phase congruency principle, known for its robustness against changes in lighting and radiometric variations, to formulate discriminative descriptors specifically tailored for multi-sensor image registration. 
	
	Intensity-based registration algorithms embrace a holistic pixel-wise approach, wherein they introduce quantitative measures that effectively embody the image's consistency for purposes of image alignment, e.g., Mutual Information (MI) \cite{MI}. Myronenko \textit{et al.} \cite{R2010} proposed an alignment method that hinges on minimizing the Residual Complexity (RC) between two images, an indicator that quantifies the sparsity of the residual image ensuing from the alignment process. Chen \textit{et al.} \cite{NTG} introduced a novel measure termed Normalized Total Gradient, assuming that the gradient of the difference between a pair of well-aligned  images exhibits a higher degree of sparsity.  Li \textit{et al.} \cite{Li2018} proposed a novel similarity measure based on the registered images can be sparsified hierarchically in the gradient domain and frequency domain with the separation of sparse errors.
	Cao \textit{et al.} presented \cite{Cao2020} a Structure Consistency Boosting (SCB) transformation, which is designed to enhance the structural resemblance between MSIs. 
	
	Beyond pairwise image registration, batch image alignment refers to the procedure of aligning a group of interconnected images. Peng \textit{et al.} \cite{RASL} utilized the sparse and low-rank decomposition and identify an optimal set of image domain transformations to align a batch of linearly correlated images. However, they vectorize each grayscale image, which destroys the inherent correlation. Zhang \textit{et al.} \cite{Zhang2021} proposed a convex image alignment method from the perspective of tensors, and provided an analysis of the error bounds. Xia \textit{et al.} \cite{Xia2023} incorporated transformed tensor-tensor products and nonconvex $l_p$ norm regularization for image alignment.

	\subsection{HSI-MSI fusion}
	Existing approaches for HSI-MSI fusion mainly encompass three  categories: deep learning-based, matrix factorization (MF)-based and tensor factorization (TF)-based methods.  Deep learning-based methods leverage neural networks to glean the inherent features  between HSI and MSI data, then the networks synthesize a fused image that effectively integrates the spatial detail and spectral attributes from both datasets \cite{Xie2022a}.	Dian \textit{et al.} \cite{ZSL} enhanced the model's adaptability by downsampling the images destined for fusion to create training data, which is subsequently employed to train the neural network.
	MF-based methods explore the internal correlations within images to reduce problem dimensions or perform component separation.
	Yokoya \textit{et al.} \cite{CNMF} proposed a Coupled Nonnegative Matrix Factorization (CNMF) fusion technique rooted in the principles of a linear spectral mixture model, which decomposes the spectral information to effectively integrate and fuse data.  Dong \textit{et al.} \cite{NSSR} developed a Nonnegative Structured Sparse Representation (NSSR) method which jointly estimated the  dictionary and the sparse coefficients.
	However, MF-based methods require the initial unfolding of three-dimensional data into matrices, a process that may inadvertently undermine the intrinsic structural integrity of the data and potentially lead to a reduction in the efficacy of image restoration or reconstruction.
	
	Recently, TF-based methods have demonstrated a superior capacity for leveraging the  multilinear features of the data in comparison to MF-based methods \cite{Zhao1}. In \cite{CP1}, the target HSI was restructured as a CP approximation and an alternating least squares algorithm called Super-resolution TEnsor REconstruction (STEREO) was developed. Zeng \textit{et al.} \cite{CP2024} utilized  CP factorization to characterize the low-rank structure of HSIs with a sparse Bayesian framework for adaptive rank determination. Li \textit{et al.} \cite{CSTF} proposed a methodology employing Coupled Sparse Tucker Factorization (CSTF) for HSI-MSI fusion. The work in \cite{LTTR} employed nonlocal clustering and low Tensor-Train rank representation (LTTR) as regularization to enhance the model's performance and robustness.   Chen \textit{et al.} \cite{TR1} developed a fusion method based on Tensor Ring (TR) decomposition and a factor smoothed regularization was used to encapsulate the inherent spatial-spectral consistency. Moreover, in \cite{XY1}, the authors utilized  tensor-product (t-product) framework to decompose the target HSI, aiming to  exploit its underlying interdependent relationships.
	
	\subsection{Registration-fusion method}
	In addition to the concurrent registration and fusion approaches mentioned in introduction, there are also research focusing on a two-stage methodology. This strategy revolves around a sequential framework that begins with the execution of image registration, followed by the subsequent undertaking of the fusion process. Zhou \textit{et al.} \cite{Zhou2020} proposed a  integrated registration and fusion method that minimized a least-squares objective function by utilizing the point spread function (PSF). Ying \textit{et al.} \cite{NED} introduce a new registration method NED for clear–blur image similarity measurement and incorporate the interpolation process into the fusion process. Guo \textit{et al.} \cite{Guo2022} utilized prior geographical information for pre-registration using Rational Polynomial Coefficient (RPC) orthorectification, then a CNN-based network is proposed to optimize the image registration and blur kernel learning jointly,  culminating in the execution of the fusion process.
	
	\section{Problem formulation } \label{B}
	In this section, we commence by presenting a comprehensive overview of the essential preliminary knowledge  related to our work.
	Then, a brief introduction to HSI-MSI registration and fusion tasks are provided.  Finally, we make some fundamental assumptions regarding the registration and fusion problem to facilitate subsequent theoretical analyses.
	\subsection{Preliminary}
	In this paper, we consistently employ the following notation conventions:  scalars are represented by lowercase letters, such as '$a$'; vectors are denoted by bold lowercase letters such as '$\mathbf{a}$'; matrices are represented by bold uppercase letters such as '$\textbf{A}$'; tensors are denoted by calligraphic letters such as '$\mathcal{A}$'; sets and fields  are denoted by hollow letters such as '$\mathbb{R}$'. Both $\Vert \cdot \Vert_F$ and $\Vert \cdot \Vert$  represent the Frobenius norm. Next, we present a concise  introduction of tensor algebra.  Readers can refer to \cite{Kolda} for a more comprehensive exposition on tensor.
	
	\begin{defn}\label{B2}(F-diagonal Tensor \cite{TSVD}) A tensor $\mathcal{X}$ is called f-diagonal if each frontal slice $\textbf{X}^{(i)}$ is a diagonal matrix.
	\end{defn}
	
	\begin{defn}\label{B3}(Tensor Singular Value Decomposition: t-SVD \cite{TSVD}) For $\mathcal{X} \in \mathbb{R}^{n_{1}\times n_{2} \times n_{3}}$ , the t-SVD of $\mathcal{X}$ is given by $\mathcal{X}=\mathcal{U}\ast \mathcal{S}\ast\mathcal{V}^{\mathcal{H}}$,
		where $\mathcal{U}\in \mathbb{R}^{n_{1}\times n_{1} \times n_{3}}$ and $\mathcal{V}\in \mathbb{R}^{n_{2}\times n_{2} \times n_{3}}$ are orthogonal tensors, $\mathcal{V}^{\mathcal{H}}$ denotes the conjugate transpose of $\mathcal{V}$ and $\mathcal{S}\in \mathbb{R}^{n_{1}\times n_{2} \times n_{3}}$ is a f-diagonal tensor. 
		The entries in $\mathcal{S}$ are called the singular fibers of $\mathcal{X}$ and $*$ denotes the t-product.
	\end{defn}
	
	\begin{defn}(Tensor Average Rank \cite{TSVD}) For  $\mathcal{X} \in \mathbb{R}^{n_{1}\times n_{2} \times n_{3}}$, the tensor average rank, denoted as $\text{rank}_a(\mathcal{X})$, is defined as $\text{rank}_a(\mathcal{X})=\frac{1}{n_3}\sum_{i=1}^{n_3}\text{rank}(\hat{\textbf{X}}^{(i)})$, where $\hat{\textbf{X}}^{(i)}$ is the $i$ th frontal slice  of $\hat{\mathcal{X}}$,  and $\hat{\mathcal{X}}:=\text{fft}(\mathcal{X},\operatorname{[\enspace]},3)$. Here $\text{fft}$ represents the Fast Fourier Transform.
	\end{defn}

	\begin{defn}\label{B4}(Tensor Nuclear Norm \cite{TSVD}) The TNN of $\mathcal{X}\in \mathbb{R}^{n_{1}\times n_{2} \times n_{3}}$ is the average of the nuclear norm of all the frontal slices of $\hat{\mathcal{X}}$, i.e., $\Vert \mathcal{X} \Vert_{\text{TNN}}=\frac{1}{n_3}\sum_{i=1}^{n_3}\Vert \hat{\textbf{X}}^{(i)} \Vert_*  $, where  $\Vert \cdot \Vert_*$ denotes the nuclear norm of a matrix.
	\end{defn}
	
	In the following, a class of functions will be presented, accompanied by their respective proximal mappings.  We initially make the following assumptions for the function $\psi$.
	\begin{assumption} \label{A1}
		(a). $\psi: \mathbb{R}\to \mathbb{R}$ is proper, lower semicontinuous and symmetric with respect to y-axis; (b). $\psi$ is concave and monotonically nondecreasing on $[0,+\infty)$ with $\psi(0)=0$.
	\end{assumption}

In fact, many commonly used functions in statistics satisfy  Assumption \ref{A1}. We list three of them as follows:
	\begin{itemize}
		\item $l_1 $\cite{L1}: $\psi^{l_1}(x)=|x|$;
		\item Logarithm \cite{log}: $\psi^{log}(x)=log(|x|+\theta), \theta>0$;
		\item MCP \cite{Zhang}:$\psi^{\text{MCP}}(x)=\left\{
		\begin{array}{rcl}
			|x|-\dfrac{x^2}{2\theta},       &      & |x|\le \theta,\\
			\dfrac{\theta}{2},     &      & |x|\textgreater \theta.
		\end{array} \right.$
	\end{itemize}
	
	We now present the definition of the proximal mapping. 
	\begin{defn}\label{B8} (Proximal Mapping) 
		The proximal operator of $z\in \mathbb{R}$ with respect to function $\psi$ is defined as 
		\begin{equation*}
			\text{prox}_{\alpha\psi}(z):=\text{arg}\min\limits_{x} \enspace \{ \alpha\psi(x)+\dfrac{1}{2}(x-z)^2\}.
		\end{equation*}
	\end{defn}
	The proximal mappings of above functions have closed-form solution, which can be found in the Appendix A.

\subsection{Problem formulation}
	HSIs and MSIs can be considered as three-dimensional tensors. We represent the target high-resolution HSI (HR-HSI) as $\mathcal{X}\in \mathbb{R}^{M\times N\times H}$, where $M$ and $N$ denote the two spatial dimensions, and $H$ represents the number of spectral bands. Similarly, the obtained HSIs and MSIs are denoted as $\mathcal{Y}\in \mathbb{R}^{m\times n\times H}$ and $\mathcal{Z}\in \mathbb{R}^{M\times N\times h}$ respectively, where $m\textless M$, $n\textless N$, and $h\textless H$.
	
	The derived HSI $\mathcal{Y}$ can be perceived as a spatially degraded variant of the target HR-HSI $\mathcal{X}$.
	Therefore, the relationship between $\mathcal{X}$ and $\mathcal{Y}$ can be expressed as 
	\begin{equation}\label{C1}
		\begin{array}{l}
			\mathcal{Y}_{(3)}=\mathcal{X}_{(3)}\textbf{BS}+\textbf{N}_1,
		\end{array}
	\end{equation} 
	where the subscript $(\cdot)_{(3)}$ denotes the mode-3 unfolding matrix of a tensor.  
	$\textbf{B}\in \mathbb{R}^{MN\times MN}$ is a spatial blurring matrix representing the hyperspectral sensor’s point spread function(PSF),  which is assumed band-independent and operates under circular boundary conditions \cite{Hysure}. $\textbf{S}\in \mathbb{R}^{MN\times mn}$ is a downsampling matrix, whose column vectors form a subset of the identity matrix. Matrix $\textbf{N}_1$ represents the noise, with the assumption that it follows the Gaussian distribution. Similarly, the acquired MSI $\mathcal{Z}$ can be considered as the  degraded version of HR-HSI $\mathcal{X}$ along the spectral mode. Therefore, the relationship between $\mathcal{X}$ and $\mathcal{Z}$ can be expressed as follows:
	\begin{equation}\label{C2}
		\begin{array}{l}
			\mathcal{Z}_{(3)}=\textbf{R}\mathcal{X}_{(3)}+\textbf{N}_2,
		\end{array}
	\end{equation} 
	where $\textbf{R}\in \mathbb{R}^{h\times H}$ is the spectral downsampling matrix, and $\textbf{N}_2$ is the Gaussian noise contained in the observed MSI.  For simplicity, we rewrite (\ref{C1}) and (\ref{C2}) as: 
	\begin{equation}\label{C3}
		\begin{array}{l}
			\mathcal{Y}=\mathcal{H}_{spa}(\mathcal{X})+\mathcal{N}_1,\quad
			\mathcal{Z}=\mathcal{H}_{spec}(\mathcal{X})+\mathcal{N}_2,
		\end{array}
	\end{equation} 
	where $\mathcal{H}_{spa}$ and $\mathcal{H}_{spec}$ represent spatial  and spectral degradation operators. Their specific expressions are as follows:
	\begin{equation*}
		\begin{aligned}
			&\mathcal{H}_{spa}(\mathcal{X}):=fold_3(unfold_3(\mathcal{X})\cdot\textbf{BS}),\\
			&\mathcal{H}_{spec}(\mathcal{X}):=fold_3(\textbf{R}\cdot unfold_3(\mathcal{X})),
		\end{aligned}
	\end{equation*} 
	where $unfold_3(\cdot)$ represents the mode-3 unfolding matrix of a tensor and $fold_3(\cdot)$ denotes its inverse operator. 

Considering that  there may be infinite solutions satisfying equation (\ref{C3}), it is necessary to investigate the prior characteristics of the images  to narrow down the scope of the solution space and find solutions that better align with our expectations. 
	
	It is noteworthy that in practical scenarios, the HSI-MSI pairs acquired   from different sensors may exhibit poor registration. Consequently, it is imperative to conduct image registration preprocessing prior to fusion. Considering the inherent discrepancies among different spectral bands within HSIs, we treat HSIs as a collection of individual grayscale images and apply spatial transformations to each spectral band of HSI to achieve piecewise registration with respect to the MSI. Specifically, we need to find a transformation $\tau$ to ensure $\mathcal{Y}\circ\tau$ is aligned with the MSI. Here $\mathcal{Y}\circ\tau$ is defined as 
	\begin{equation*}
		\begin{array}{l}
			(\mathcal{Y}\circ\tau)_i=\mathcal{Y}_i\circ\tau_i,
		\end{array}
	\end{equation*} 	
	where $\tau$ denotes the mapping that consists of $H$ transformations $\{\tau_i, i=1,\cdots,H\}$ and $\mathcal{Y}\circ\tau$  represents image $\mathcal{Y}$ warped by $\tau$. The $(\cdot)_i$ represents corresponding $i$th band. 

	\subsection{Some fundamental assumptions}
	With respect to the issues of image fusion and registration tasks, we make some fundamental assumptions to facilitate further theoretical analysis. For degradation operators $\mathcal{H}_{spa}$ and $\mathcal{H}_{spec}$, we first introduce the restricted  eigenvalue (RE) condition as follows.

\begin{defn}\label{REC} The operator $\mathcal{H}$ is said to satisfy the restricted eigenvalue condition relative to constant $s$ (REC($s$)) if 
		$$
		\xi(s):=\min_{\mathcal{X}\in \mathbb{X}_{s}}\dfrac{\Vert \mathcal{H}(\mathcal{X}) \Vert_F}{\Vert \mathcal{X} \Vert_F}>0,
		$$
		where $\mathbb{X}_{s}$ represents some set related to  $s>0$.
	\end{defn}
	
	This definition directly extends  the notion of the restricted tensor eigenvalue condition (RTEC) \cite{Yu2024}. A similar restricted eigenvalue condition  for matrices was introduced in \cite{B2009}. 

The theoretical analysis in this paper is primarily built upon the RE  condition. Therefore, for the sake of clarity, we initially elucidate the role of  RE condition within the context of HSI-MSI  fusion:
	\begin{itemize}
		\item Intuitively,  RE condition is designed to ensure that a non-zero HSI signal $\mathcal{X}$ does not become fully nullified following the degradation process.
		
		\item 
		Certain pixel regions in image $\mathcal{X}$ may be infinitely large, but due to their discarding during downsampling, $\mathcal{X}$ still complies with equations (\ref{C1}) and (\ref{C2}). The RE condition effectively precludes this scenario, as demonstrated through a proof by contradiction:  Assume, for the sake of argument, that individual pixels within the set $\mathcal{X}$ were to approach infinity, then $$\Vert\mathcal{Z} \Vert=\Vert\mathcal{H}_{spec}(\mathcal{X}) \Vert>\xi\Vert \mathcal{X} \Vert\to +\infty,$$
		which contradicts $\Vert \mathcal{Z} \Vert$ being a constant. 
	\end{itemize}
	From a counter perspective, it is impractical to use an entirely zero-valued image for fusion purposes, and moreover, the distribution of pixel values in images is typically bounded. Therefore, the RE condition is reasonable, which  circumvents these extreme cases that could otherwise disrupt our  further theoretical analysiss. 
	
	Regarding the registration problem, it is necessary to stipulate that function $\Gamma(\tau):=\mathcal{Y}\circ\tau$ is twice continuously differentiable in our method. In fact, the manifolds formed by domain transformed images may not be continuously differentiable due to the appearance of sharp edges \cite{Do2005}. In our case, however, the digital images $\mathcal{Y}_i\circ\tau_i$ can be viewed as resampling transformations of an ideal bandlimited reconstruction obtained from the digital image $\mathcal{Y}_i$, where the mapping is indeed twice continuously differentiable \cite{RASL}. 

 \section{The proposed RAF-NLRGS model}\label{D}
\begin{figure}
		\centering
        \includegraphics[width=0.55\textwidth,height=0.4\textwidth]{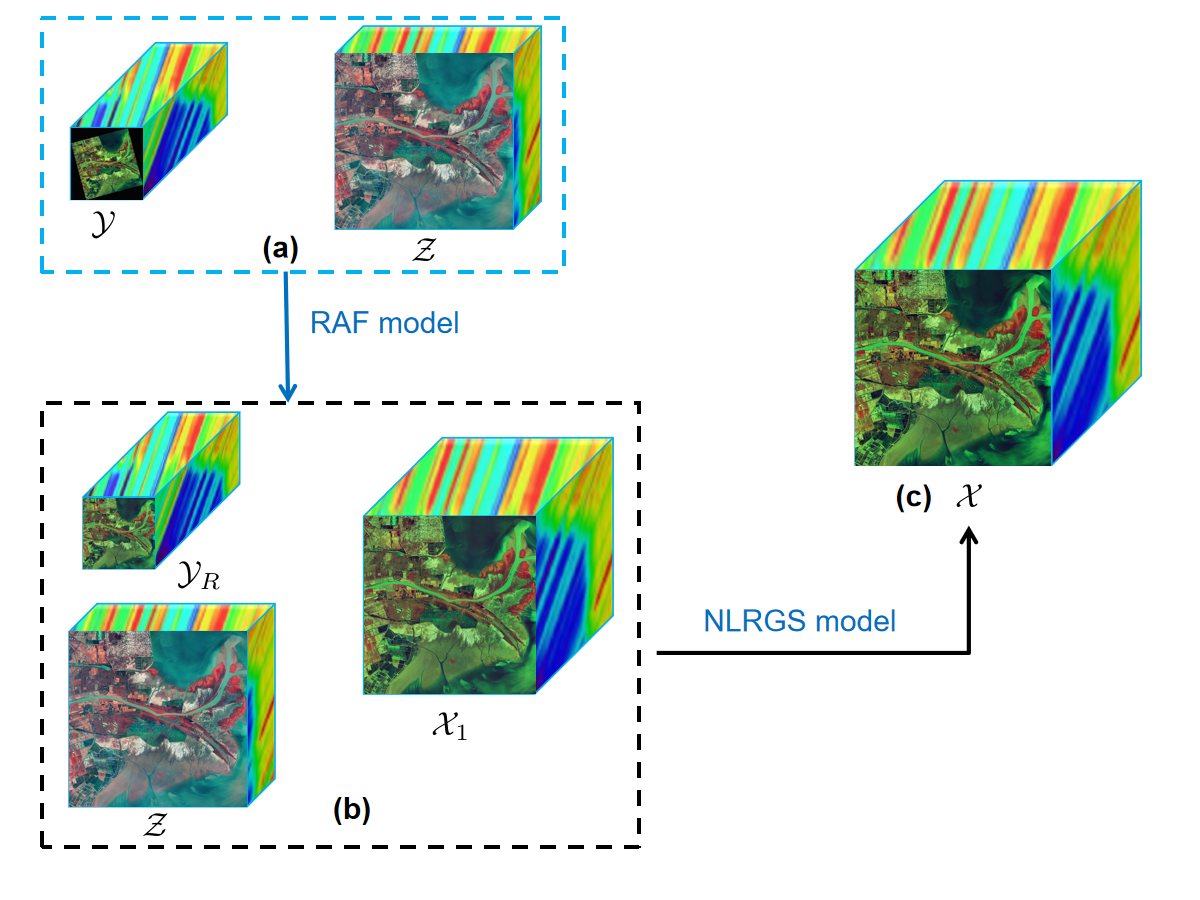}
		\vspace{-5pt}
		\caption{\textbf{(a)} Unregistered image pairs: $\mathcal{Y}$ and $\mathcal{Z}$. \textbf{(b)} Registered image pairs: $\mathcal{Y}_{R}$, $\mathcal{Z}$ and preliminary fused image $\mathcal{X}_1$.  \textbf{(c)} Final fused image $\mathcal{X}$.}
		\vspace{-6pt}
		\label{fig34}
\end{figure}

	In this section, we present the RAF-NLRGS methodology for HSI-MSI registration and fusion. The RAF model supplies the NLRGS model with registered image pairs and excellent initial points. The NLRGS model, on the other hand, manages to attain fusion results of superior quality and enhanced stability, thereby forming the proposed RAF-NLRGS framework, as illustrated in Figure \ref{fig34}.

	\subsection{RAF model for image registration and fusion}
	In this subsection, we commence by introducing the RAF model, a novel framework designed for the concurrent execution of HSI-MSI registration and fusion tasks. Subsequently, we delineate the associated solving algorithm, complemented by  the corresponding convergence analysis.
	
	\subsubsection{The RAF model}
	Given that HSIs typically encompass numerous spectral bands with high spectral correlation, we leverage subspace representation for $\mathcal{X}$, i.e.,
	\begin{equation}\label{D8}
		\mathcal{X} = \mathcal{L}\times_3\textbf{D},
	\end{equation} 
	where $\mathcal{L}\in \mathbb{R}^{M\times N\times L}$  is  the coefficient tensor,  $\textbf{D}\in \mathbb{R}^{H\times L}$ denodes the dictionary and $L$ represents the dimension of the subspace spanned by the columns of $\textbf{D}$. To estimate dictionary $\textbf{D}$, a commonly adopted assumption is that $\mathcal{Y}$ and $\mathcal{X}$ share a common spectral subspace. Employing this principle,  we estimate the dictionary  $\textbf{D}$ from the LR-HSI by applying truncated singular value decomposition (svd), i.e.,
	\begin{equation}\label{D9}
		\begin{array}{l}
			[\textbf{U},\Sigma,\textbf{V}]=\text{svd}(\textbf{Y}_{(3)})\quad \text{and}  \quad \textbf{D}=\textbf{U}(: , 1:L),
		\end{array}
	\end{equation}
	where  $\Sigma$ denotes the matrix whose diagonal elements are the singular values of $\textbf{Y}_{(3)}$. $\textbf{U}$ and $\textbf{V}$ are the left and right singular matrices, respectively.
	
	Equations (\ref{D8}) utilized the spectral correlation of HSI to reduce the dimensionality of the problem.  Next, we explore the spatial correlation of coefficient tensor $\mathcal{L}$ to further enhance the model's performance. Specifically, we adopt the tensor nuclear norm (TNN) to characterize the low-rankness of coefficient tensor. 

Moreover,  it is recognized that disparities in imaging satellites, illumination conditions, and acquisition times can engender significant spectral variations between images \cite{Guo2022}, i.e., $\mathcal{Y}$ and $\mathcal{Z}$ come from different HR-HSIs. Thus, we incorporate a linear transformation  to the spectral dimension of $\mathcal{Z}$.
	Finally, the RAF model is developed as follows:
	\begin{equation}\label{RAF}
		\begin{aligned}
			\min\limits_{\mathcal{L},\tau,\textbf{A},\mathcal{B}}  \quad& \Vert \mathcal{L}\Vert_{\text{TNN}},\\ s.t. \quad&
			\mathcal{H}_{spa}(\mathcal{L}\times_3\textbf{D})=\mathcal{Y}\circ\tau,  \\&  \mathcal{H}_{spec} (\mathcal{L}\times_3\textbf{D})=\mathcal{Z}\times_3 \textbf{A}+\mathcal{B},
		\end{aligned}
	\end{equation}
	where $\textbf{A}\in \mathbb{R}^{h\times h}$ and $\mathcal{B}\in\mathbb{R}^{M\times N\times h}$ are utilized to eradicate spectral discrepancies between images.

 \subsubsection{GGN algorithm for solving the RAF model}
	As a matter of fact, the equality constraint of (\ref{RAF}) is highly nonlinear due to the domain transformations $\tau$, which complicates the minimization task. Therefore, we employ the Generalized Gauss–Newton (GGN) algorithm, which involves leveraging a first-order approximation of $\tau$ when the variations in $\tau$ are relatively insignificant \cite{RASL}. Specifically, under small perturbation $\Delta\tau$, we have $\mathcal{Y}\circ(\tau+\Delta\tau)\approx \mathcal{Y}\circ\tau+\mathcal{J}(\Delta\tau)$, with $\mathcal{J}(\Delta\tau):=fold_3((\sum_{i=1}^H\textbf{J}_i\Delta\tau\epsilon_i\epsilon_i^\top)^\top)$, where $\textbf{J}_i=\dfrac{\partial}{\partial_{\xi}}\text{vec}(\mathcal{Y}_i\circ\xi)|_{\xi=\tau_i}$ is the Jacobian matrix of the $i$th image with respect to the transformation parameters $\tau_i$ and $\{\epsilon_i\}$ denotes the standard basis for $\mathbb{R}^H$. After adopting a local first-order approximation for a given $\tau^k$, we obatin the following subproblem:
	\begin{equation}\label{C9}
		\begin{aligned}
			\min\limits_{\mathcal{L},\Delta\tau,\textbf{A},\mathcal{B}}  \quad& \Vert \mathcal{L}\Vert_{\text{TNN}},\\ s.t. \quad&
			\mathcal{H}_{spa}(\mathcal{L}\times_3\textbf{D})=\mathcal{Y}\circ\tau^k+\mathcal{J}(\Delta\tau),  \\&  \mathcal{H}_{spec} (\mathcal{L}\times_3\textbf{D})=\mathcal{Z}\times_3 \textbf{A}+\mathcal{B}.
		\end{aligned}
	\end{equation}
	Considering that subproblem (\ref{C9}) constitutes a local linearization of RAF model at the point $\tau^k$, we proceed to treat $\tau^k + \Delta\tau$ as the next iterative estimate and undertake another cycle of linearization. The entire procedure described above is summarized in Algorithm \ref{AL1}.
	\begin{algorithm}
		\caption{ GGN algorithm for solving RAF model}
		\parskip 0.73mm
		\hspace*{0.02in} {\bf Input:} ${\mathcal{Y}}$, ${\mathcal{Z}}$, $\textbf{D}$.  
		
		\hspace*{0.02in} \textbf{Initialize:} ${\mathcal{L}}^{0}$,  $\tau^0$,  $\textbf{A}^0$, $\mathcal{B}^0$. Set $k=0$.
		
		\quad1: Compute Jacobian matrices at $\tau_i^k$:
		$$
		\textbf{J}_i=\dfrac{\partial}{\partial x}\left( \dfrac{\text{vec}(\mathcal{Y}(:,:,i)\circ x)}{\Vert \text{vec}(\mathcal{Y}(:,:,i)\circ x) \Vert} \right)|_{x=\tau_i^k}, i=1,...H;
		$$
		
		\quad2: Compute $(\mathcal{L}^{k+1}, \Delta\tau^{k+1}, \textbf{A}^{k+1}, \mathcal{B}^{k+1})$  by solving (\ref{C9});
		
		\quad3: Update $\tau^{k+1}=\tau^k+\Delta\tau^{k+1}$;
		
		%
		
		\quad4: If converge, algorithm terminates;\quad  
		
		\qquad else, set $k = k+1$  and goto step1;
		
		\hspace*{0.02in} {\bf Output:}  $\mathcal{L}^{k+1}$, $\tau^{k+1}$, $\textbf{A}^{k+1}, \mathcal{B}^{k+1}$.
		\label{AL1}
	\end{algorithm}
	
	For the termination criterion of the algorithm, we define
	\begin{equation*}
		\kappa_{raf}:=\dfrac{\Vert \mathcal{H}_{spec}(\mathcal{Y}\circ\tau^{k+1})-\mathcal{H}_{spa}(\mathcal{Z}\times_3 \textbf{A}^{k+1}+\mathcal{B}^{k+1}) \Vert}{\Vert \mathcal{H}_{spa}(\mathcal{Z})\Vert}.
	\end{equation*}
	The algorithm terminates when $\kappa_{raf}$ falls below a predetermined threshold.  Next, we delve into the  convergence properties  associated with  Algorithm \ref{AL1}.
	\begin{theorem} 
		Suppose the operator $\mathcal{H}_{spa}$ satisfies REC($s$) on set $\mathbb{X}_{s}:=\{\mathcal{X}\in \mathbb{R}^{M\times N \times H}~|~rank(\mathcal{X}_{(3)})\le s\}$, where $s$ denotes the dimension of the null space of $\mathcal{H}_{spa}(\cdot)$,  then Algorithm \ref{AL1} converges quadratically in the neighborhood of any strongly unique local minimum of (\ref{RAF}).
	\end{theorem}
	
	The proof is provided in the Appendix B.
	\begin{remark}
		Within the framework of the RAF model, the obtained $\mathcal{X}$ can be decomposed into $\mathcal{L}\times_3\textbf{D}$, which indicates that $rank(\mathcal{X}_{(3)}) \le rank(\mathcal{L}_{(3)})\le L$. Consequently, $L$ serves as an upper threshold on the value of $s$, inherently suggesting that $s$ is constrained from becoming excessively large.
	\end{remark}
	
	\subsubsection{sGS-ADMM for solving model (\ref{C9})}
	Although model (\ref{C9}) is convex, the direct utilization of ADMM for solving it may fail to converge given that the number of variables surpasses two \cite{Chen2014}. 
	Therefore, we employ the symmetric Gauss–Seidel ADMM \cite{Li2016, Chen2017}. Firstly, by denoting $\bar{\mathcal{Z}}:=\mathcal{Z}\times_3 \textbf{A}+\mathcal{B}$, the augmented Lagrangian function of (\ref{C9}) is defined as
	\begin{equation}\label{C10}
		\begin{aligned}
		  \Theta(\mathcal{L},\Delta\tau,\textbf{A},\mathcal{B},\mathcal{O}_1,\mathcal{O}_2):=&  \dfrac{\mu_{1}}{2}\Vert \mathcal{H}_{spa}(\mathcal{L}\times_3\textbf{D})-(\mathcal{Y}\circ\tau^k+\mathcal{J}(\Delta\tau))+\dfrac{\mathcal{O}_1}{\mu_1} \Vert_F^2\\&+\dfrac{\mu_{2}}{2}\Vert \mathcal{H}_{spec} (\mathcal{L}\times_3\textbf{D})-\bar{\mathcal{Z}}+\dfrac{\mathcal{O}_2}{\mu_2} \Vert_F^2 +\Vert \mathcal{L} \Vert_{\text{TNN}},
		\end{aligned}
	\end{equation}
	where the symbols $\mathcal{O}_1$ and $\mathcal{O}_2$ represent Lagrangian multipliers.  $\mu_1$ and $\mu_2$  are two positive penalty parameters.  The iterative scheme of sGS-ADMM is given explicitly by
	\begin{equation}\label{C12}
		\begin{aligned}
			\Delta\tau^{t+\frac{1}{2}}&=\text{arg}\min\limits_{\Delta\tau}\{ \Theta(\mathcal{L}^t,\Delta\tau,\textbf{A}^t,\mathcal{B}^t,\mathcal{O}_1^t,\mathcal{O}_2^t) \},\\
			\mathcal{L}^{t+1}&=\text{arg}\min\limits_{\mathcal{L}}\{ \Theta(\mathcal{L},\Delta\tau^{t+\frac{1}{2}},\textbf{A}^t,\mathcal{B}^t,\mathcal{O}_1^t,\mathcal{O}_2^t) \},\\
			\Delta\tau^{t+1}&=\text{arg}\min\limits_{\Delta\tau}\{ \Theta(\mathcal{L}^{t+1},\Delta\tau,\textbf{A}^t,\mathcal{B}^t,\mathcal{O}_1^t,\mathcal{O}_2^t) \},\\
			(\textbf{A}^{t+1},\mathcal{B}^{t+1})&=\text{arg}\min\limits_{\textbf{A},\mathcal{B}}\{ \Theta(\mathcal{L}^{t+1},\Delta\tau^{t+1},\textbf{A},\mathcal{B},\mathcal{O}_1^t,\mathcal{O}_2^t) \},\\
	\mathcal{O}_1^{t+1}&=\mathcal{O}_1^t+\rho\mu_1(\mathcal{H}_{spa}(\mathcal{L}^{t+1}\times_3\textbf{D})-(\mathcal{Y}\circ\tau^k+\mathcal{J}(\Delta\tau^{t+1})),\\
			\mathcal{O}_2^{t+1}&=\mathcal{O}_2^t+\rho\mu_2(\mathcal{H}_{spec}(\mathcal{L}^{t+1}\times_3\textbf{D})-\bar{\mathcal{Z}}^{t+1}),
		\end{aligned}
	\end{equation}
	where $\rho\in(0,(1+\sqrt{5})/2)$  is the step length. Subsequently,  we proceed to solve each of the subproblems.
	
	\textbf{a.} The subproblem $\Delta\tau$ is  
	\begin{equation}\label{C18}
		\begin{aligned}
			\min\limits_{\Delta\tau} \enspace \Vert\mathcal{J}(\Delta\tau)+\mathcal{Y}\circ\tau^k-\mathcal{H}_{spa}(\mathcal{L}\times_3\textbf{D})-\frac{\mathcal{O}_1}{\mu_1} \Vert_F^2.
		\end{aligned}
	\end{equation}
	The optimal solution of (\ref{C18}) can be  given  by
	\begin{equation}\label{d21}
		\begin{aligned}
			\Delta\tau=\sum_{i=1}^{H}\textbf{J}_i^{\dagger}\textbf{M}\varepsilon_i\varepsilon_i^\top,
		\end{aligned}
	\end{equation}
	where $\textbf{M}:=(\mathcal{H}_{spa}(\mathcal{L}\times_3\textbf{D})+\frac{\mathcal{O}_1}{\mu_1}-\mathcal{Y}\circ\tau^k)_{(3)}^\top$ and $\textbf{J}_i^{\dagger}$ denotes the Moore-Penrose pseudoinverse of $\textbf{J}_i$.
	
	\textbf{b.} The subproblem $\mathcal{L}$ is 
	\begin{equation}\label{C13}
		\begin{aligned}
			&\min\limits_{\mathcal{L}}\enspace \dfrac{\mu_{1}}{2}\Vert \mathcal{H}_{spa}(\mathcal{L}\times_3\textbf{D})-(\mathcal{Y}\circ\tau^k+\mathcal{J}(\Delta\tau))+\frac{\mathcal{O}_1}{\mu_1} \Vert_F^2+\dfrac{\mu_{2}}{2}\Vert \mathcal{H}_{spec}(\mathcal{L}\times_3\textbf{D})-\bar{\mathcal{Z}}+\frac{\mathcal{O}_2}{\mu_2}\Vert_F^2+\Vert \mathcal{L} \Vert_{\text{TNN}}.
		\end{aligned}
	\end{equation}
	By introducing the variable $\mathcal{G}=\mathcal{L}$, (\ref{C13}) is reformulated as
	\begin{equation*}\label{C14}
		\begin{aligned}
			&\min\limits_{\mathcal{L},\mathcal{G}}\enspace \dfrac{\mu_{1}}{2}\Vert \mathcal{H}_{spa}(\mathcal{L}\times_3\textbf{D})-(\mathcal{Y}\circ\tau^k+\mathcal{J}(\Delta\tau))+\frac{\mathcal{O}_1}{\mu_1} \Vert_F^2+\dfrac{\mu_{2}}{2}\Vert \mathcal{H}_{spec}(\mathcal{L}\times_3\textbf{D})-\bar{\mathcal{Z}}+\frac{\mathcal{O}_2}{\mu_2}\Vert_F^2+\Vert \mathcal{G} \Vert_{\text{TNN}},\\
			&~s.t.\quad \mathcal{L}=\mathcal{G}.
		\end{aligned}
	\end{equation*}
	The augmented Lagrangian function  is 
	\begin{equation*}\label{C15}
		\begin{aligned}
			&\min\limits_{\mathcal{L},\mathcal{G},\mathcal{O}}\enspace \dfrac{\mu_{1}}{2}\Vert \mathcal{H}_{spa}(\mathcal{L}\times_3\textbf{D})-(\mathcal{Y}\circ\tau^k+\mathcal{J}(\Delta\tau))+\frac{\mathcal{O}_1}{\mu_1} \Vert_F^2+\dfrac{\mu_{2}}{2}\Vert \mathcal{H}_{spec}(\mathcal{L}\times_3\textbf{D})-\bar{\mathcal{Z}}+\frac{\mathcal{O}_2}{\mu_2}\Vert_F^2+\Vert \mathcal{G} \Vert_{\text{TNN}},\\
			&\qquad+\dfrac{\nu}{2}\Vert \mathcal{L}-\mathcal{G}+\dfrac{\mathcal{O}}{\nu} \Vert_F^2.
		\end{aligned}
	\end{equation*}
	The variable  $\mathcal{L}$ can be updated by soloving the optimality condition  with respect to $\mathcal{L}$, which is a Sylvester equation:
	\begin{equation}\label{C166}
		\textbf{H}_1\mathcal{L}_{(3)}+\mathcal{L}_{(3)}\textbf{H}_2=\textbf{H}_3.
	\end{equation}
	Given that $\textbf{D}^\top\textbf{D}=\textbf{I}$,  we can denote
	\begin{equation}
		\begin{aligned}
			&\textbf{H}_1:=\mu_2(\textbf{RD})^\top\textbf{RD}+\nu\textbf{I}_{L},\\
			&\textbf{H}_2:=\mu_1\textbf{BS}(\textbf{BS})^\top,\\
			&\textbf{H}_3:=\mu_1\textbf{P}(\textbf{BS})^\top+\mu_2(\textbf{RD})^\top\textbf{Q}+(\nu\mathcal{G}-\mathcal{O})_{(3)},
		\end{aligned}
	\end{equation}
	where
	\begin{equation*}
		\begin{aligned}
			&\textbf{P}=(\mathcal{Y}\circ\tau^k+\mathcal{J}(\Delta\tau)-\mathcal{O}_1/\mu_1)_{(3)},\\
			&\textbf{Q}=(\bar{\mathcal{Z}}-\mathcal{O}_2/\mu_2)_{(3)}.
		\end{aligned}
	\end{equation*}
	Equation (\ref{C166}) has a closed-form solution which can be solved efficiently by Algorithm \ref{AL22}. 
	\begin{algorithm}
		\caption{A closed-form solution of (\ref{C166})}
		\parskip 0.7mm
		\hspace*{0.02in} {\bf Input:} $\textbf{H}_1$, $\textbf{H}_2$, $\textbf{H}_3$, $\textbf{B}$, $\textbf{S}$.
		
		\quad1: Eigen-decomposition of $\textbf{B}: \textbf{B}=\textbf{FKF}^H$;
		
		\quad2: $\widetilde{\textbf{K}}=\textbf{K}(\textbf{1}_d\otimes\textbf{1}_{mn})$;
		
		\quad3: Eigen-decomposition of $\textbf{H}_1:\textbf{H}_1=\textbf{Q}_1\Lambda\textbf{Q}_1^H$;
		
		\quad4: $\widetilde{\textbf{H}}_3=\textbf{Q}_1^{-1}\textbf{H}_3\textbf{F}$;
		
		\quad5: \textbf{for} $i= 1:L$ \textbf{do}
		
		\quad6:  $\widetilde{\textbf{L}}(i,:)=\lambda_i^{-1}(\widetilde{\textbf{H}}_3)_i-\lambda_i^{-1}(\widetilde{\textbf{H}}_3)_i\widetilde{\textbf{K}}(\lambda_id\textbf{I}_n+\sum_{t=1}^{d}\textbf{K}_t^2)\widetilde{\textbf{K}}^H$;
		
		\quad7: \textbf{end} \textbf{for}
		
		\quad8: Set $\textbf{L}= \textbf{Q}_1\widetilde{\textbf{L}}\textbf{F}^H$; 
		
		\hspace*{0.02in} {\bf Output:}  $\mathcal{L}=fold_3(\textbf{L})$.
		\label{AL22}
	\end{algorithm}

	In this algorithm, $\lambda_i, i=1,...,L$ are the diagonal elements of $\Lambda$ and $d$ is the spatial degradation factor. Further particulars of the algorithm can be located in reference \cite{sy}.

	The $\mathcal{G}$ subproblem is 
	\begin{equation}\label{C16}
		\begin{aligned}
			\mathcal{G}^{t+1}\in\text{arg}\min\limits_{\mathcal{G}}\Vert \mathcal{G} \Vert_{\text{TNN}}+\dfrac{\nu}{2}\Vert \mathcal{G}-\mathcal{L}-\dfrac{\mathcal{O}}{\nu}\Vert_F^2.
		\end{aligned}
	\end{equation}
	Capitalizing on tensor Singular Value Thresholding (t-SVT, \cite{TSVD}), model (\ref{C16}) admits a closed-form solution, that is
	\begin{equation}\label{C17}
		\begin{aligned}
			\mathcal{G}^{t+1}=\mathcal{U}*\mathcal{D}*\mathcal{V}^{\mathcal{H}},
		\end{aligned}
	\end{equation}
	where $\mathcal{U}*\mathcal{S}*\mathcal{V}^{\mathcal{H}}$ represents the t-SVD of $(\mathcal{L}+\mathcal{O}/\nu)$. $\mathcal{D}$ and $\mathcal{S}$ are f-diagonal tensors with  $\mathcal{D}=\text{ifft}(\max\{\hat{\mathcal{S}}-\frac{1}{\nu},0\},[\enspace],3)$, where $\hat{\mathcal{S}}:=\text{fft}(\mathcal{S},[\enspace],3)$.
	Lastly, the iterative update for the Lagrangian multiplier $\mathcal{O}$ can be given by
	$$
	\mathcal{O}=\mathcal{O}+\nu(\mathcal{L}-\mathcal{G}).
	$$
	The variables $\mathcal{L}, \mathcal{G}$ and $\mathcal{O}$ are updated alternately, ultimately enabling us to solve the subproblem (\ref{C13}).
	
	\textbf{c.} The subproblem $(\textbf{A},\mathcal{B})$ can be reformulated as
	\begin{equation}\label{D22}
		\begin{aligned}
			\min\limits_{\textbf{A},\mathcal{B}}\enspace \Vert (\textbf{A},\mathcal{B}_{(3)})\textbf{Z}-\textbf{RD}\mathcal{L}_{(3)} \Vert_F^2,
		\end{aligned}
	\end{equation}
	where $\textbf{Z}:=(\mathcal{Z}_{(3)},\textbf{I})^\top$ and $\textbf{I}$ is the unit matrix. Then variables $\textbf{A}$ and $\mathcal{B}$ can be updated by
	\begin{equation}\label{D23}
		\begin{aligned}
			(\textbf{A},\mathcal{B}_{(3)})=\textbf{RD}\mathcal{L}_{(3)}\textbf{Z}^\dagger.
		\end{aligned}
	\end{equation}

Finally,  the pseudocode of above process is summarized into Algorithm \ref{AL2}.
	\begin{algorithm}
		\caption{sGS-ADMM algorithm for solving  (\ref{C9})}
		\parskip 0.72mm
		\hspace*{0.02in} {\bf Input:} ${\mathcal{Y}}$, ${\mathcal{Z}}$, $\tau^k, \mu_1, \mu_2, \nu,  \theta$.  
		
		\hspace*{0.02in} \textbf{Initialize:} ${\mathcal{L}}^{0}$, ${\mathcal{G}}^{0}$, $\Delta\tau^0$, $\textbf{A}^{0}, \mathcal{B}^0$. Set $t=0$.
		
		\quad1: Compute $\Delta\tau^{t+\frac{1}{2}}$  by  (\ref{d21});
		
		\quad2: Compute $\mathcal{L}^{t+1}$  by solving (\ref{C13});
		
		\quad3: Compute $\Delta\tau^{t+1}$  by  (\ref{d21});
		
		\quad4: Compute $(\textbf{A}^{t+1},\mathcal{B}^{t+1})$  by   (\ref{D23});
		
		\quad5: Update multipliers $\mathcal{O}_1^{t+1}$ and $\mathcal{O}_2^{t+1}$ as (\ref{C12}).
		
		\quad6: If converge, algorithm terminates;\quad   
		
		\qquad else, set $t = t+1$ and goto step1;
		
		\hspace*{0.02in} {\bf Output:}  $\mathcal{L}^{t+1}, \Delta\tau^{t+1}$, $\textbf{A}^{t+1}$ and $\mathcal{B}^{t+1}$.
		\label{AL2}
	\end{algorithm}

 Note that the objective function of (\ref{C9})  is nonsmooth with respect to $\mathcal{L}$, and other blocks vanishes. By [Theorem 3,\cite{Li2016}],  we can show the convergence of Algorithm \ref{AL2}, which is summarized in the following theorem.
	\begin{theorem}\label{T1} Suppose that $\rho\in(0,(1+\sqrt{5})/2)$. Let the sequence $\{\mathcal{L}^t,\Delta\tau^t,(\textbf{A}^t,\mathcal{B}^t),\mathcal{O}_1^t,\mathcal{O}_2^t\}_{t\in\mathbb{N}}$ be generated by Algorithm \ref{AL2}. Then the sequence $\{\mathcal{L}^t,\Delta\tau^t,(\textbf{A}^t,\mathcal{B}^t)\}_{t\in\mathbb{N}}$ converges to an optimal solution of (\ref{C9}) and the sequence $\{\mathcal{O}_1^t,\mathcal{O}_2^t\}_{t\in\mathbb{N}}$ converges to an optimal solution to the dual of (\ref{C9}).
	\end{theorem}

	\subsection{The NLRGS model and algorithm}
	Considering that the RAF model performs multiple tasks simultaneously, it may not consistently  yield the optimal fusion outcome. Therefore, in this section, we propose the novel NLRGS model, specifically designed to enhance the fusion performance and robustness.
	
	\subsubsection{The NLRGS model}
	Presently, the majority of fusion algorithms primarily  focus on designing strategies to reconstruct the principal image content with superior quality. Nevertheless, they frequently fail to capitalize on the potential of extracting additional information from the residual components within the images. This oversight may lead to an inefficient use of the available image data, and a missed opportunity for enhancing the overall fusion performance. Consequently, we propose to partition the target $\mathcal{X}$ into two distinct components: a primary component, $\widetilde{\mathcal{L}}$, and a residual component, $\widetilde{\mathcal{E}}$, such that $\mathcal{X}$ can be expressed as $\mathcal{X} = \widetilde{\mathcal{L}} + \widetilde{\mathcal{E}}$. The core objective is to reconstruct both $\widetilde{\mathcal{L}}$ and $\widetilde{\mathcal{E}}$ by thoroughly exploring and leveraging their unique characteristics and information content.
	
	With respect to the primary component $\widetilde{\mathcal{L}}$,  we adopt the subspace representation approach as utilized in (\ref{D8}).
	In terms of dictionary selection, we employ the same approach:
	\begin{equation}
		\begin{array}{l}
			[\textbf{U},\Sigma,\textbf{V}]=\text{svd}(\textbf{Y}_{R})\quad \text{and}  \quad \textbf{D}_{\mathcal{L}}=\textbf{U}(: , 1:L_1),
		\end{array}
	\end{equation}
	where $\textbf{Y}_R:=unfold_3(\mathcal{Y}_R)$ and $\mathcal{Y}_R$ represents the registered HSI obtained from RAF model. Observing that the coefficient tensor under dictionary $\textbf{D}_{\mathcal{L}}$ retains a comprehensive spatial structure, we incorporate the nonlocal self-similarity inherent in images, which clusters similar patches together to form 4th-order tensors. This is designed to bolster the low-rank prior characteristic of the data.  
	
	Regarding the residual component $\widetilde{\mathcal{E}}$, directly updating it can be computationally expensive due to its size being equivalent to that of the target $\mathcal{X}$. Therefore, we also project $\widetilde{\mathcal{E}}$ into a low-dimensional subspace along the spectral mode, i.e.,
	\begin{equation}\label{C21}
		\widetilde{\mathcal{E}} = \mathcal{E}\times_3\textbf{D}_{\mathcal{E}},
	\end{equation} 
	where $\mathcal{E}\in \mathbb{R}^{M\times N\times L_2}$  represents  the coefficient tensor,  $\textbf{D}_{\mathcal{E}}\in \mathbb{R}^{H\times L_2}$ denodes the dictionary and $L_2$ is the dimension of the subspace spanned by the columns of $\textbf{D}_{\mathcal{E}}$.
	This not only reduces the dimensionality of the problem but also enhances the robustness of the high-frequency information in $\widetilde{\mathcal{E}}$. As for the estimation of $\textbf{D}_{\mathcal{E}}$, we intuitively consider that the subspace generated by the column vectors of $\textbf{D}_{\mathcal{E}}$ should have no intersection with the subspace spanned by $\textbf{D}_{\mathcal{L}}$ associated with the primary component. Therefore, we select basis vectors from the orthogonal complement of $\textbf{D}_{\mathcal{L}}$ to construct $\textbf{D}_{\mathcal{E}}$, that is:
	\begin{equation}\label{Cc9}
		\begin{array}{l}
			\textbf{D}_{\mathcal{E}}=\textbf{U}( :~ , L_1+1:L_1+L_2),
		\end{array}
	\end{equation}
	with the condition that $L_1+L_2\le H$.
	Now, let us turn our attention to the coefficient $\mathcal{E}$. From a foundational standpoint,  if a particular feature manifests in one spectral band, it is highly probable that this feature will also be present in adjacent bands.  To exploit this inherent correlation, a group sparse prior is imposed along the spectral dimension as a regularization strategy.  Consequently,  the NLRGS model is formulated as follows:
	\begin{equation}\label{NLRGS}
		\begin{aligned}
			\min\limits_{\mathcal{L},\mathcal{E}}  \quad& \alpha\sum_{i=1}^{n}\Vert \mathcal{L}_i\Vert_{\psi}+\beta\Vert \mathcal{E} \Vert_{l_F^\psi},\\ s.t. \quad&
			\mathcal{H}_{spa}(\mathcal{L}\times_3\textbf{D}_{\mathcal{L}}+\mathcal{E}\times_3\textbf{D}_{\mathcal{E}})=\mathcal{Y}_{R},  \\&  \mathcal{H}_{spec} (\mathcal{L}\times_3\textbf{D}_{\mathcal{L}}+\mathcal{E}\times_3\textbf{D}_{\mathcal{E}})=\mathcal{Z},
		\end{aligned}
	\end{equation}
	where $\alpha$ and $\beta$ are hyper-parameters to control the weight of different terms. The tensors  $\mathcal{L}_i$, $i = 1,2,\dots, n$ represent the clustered blocks of $\mathcal{L}$. They originally were fourth-order tensors; however, we have merged their final two dimensions to transform them into third-order ones. The specific process can be referred to in \cite{LTMR}.  The$\Vert \mathcal{L} \Vert_{\psi}$ and $\Vert \mathcal{E} \Vert_{l_F^\psi}$ are defined as 
	$$
	\Vert \mathcal{L} \Vert_{\psi}:=\dfrac{1}{L_1}\sum_{k=1}^{L_1}\sum_{i=1}^{\min\{M,N\}}\psi(\sigma_i(\hat{\textbf{L}}^{(k)})),
	$$
	and
	$$
	\Vert \mathcal{E} \Vert_{l_F^\psi}:=\sum_{i=1}^{M}\sum_{j=1}^{N}\psi(\Vert \mathcal{E}(i,j,:) \Vert_F),
	$$
	respectively, where $\psi$ is selected as MCP function.  Next, we present relevant proximal mappings for solving the model.
	\begin{theorem}\cite{Yu2024}\label{B9} 
		For a three-order tensor $\mathcal{A}$ and $\alpha\textgreater0$, a minimizer to
		\begin{equation*}
			\min\limits_{\mathcal{X}} \alpha\Vert \mathcal{X} \Vert_{\psi}+\dfrac{1}{2}\Vert \mathcal{X}-\mathcal{A} \Vert_F^2
		\end{equation*}
		is given as follows:
		\begin{equation*}
			\mathcal{X}=\mathcal{U}*\mathcal{D}*\mathcal{V}^{\mathcal{H}},
		\end{equation*}
		where $\mathcal{A}=\mathcal{U}*\mathcal{S}*\mathcal{V}^{\mathcal{H}}$ denotes the t-SVD of $\mathcal{A}$. $\mathcal{D}$ and $\mathcal{S}$ are f-diagonal tensors with $\hat{\textbf{D}}_{i,i}^{(k)}=\text{prox}_{\alpha\psi}(\hat{\textbf{S}}_{i,i}^{(k)})$.
	\end{theorem}

According to the definition of $\Vert \cdot \Vert_{l_F^\psi}$,  we have
	\begin{equation} \label{D26}
		\begin{aligned}
			\text{prox}_{\Vert\cdot\Vert_{l_F^\psi}}(\mathcal{Z})&=
			\text{arg}\min_{\mathcal{X}}\enspace \Vert\mathcal{X}\Vert_{l_F^\psi}+\dfrac{1}{2}\Vert \mathcal{X}-\mathcal{Z} \Vert^2_F \\&=\text{arg}\min_{\mathcal{X}}\sum_{i=1}^{M}\sum_{j=1}^{N}(\psi(\Vert \mathcal{X}_{i,j} \Vert)+\dfrac{1}{2}\Vert \mathcal{X}_{ij}-\mathcal{Z}_{ij} \Vert^2_F),
		\end{aligned}
	\end{equation}
	where subscript $(\cdot)_{ij}$ represents the $(i,j)$th mode-3 fiber of a tensor. Note that (\ref{D26}) is separable, it is equivalent to 
	\begin{equation}\label{D27}
		\begin{aligned}
			\left(\text{prox}_{\beta \Vert\cdot\Vert_{l_F^\psi}}(\mathcal{Z})\right)_{ij}&=\text{arg}\min_{\textbf{x}}\enspace \psi(\textbf{x})+\dfrac{1}{2}\Vert \textbf{x}-\mathcal{Z}_{ij} \Vert^2_F \\&=\text{prox}_{\psi}(\mathcal{Z}_{ij}),
		\end{aligned}
	\end{equation}
	for $i=1,...M, j=1,...N$. Thus, problem (\ref{D26}) can be solved fiber-by-fiber according to [Theorem 1,\cite{Wen2018}].
	
	\subsubsection{Error bound of NLRGS model}
	Subsequently,  we establish the error bound for the NLRGS model, which investigates the discrepancy between the model's optimal solution and the true solution under worst-case conditions.
	
	\begin{theorem}
		Let $(\hat{\mathcal{L}},\hat{\mathcal{E}})$ be an optimal solution of the NLRGS model and $(\mathcal{L}^\star,\mathcal{E}^\star)$ represent the underlying true values. Assume that the operator  $\mathcal{H}_{spec}(\cdot)$ satisfies REC($s$) on $\mathbb{X}_{s}:=\{\mathcal{X}|\Vert \mathcal{X} \Vert_{\psi}\le s\}$, where $s:=r\psi(r^{-\frac{1}{2}}\Vert \hat{\mathcal{L}}-\mathcal{L^\star} \Vert_F)$ with $r:=\min \{M,N\}$.  Then,  provided that $\beta>\alpha r v$, we have
		\begin{equation}
			\begin{aligned}
				\psi(\Vert \hat{\mathcal{E}}-\mathcal{E}^\star\Vert_F)\le\Vert \hat{\mathcal{E}}-\mathcal{E}^\star \Vert_{l_F^\psi}
				\le\frac{2\beta\Vert \mathcal{E}^\star \Vert_{l_F^\psi}}{\beta-\alpha rv}
			\end{aligned}
		\end{equation}
		and
		\begin{equation}
			\begin{aligned}
				\psi(\Vert \hat{\mathcal{L}}-\mathcal{L}^\star \Vert_F)\le\Vert \hat{\mathcal{L}}-\mathcal{L}^\star \Vert_{l_F^\psi}
				\le \frac{2hw\beta\Vert \mathcal{E}^\star \Vert_{l_F^\psi}}{\beta-\alpha rv},
			\end{aligned}
		\end{equation}
		where $v$,  $w$ are positive constants and $h:=MN$. Furthermore, it follows that:
		\begin{equation}\label{D30}
			\begin{aligned}
				\psi(\Vert \hat{\mathcal{L}}-\mathcal{L}^\star \Vert_F+\Vert \hat{\mathcal{E}}-\mathcal{E}^\star \Vert_F)
				\le \frac{2\beta(hw+1)\Vert \mathcal{E}^\star \Vert_{l_F^\psi}}{\beta-\alpha rv}.
			\end{aligned}
		\end{equation}
	\end{theorem}
	The proofs are  provided in the Appendix C.
	
	\begin{remark}
		\textbf{a).} From the proof, constants $v$ and $w$ are positively correlated with $\frac{\Vert\textbf{RD}_{\mathcal{E}}\Vert}{\sqrt{r}\xi}$ and $\frac{\Vert\textbf{RD}_{\mathcal{E}}\Vert}{\sqrt{h}\xi}$, respectively. As inequality (\ref{D30}) indicates, reducing the values of $v$ and $w$ serves to a lower error bound. This necessitates maximizing the value of $\xi$, which in turn implies that the primary component $\mathcal{L}$ should ideally have a lower rank. The above analysis suggests that a lower rank of the true principal component leads to a superior recovery outcome.
		\textbf{b).} Considering that nonlocal clustering  serves to augment the low-rank property of the data without fundamentally altering its characteristics, we elect not to incorporate this operation in our theoretical analysis.
		
	\end{remark} 
	
	\subsubsection{Proximal Alternating Optimization algorithm}
	In order to solve the NLRGS model, we penalize the constraints into the objective, leading to the following formulation:
	\begin{equation}\label{C23}
		\begin{aligned}
			\min\limits_{\mathcal{L},\mathcal{E}}  g(\mathcal{L},\mathcal{E}):&=\Vert \mathcal{H}_{spa}(\mathcal{L}\times_3\textbf{D}_{\mathcal{L}}+\mathcal{E}\times_3\textbf{D}_{\mathcal{E}})-\mathcal{Y}_{R} \Vert_F^2+\Vert \mathcal{H}_{spec} (\mathcal{L}\times_3\textbf{D}_{\mathcal{L}}+\mathcal{E}\times_3\textbf{D}_{\mathcal{E}})-\mathcal{Z} \Vert_F^2 \\
			&+\alpha\sum_{i=1}^{n}\Vert \mathcal{L}_i \Vert_{\psi}+\beta\Vert \mathcal{E} \Vert_{l_F^\psi}.
		\end{aligned}
	\end{equation}
	Considering that the objective function of model (\ref{C23}) involves two variables, $\mathcal{L}$ and $\mathcal{E}$, we incorporate  proximal terms into the subproblem and adopt an alternating optimization strategy. In each iteration, we update $\mathcal{L}$ and $\mathcal{E}$ alternately, while keeping the other variable fixed. This proposed algorithm, referred to as the Proximal Alternating Optimization (PAO) algorithm, is employed to solve (\ref{C23}). Specifically, the subproblem $\mathcal{L}$ is
	
	\begin{equation}
		\begin{aligned}\label{C24}
			\min\limits_{\mathcal{L}} &\quad\Vert \mathcal{H}_{spa}(\mathcal{L}\times_3\textbf{D}_{\mathcal{L}})-(\mathcal{Y}_R-\mathcal{H}_{spa}(\mathcal{E}^k\times_3\textbf{D}_{\mathcal{E}})) \Vert_F^2+\Vert \mathcal{H}_{spec}(\mathcal{L}\times_3\textbf{D}_{\mathcal{L}})-(\mathcal{Z}-\mathcal{H}_{spec}(\mathcal{E}^k\times_3\textbf{D}_{\mathcal{E}})) \Vert_F^2 \\&
			+\alpha\sum_{i=1}^{n}\Vert \mathcal{L}_i \Vert_{\psi}+\dfrac{\lambda}{2}\Vert \mathcal{L}-\mathcal{L}^k \Vert_F^2,
		\end{aligned}
	\end{equation}
	and subproblem $\mathcal{G}$ is
	\begin{equation}\label{C25}
		\begin{aligned}
			\min\limits_{\mathcal{E}}&\quad \Vert\mathcal{H}_{spa}(\mathcal{E}\times_3\textbf{D}_{\mathcal{E}})-(\mathcal{Y}_R-\mathcal{H}_{spa}(\mathcal{L}^{k+1}\times_3\textbf{D}_{\mathcal{L}})) \Vert_F^2+\Vert \mathcal{H}_{spec}(\mathcal{E}\times_3\textbf{D}_{\mathcal{E}})-(\mathcal{Z}-\mathcal{H}_{spec}(\mathcal{L}^{k+1}\times_3\textbf{D}_{\mathcal{L}})) \Vert_F^2 \\&
			+\beta\Vert \mathcal{E} \Vert_{l_F^\psi}+\dfrac{\lambda}{2}\Vert \mathcal{E}-\mathcal{E}^k \Vert_F^2,
		\end{aligned}
	\end{equation}
	where $\lambda$ is a positive proximal parameter.
	The pseudocode of above process is summarized into Algorithm \ref{AL3}.
	\begin{algorithm}
		\caption{ PAO algorithm for solving the model (\ref{C23})}
		\parskip 0.7mm
		\hspace*{0.02in} {\bf Input:} ${\mathcal{Y}_R}$, ${\mathcal{Z}}$, $ \alpha, \beta, \lambda,  \theta$.  
		
		\hspace*{0.02in} \textbf{Initialize:} ${\mathcal{L}}^{0}$, ${\mathcal{E}}^{0}$. Set $k=0$.
		
		\quad1: Compute $\mathcal{L}^{k+1}$  by  solving the subproblem (\ref{C24});
		
		\quad2: Compute $\mathcal{E}^{k+1}$  by  solving the subproblem (\ref{C25});

		\quad4: If converge, algorithm terminates;\quad 
  
            \qquad else, set $k = k+1$   and goto step1;
		
		\hspace*{0.02in} {\bf Output:}  $\mathcal{X}=\mathcal{L}^{k+1}\times_3\textbf{D}_\mathcal{L}+\mathcal{E}^{k+1}\times_3\textbf{D}_\mathcal{E}$.
		\label{AL3}
	\end{algorithm}

For the termination criterion, we  define 
	\begin{equation*}
		\kappa_{\mathcal{L}}:=\dfrac{\Vert \mathcal{L}^{k+1}-\mathcal{L}^{k} \Vert}{\Vert\mathcal{L}^{k}\Vert}
	\end{equation*}
	and
	\begin{equation*}
		\kappa_{\mathcal{E}}:=\dfrac{\Vert \mathcal{E}^{k+1}-\text{prox}_{\frac{\beta}{\lambda}\Vert \cdot\Vert_{l_F^\psi}}(\mathcal{E}^k-\nabla_{\mathcal{E}}q(\mathcal{L}^{k+1},\mathcal{E}^{k+1})) \Vert}{1+\Vert\mathcal{L}^{k+1}\Vert+\Vert\mathcal{E}^{k+1}\Vert},
	\end{equation*}
	where $q$ is defined as
	\begin{equation}
		\begin{aligned}
			q(\mathcal{L},\mathcal{E}):=\Vert \mathcal{H}_{spa}(\mathcal{L}\times_3\textbf{D}_{\mathcal{L}}+\mathcal{E}\times_3\textbf{D}_{\mathcal{E}})-\mathcal{Y}_{R} \Vert_F^2+\Vert \mathcal{H}_{spec} (\mathcal{L}\times_3\textbf{D}_{\mathcal{L}}+\mathcal{E}\times_3\textbf{D}_{\mathcal{E}})-\mathcal{Z} \Vert_F^2 .
		\end{aligned}
	\end{equation}
	Then we define $\kappa_{nlrgs}:=\text{max}\{\kappa_{\mathcal{L}}, \kappa_{\mathcal{E}}\}$.
	The algorithm terminates when $\kappa_{nlrgs}$ descends beneath a predetermined tolerance threshold. Subsequently, we present the convergence analysis for Algorithm \ref{AL3}, demonstrating that the sequence generated globally converges to a critical point of model (\ref{C23}).

	
	
	\begin{theorem}\label{T2} Let $\{\mathcal{V}^{k}=(\mathcal{L}^k,\mathcal{E}^k)\}_{k\in\mathbb{N}}$ be the sequence generated by Algorithm \ref{AL3}. If  $\{\mathcal{V}^{k}\}_{k\in\mathbb{N}}$ is bounded, then $\{\mathcal{V}^{k}\}_{k\in\mathbb{N}}$ converges to a critical point of $g$ and  
		$$
		\sum_{k=0}^{+\infty}||\mathcal{V}^{k+1}-\mathcal{V}^{k}|| < +\infty.
		$$
	\end{theorem}
	The proofs are  provided in the Appendix D.
	
	To tackle model (\ref{C24}), we continue to employ the ADMM algorithm, initially introducing a new variable $\mathcal{G}=\mathcal{L}$. This  reformulates model (\ref{C24}) into the following form:
	\begin{equation*}\label{C26}
		\begin{aligned}
			\min\limits_{\mathcal{L},\mathcal{G}}& \quad\Vert \mathcal{H}_{spa}(\mathcal{L}\times_3\textbf{D}_{\mathcal{L}})-(\mathcal{Y}_R-\mathcal{H}_{spa}(\mathcal{E}^k\times_3\textbf{D}_{\mathcal{E}})) \Vert_F^2+\Vert \mathcal{H}_{spec}(\mathcal{L}\times_3\textbf{D}_{\mathcal{L}})-(\mathcal{Z}-\mathcal{H}_{spec}(\mathcal{E}^k\times_3\textbf{D}_{\mathcal{E}})) \Vert_F^2 \\&
			+\alpha\sum_{i=1}^{n}\Vert \mathcal{G}_i \Vert_{\psi}+\dfrac{\lambda}{2}\Vert \mathcal{L}-\mathcal{L}^k \Vert_F^2,\\
			s.t &\quad \mathcal{L}=\mathcal{G},
		\end{aligned}
	\end{equation*}
	where $\mathcal{G}_i, i=1,...,n$ represents the $i$th clustering block of $\mathcal{G}$. The augmented Lagrange function for the model  can be expressed as follows:
	\begin{equation*}\label{C27}
		\begin{aligned}
			\min\limits_{\mathcal{L},\mathcal{G},\mathcal{O}^k}& \quad\Vert \mathcal{H}_{spa}(\mathcal{L}\times_3\textbf{D}_{\mathcal{L}})-(\mathcal{Y}_R-\mathcal{H}_{spa}(\mathcal{E}^k\times_3\textbf{D}_{\mathcal{E}})) \Vert_F^2+\Vert \mathcal{H}_{spec}(\mathcal{L}\times_3\textbf{D}_{\mathcal{L}})-(\mathcal{Z}-\mathcal{H}_{spec}(\mathcal{E}^k\times_3\textbf{D}_{\mathcal{E}})) \Vert_F^2 \\
			+&\alpha\sum_{i=1}^{n}\Vert \mathcal{G}_i \Vert_{\psi}+\dfrac{\mu_{\mathcal{L}}}{2}\Vert\mathcal{L}-\mathcal{G}+\dfrac{\mathcal{O}^k}{\mu_{\mathcal{L}}} \Vert_F^2+\dfrac{\lambda}{2}\Vert \mathcal{L}-\mathcal{L}^k \Vert_F^2,
		\end{aligned}
	\end{equation*}
	where $\mathcal{O}^k$ is the augmented Lagrange multiplier and  $\mu_\mathcal{L}$ is the penalty parameter.
	Then we proceed to tackle $\mathcal{L}$  and $\mathcal{G}$ subproblem alternately. Concretely, $\mathcal{L}$ subproblem is 
	\begin{equation}\label{D14}
		\begin{aligned}
			\min\limits_{\mathcal{L}} &\quad\Vert \mathcal{H}_{spa}(\mathcal{L}\times_3\textbf{D}_{\mathcal{L}})-(\mathcal{Y}_R-\mathcal{H}_{spa}(\mathcal{E}^k\times_3\textbf{D}_{\mathcal{E}})) \Vert_F^2+\Vert \mathcal{H}_{spec}(\mathcal{L}\times_3\textbf{D}_{\mathcal{L}})-(\mathcal{Z}-\mathcal{H}_{spec}(\mathcal{E}^k\times_3\textbf{D}_{\mathcal{E}})) \Vert_F^2 \\
			&+\dfrac{\mu_{\mathcal{L}}}{2}\Vert \mathcal{L}-\mathcal{G}^j+\dfrac{\mathcal{O}^k_j}{\mu_{\mathcal{L}}} \Vert_F^2+\dfrac{\lambda}{2}\Vert \mathcal{L}-\mathcal{L}^k \Vert_F^2.
		\end{aligned}
	\end{equation}
	The optimal condition of (\ref{D14})  satisfies the form of equation (\ref{C166}) since $\textbf{D}_{\mathcal{L}}^\top\textbf{D}_{\mathcal{L}}=\textbf{I}$, thus we can still utilize Algorithm \ref{AL22} to solve subproblem $\mathcal{L}$ efficiently, where
	\begin{equation*}
		\begin{aligned}
			&\textbf{H}_1:=(\textbf{RD}_{\mathcal{L}})^\top\textbf{RD}_{\mathcal{L}}+\frac{\mu_\mathcal{L}+\lambda}{2}\textbf{I}_{L_1},\\
			&\textbf{H}_2:=\textbf{BS}(\textbf{BS})^\top,\\
			&\textbf{H}_3:=\textbf{D}_{\mathcal{L}}^\top(\textbf{Y}_R-\textbf{D}_{\mathcal{E}}\mathcal{E}_{(3)}\textbf{BS})(\textbf{BS})^\top+(\textbf{RD}_{\mathcal{L}})^\top\mathcal{Z}_{(3)}\\&\quad-(\textbf{RD}_{\mathcal{L}})^\top\textbf{RD}_{\mathcal{E}}\mathcal{E}_{(3)}+\frac{(\mu_\mathcal{L}\mathcal{G}^j-\mathcal{O}_j^k+\lambda\mathcal{L}^k)_{(3)}}{2}.
		\end{aligned}
	\end{equation*}
	The $\mathcal{G}$ subproblem is 
	\begin{equation*}\label{C30}
		\begin{aligned}
			\mathcal{G}^{j+1}\in\text{arg}\min\limits_{\mathcal{G}} \sum_{i=1}^{n}(\Vert \mathcal{G}_i \Vert_{\psi}+\dfrac{\mu_{\mathcal{L}}}{2\alpha}\Vert \mathcal{G}_i-\mathcal{L}^{j+1}_i-\dfrac{\mathcal{O}^k_{ji}}{\mu_{\mathcal{L}}} \Vert_F^2),
		\end{aligned}
	\end{equation*}
	which can be divided into $n$ independent subproblems:
	\begin{equation}\label{C31}
		\begin{aligned}
			\min\limits_{\mathcal{G}_i} \enspace\alpha\Vert \mathcal{G}_i \Vert_{\psi}+\dfrac{\mu_{\mathcal{L}}}{2}\Vert \mathcal{G}_i-\mathcal{L}^{j+1}_i-\dfrac{\mathcal{O}^k_{ji}}{\mu_{\mathcal{L}}} \Vert_F^2.
		\end{aligned}
	\end{equation}
	Then, by Theorem \ref{B9}, we obtain a closed form solution
	\begin{equation}\label{C32}
		\begin{aligned}
			\mathcal{G}_i^{j+1}=\mathcal{U}_i*\mathcal{D}_i*\mathcal{V}_i^\mathcal{H},
		\end{aligned}
	\end{equation}
	where $\mathcal{U}_i*\mathcal{S}_i*\mathcal{V}_i^\mathcal{H}$ represents the t-SVD of ($\mathcal{L}_i^{j+1}+\dfrac{\mathcal{O}_{ji}^k}{\mu_{\mathcal{L}}}$) and $\hat{\textbf{D}}_{i,t,t}^{(k)}=\text{prox}_{\frac{\alpha}{\mu_{\mathcal{L}}}\psi}(\hat{\textbf{S}}_{i,t,t}^{(k)})$. We summarize the pseudocode of above process into Algorithm \ref{AL4}.
	
	\begin{algorithm}
		\caption{ ADMM algorithm for solving the model (\ref{C24})}
		\parskip 0.68mm
		\hspace*{0.02in} {\bf Input:} ${\mathcal{Y}_R}$, ${\mathcal{Z}}$, $ \alpha,\mu_\mathcal{L}, \lambda, \theta$.  
		
		\hspace*{0.02in} \textbf{Initialize:} ${\mathcal{L}}^{0}$, ${\mathcal{G}}^{0}$, $\mathcal{O}^k_0$. Set $j=0$.
		
		\quad1: Compute $\mathcal{L}^{j+1}$  by  solving the subproblem (\ref{D14});
		
		\quad2: Compute $\mathcal{G}^{j+1}_i$  by  equation (\ref{C32}) and then consolidate
		
		\qquad  them to obtain $\mathcal{G}^{j+1}$;
		
		\quad3: Compute $\mathcal{O}_{j+1}^k$  by  $\mathcal{O}_{j+1}^k=\mathcal{O}_j^k+\mu_{\mathcal{L}}(\mathcal{L}^{j+1}-\mathcal{G}^{j+1})$;
		
		\quad3: If converge, algorithm terminates;\quad   
		
		\qquad else, set $j = j+1$ and goto step1;
		
		\hspace*{0.02in} {\bf Output:} $\mathcal{L}=\mathcal{L}^{j+1}$.
		\label{AL4}
	\end{algorithm}
	
	In a similar manner, by introducing a new variable $\mathcal{F}=\mathcal{E}$, the model (\ref{C25}) can be transformed into
	\begin{equation*}\label{D18}
		\begin{aligned}
			\min\limits_{\mathcal{E},\mathcal{F}} &\quad \Vert\mathcal{H}_{spa}(\mathcal{E}\times_3\textbf{D}_{\mathcal{E}})-(\mathcal{Y}_R-\mathcal{H}_{spa}(\mathcal{L}^{k+1}\times_3\textbf{D}_{\mathcal{L}})) \Vert_F^2+\Vert \mathcal{H}_{spec}(\mathcal{E}\times_3\textbf{D}_{\mathcal{E}})-(\mathcal{Z}-\mathcal{H}_{spec}(\mathcal{L}^{k+1}\times_3\textbf{D}_{\mathcal{L}})) \Vert_F^2 \\&
			+\beta\Vert \mathcal{E} \Vert_{l_F^\psi}+\dfrac{\lambda}{2}\Vert \mathcal{E}-\mathcal{E}^k \Vert_F^2,\\
			s.t &\quad \mathcal{E}=\mathcal{F}.\quad
			\vspace{-3pt}
		\end{aligned}
	\end{equation*}
	We address the subproblem $\mathcal{E}$ by employing the BCD algorithm, which necessitates initially incorporating the constraints into the objective function. This process leads us to formulate the following model:
	\begin{equation}\label{D19}
		\begin{aligned}
			\min\limits_{\mathcal{E},\mathcal{F}} &\quad \Vert\mathcal{H}_{spa}(\mathcal{E}\times_3\textbf{D}_{\mathcal{E}})-(\mathcal{Y}_R-\mathcal{H}_{spa}(\mathcal{L}^{k+1}\times_3\textbf{D}_{\mathcal{L}})) \Vert_F^2+\Vert \mathcal{H}_{spec}(\mathcal{E}\times_3\textbf{D}_{\mathcal{E}})-(\mathcal{Z}-\mathcal{H}_{spec}(\mathcal{L}^{k+1}\times_3\textbf{D}_{\mathcal{L}})) \Vert_F^2 \\&
			+\beta\Vert \mathcal{F} \Vert_{l_F^\psi}+\dfrac{\mu_{\mathcal{E}}}{2}\Vert \mathcal{E}-\mathcal{F} \Vert_F^2+\dfrac{\lambda}{2}\Vert \mathcal{E}-\mathcal{E}^k \Vert_F^2.
		\end{aligned}
	\end{equation}
	To solve the model (\ref{D19}), we update $\mathcal{E}$ and $\mathcal{F}$ alternately. More precisely, the subproblem $\mathcal{E}$ can be represented as
	\begin{equation}\label{D20}
		\begin{aligned}
			\min\limits_{\mathcal{E}} &\quad \Vert\mathcal{H}_{spa}(\mathcal{E}\times_3\textbf{D}_{\mathcal{E}})-(\mathcal{Y}_R-\mathcal{H}_{spa}(\mathcal{L}^{k+1}\times_3\textbf{D}_{\mathcal{L}})) \Vert_F^2+\Vert \mathcal{H}_{spec}(\mathcal{E}\times_3\textbf{D}_{\mathcal{E}})-(\mathcal{Z}-\mathcal{H}_{spec}(\mathcal{L}^{k+1}\times_3\textbf{D}_{\mathcal{L}})) \Vert_F^2 \\&+\dfrac{\mu_{\mathcal{E}}}{2}\Vert \mathcal{E}-\mathcal{F} \Vert_F^2+\dfrac{\lambda}{2}\Vert \mathcal{E}-\mathcal{E}^k \Vert_F^2.
		\end{aligned}
	\end{equation}
	Similar to  subproblem $\mathcal{L}$, the optimal condition of  (\ref{D20}) is also a Sylvester equation since $\textbf{D}_{\mathcal{E}}^\top\textbf{D}_{\mathcal{E}}=\textbf{I}$, which can be solved efficiently by Algorithm \ref{AL22}. The $\mathcal{F}$ subproblem is 
	\begin{equation}\label{D21}
		\begin{aligned}
			\mathcal{F}^{t+1}\in\text{arg}\min\limits_{\mathcal{F}} \beta\Vert \mathcal{F}\Vert_{l_F^\psi}+\dfrac{\mu_{\mathcal{E}}}{2}\Vert \mathcal{F}-\mathcal{E}^{t+1} \Vert_F^2,
		\end{aligned}
	\end{equation}
	which  has a closed form solution according to (\ref{D27}). 
	The pseudocode  is summarized in Algorithm \ref{AL5}.
	\begin{algorithm}
		\caption{ BCD for solving the model (\ref{C25})}
		\parskip 0.66mm
		\hspace*{0.02in} {\bf Input:} ${\mathcal{Y}_R}$, ${\mathcal{Z}}$, $ \beta, \mu_\mathcal{E}, \lambda$, $\theta$.  
		
		\hspace*{0.02in} \textbf{Initialize:} ${\mathcal{E}}^{0}$, ${\mathcal{F}}^{0}$. Set $t=0$.
		
		\quad1: Compute $\mathcal{E}^{t+1}$  by  solving  (\ref{D20});
		
		\quad2: Compute $\mathcal{F}^{t+1}$  by  solving  (\ref{D21});
		
		\quad3: If converge, algorithm terminates;\quad   
		
		\qquad else, set $t = t+1$ and goto step1;
		
		\hspace*{0.02in} {\bf Output:} $\mathcal{E}=\mathcal{E}^{t+1}$.
		\label{AL5}
	\end{algorithm}

\section{NUMERICAL EXPERIMENTS}\label{NN}
	In this section, to rigorously evaluate the effectiveness of the proposed RAF-NLRGS method, we conduct  a series of experiments on publicly available HSI datasets, including:
	
	\textit{1) Pavia University and Pavia Center \cite{Pavia}:} These two scenes were captured by the ROSIS sensor during a flight campaign conducted in Pavia. The Pavia University and Pavia Center datasets, following processing, respectively encompass 103 and 102 spectral bands, each spanning a wavelength spectrum from 0.43µm to 0.86µm.  We select 93 bands from both scenes and cropped a region of interest from the up-left corner, resulting in the image size of 256 × 256× 93. 
	To generate the LR-HSI  with a size of 64 × 64 × 93, we apply a uniform downsampling with a ratio of 4 to the HR-HSI. Additionally, the HR-MSI  is generated using a four-band IKONOS-like reflectance spectral response filter \cite{sy}.

	\textit{2) CAVE$^1$:} The CAVE dataset comprises images with 31 bands, spanning the spectral range from 400nm to 700nm at 10nm intervals. All images within the dataset are uniform in dimensions, measuring 512×512×31.  We select the `Face', `Peppers', and `Superballs' for our experiments. To generate the LR-HSIs, we systematically downsampled these images at different ratios. More precisely, the spatial dimensions of the `Face' and `Peppers' images are decreased by factors of 8 and 16, respectively, while the `Superballs' image is scaled down by a factor of 32. Besides, we obtain the MSIs with three bands by the Nikon D700 camera$^2$, i.e., h = 3.
	
	\textit{3) Real Dataset (GF1-GF5):} In the GF5-GF1 dataset,  the size of LR-HSI is 1161×1129×150, and the size of HR-MSI is 2322×2258×4. From the HSI, we isolate bands 11 to 110 and extracted a 1024x1024 pixel area from the top-left quadrant, employing a sampling step of 4, to give rise to an LR-HSI measuring 256x256 pixels across 100 bands. Concurrently, a 2048x2048 pixel segment is taken from the identical corner of the MSI, also utilizing a sampling interval of 2, producing an MSI of 1024x1024 pixels with 4 bands.

	\footnotetext[1]{https://www.cs.columbia.edu/CAVE/databases/multispectral/ }
	\footnotetext[2]{https://maxmax.com/spectral\_response.htm}
	\footnotetext[3]{Although the code for this paper has not been open-sourced, we deemed it necessary to compare our method with theirs, hence we reproduced their approach. All experimental results presented are from this reproduction.}
	Moreover,  to impartially assess  the quality of the reconstructed HR-HSI, we choose  the following  evaluation metrics: peak signal-to-noise ratio (PSNR) \cite{MSE},  structural similarity (SSIM) \cite{MSE}, relative dimensionless global error in synthesis (ERGAS) \cite{WR}, and spectral angle mapper  (SAM) \cite{WR}.
	
	\subsection{Compared methods}
	To assess the performance of our proposed RAF-NLRGS method, we compare it with several current state-of-the-art (SOTA) techniques. For the unregistered case, we selected for comparison with two well-known registration algorithms: the feature-based SIFT \cite{SIFT} algorithm and  intensity-based  NTG \cite{NTG} method. Then, we performed fusion on the registered images using classical algorithms: LTMR \cite{LTMR} and SCOTT \cite{P2020}. Additionally, we consider three related joint registration and fusion approaches: SHR$^3$ \cite{Fu2020}, NED \cite{NED} and DFMF \cite{Guo2022}. Specifically,  SHR is a technique that performs registration and fusion concurrently, utilizing spectral dictionary learning and sparse coding. NED is a two-stage algorithm, which  incorporates the interpolation process into unmixing for the fusion task. DFMF is a deep learning-based method comprising two components: a registration network and a fusion network.
	
	To independently verify the efficacy of the NLRGS model, we conduct comparisons against several SOTA fusion algorithms under the condition of perfect registration. Specifically, given that our approach embraces subspace representation (SR), tensor representation (TR), residual modeling and group sparsity prior, we draw comparisons against a classical SR-based method (Hysure \cite{Hysure}), some famous TR-based methods (SCOTT, LTMR \cite{LTMR} and LTTR \cite{LTTR}),  S$^4$-LRR \cite{S4}, which utilize the group sparsity and IR-TenSR \cite{IR} which emphasizes residual reconstruction. Additionally, we compare with  ZSL \cite{ZSL}, an excellent method based on Convolutional Neural Networks.
	The parameters are fine-tuned  to obtain the best outputs following the suggestions in the reference papers. 

 \subsection{Parameters Setting}
	In this subsection,  we outline the specific parameters configuration employed within our proposed methodology.
	
	\subsubsection{RAF model} 
	Overall, the RAF model is robust with respect to its parameters. In all scenarios, we set both augmentation penalty parameters, $\mu_1$ and $\mu_2$, to a value of 10. The subspace dimension $L$ is typically configured within the range of 2 to 4. The parameter $\nu$ in the subproblem is selected to 0.1, and the step size $\rho$ is set to the value of 1.618. Additionally, the MCP parameter $\theta$ is configured to 8.
	
	\subsubsection{NLRGS model} 
	As part of the post-processing stage for the RAF model, we strongly recommend maintaining consistency between the value of $L_1$ and $L$ within RAF model.  Meanwhile, the parameter $L_2$ can generally be retained at a value close to 20. 
	The hyperparameter $\alpha$ and $\beta$ are employed to adjust the weight of the low-rank and sparse term, which have been assigned a value within [1 $\times 10^{-3}$, $5 \times 10^{-3}$]. Moreover, we consistently set the MCP parameter $\theta$ to either 8 or 9 across all the datasets. Furthermore, there are two penalty parameters: $\mu_{\mathcal{L}}$ and $\mu_{\mathcal{E}}$, both of which are systematically adjusted within the interval of [\( 1 \times 10^{-3} \), \( 1 \times 10^{-2} \)]. $\lambda$ is the proximal parameter and we set it to $1\times 10^{-4}$. Regarding the nonlocal term, the number of patches $n$ is  configured to fall within the interval of $[150, 250]$. The dimensions of each patch cube are set to 6 or 7 with a 4-unit overlap.

 As for the stopping criterion, $\kappa_{raf}$ and $\kappa_{nlrgs}$ control whether the GGN and PAO algorithms terminate respectively. From Figure \ref{DF4}, setting them within the range from $1\times 10^{-2}$ to $1\times 10^{-5}$ is acceptable for different datasets.

\subsection{Experimental results under non-aligned conditions}
	In this subsection, our objective is to illustrate the effectiveness and robustness of the RAF-NLRGS approach in registration and fusion tasks through extensive experimental evaluations.
	
	\subsubsection{The experimental results}
	Specifically, we consider several geometric manipulations, encompassing three linear transforms: translations, rotations, and flipping, along with two nonlinear transforms: barrel and pincushion distortions. These transformations are individually applied to the HSI. Respective experimental outcomes pertaining to the `Pavia University', `Pavia Center', and `Face' datasets are recorded in Table \ref{DT2}.
\begin{table*}[hpbt]
		\renewcommand\arraystretch{1.36}
		\caption{The experimental results of RAF-NLRGS method on three HSI datasets.} 
		\vspace{-0.05cm}      
		\centering
		\label{DT2}               
		\begin{tabular}{p{1.3cm}p{1cm}|p{0.7cm}p{0.7cm}p{0.7cm}p{0.7cm}|p{0.7cm}p{0.7cm}p{0.7cm}p{0.7cm}|p{0.7cm}p{0.7cm}p{0.7cm}p{0.7cm}}
			\hline
			\textbf{Transform} & \textbf{Stage} & PSNR &SSIM & ERGAS& SAM &  PSNR &SSIM & ERGAS& SAM &  PSNR &SSIM & ERGAS& SAM 
			\\ \hline
			\multicolumn{14}{c}{\textbf{\qquad\qquad\qquad Pavia University (sf = 4)}\qquad\quad~\textbf{Pavia Center (sf = 4)} \qquad\qquad\quad \textbf{Face (sf = 8)}} \\
			\hline
			Translation & RAF & 41.97 & 0.989 & 1.263  & 2.254 &  44.57 & 0.993 & 1.062  & 2.988 &  41.29 & 0.970 & 1.644  & 15.601  \\ 
			& NLRGS	 & 43.57 & 0.991 & 1.063  & 2.075 & 45.79 & 0.994 & 0.953  & 2.652 &  42.86 & 0.976 & 1.502  & 15.315 \\
			\cline{2-14}	Rotation & RAF& 41.95 & 0.989 & 1.265  & 2.257& 44.66 & 0.993 & 1.047  & 2.855 & 41.28 & 0.962 & 1.789  & 15.862 \\ 
			& NLRGS	 & 43.55 & 0.991 & 1.066  & 2.078 & 45.72 & 0.994 & 0.962  & 2.682 & 42.77 & 0.977 & 1.534  & 15.251  \\ 
			\cline{2-14} Flip & RAF & 41.95 & 0.989 & 1.263  & 2.242&  44.34 & 0.993 & 1.104  & 2.950&41.20 & 0.956 & 1.788  & 15.787 \\
			& NLRGS	& 43.54 & 0.989 & 1.067  & 2.066 &  45.63 & 0.994 & 0.986  & 2.786 &  42.83 & 0.977 & 1.509 & 15.226  \\
			\hline 
			barrel & RAF& 41.94 & 0.989 & 1.272  & 2.265 & 44.28 & 0.993 & 1.253  & 3.129 & 40.24 & 0.962 & 1.816  & 15.095 \\
			& NLRGS	&  43.54 & 0.991 & 1.068  & 2.081	&  45.45 & 0.994 & 1.053  & 2.818 &  42.62 & 0.979 & 1.538  & 14.338  \\
			\cline{2-14}
			pincushion & RAF& 41.91 & 0.989 & 1.271  & 2.254 & 44.12 & 0.992 & 1.276  & 3.173 & 40.09 & 0.960 & 1.827  & 15.841 \\
			& NLRGS	&  43.53 & 0.991 & 1.070  & 2.017 & 45.32 & 0.994 & 1.054  & 2.810  &  42.41 & 0.976 & 1.596  & 15.362 \\
			\hline
		\end{tabular}
	\end{table*}
 \begin{figure*}[htbp]
		\vspace{-6pt}
		\centering
		\subfloat[HSI]{\includegraphics[width=2.2cm]{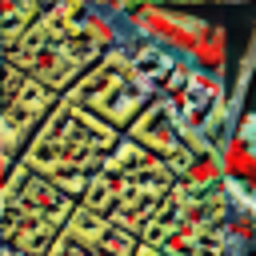}} 
		\subfloat[Translation]{\includegraphics[width=2.2cm]{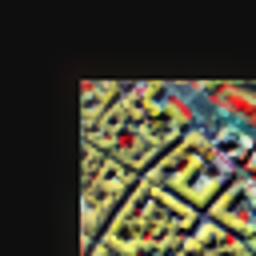}} 
		\subfloat[Rotation]{\includegraphics[width=2.2cm]{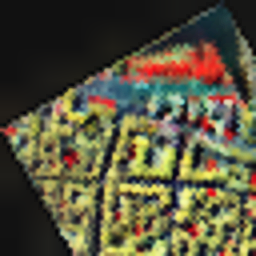}} 
		\subfloat[Flip]{\includegraphics[width=2.2cm]{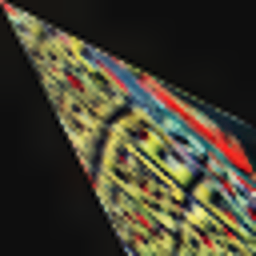}} 
		\subfloat[Barrel]{\includegraphics[width=2.2cm]{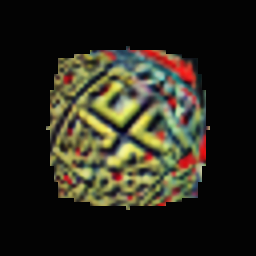}} 
		\subfloat[Pincushion]{\includegraphics[width=2.2cm]{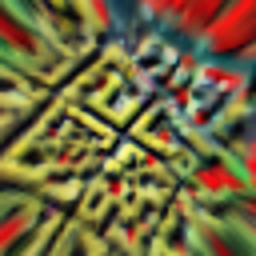}} 
		\subfloat[]{\includegraphics[width=2.2cm]{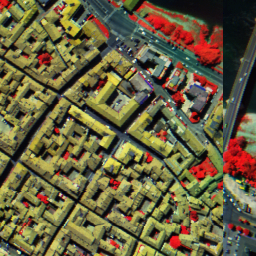}}
		
		\vspace{-7pt}
	~	\subfloat[MSI]{\includegraphics[width=2.2cm]{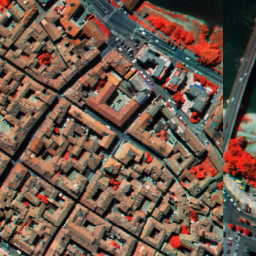}}
		\subfloat[]{\includegraphics[width=2.2cm]{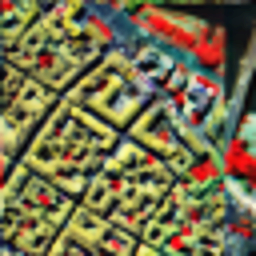}}
		\subfloat[]{\includegraphics[width=2.2cm]{overleaf_NLRGS/PC_rhsi}}
		\subfloat[]{\includegraphics[width=2.2cm]{overleaf_NLRGS/PC_rhsi}}
		\subfloat[]{\includegraphics[width=2.2cm]{overleaf_NLRGS/PC_rhsi}}
		\subfloat[]{\includegraphics[width=2.2cm]{overleaf_NLRGS/PC_rhsi}}
		\subfloat[]{\includegraphics[width=2.1cm]{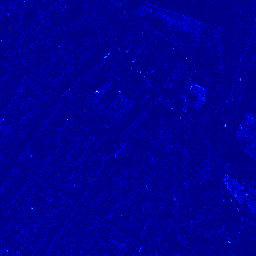}}\!
		\subfloat{\includegraphics[width=0.3cm,height=2.2cm]{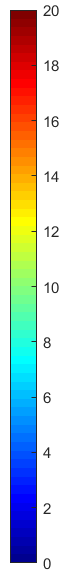}}
		
		\caption{Visualization of the registration and fusion efficacy for the RAF-NLRGS model (pseudocolor image, bands 20, 40, 60). (a) True HSI. (h) MSI. (b)-(f) Transformed HSIs.  (i)-(m) The corresponding registrated HSIs. (g) The fused HSI. (n) The error image.}
		\label{DF1}
		\vspace{-4pt}
	\end{figure*}
 
 The row labeled `RAF' showcases the evaluation metrics for images reconstructed via the RAF methodology, while the row marked `NLRGS' exhibits the corresponding metrics for images subsequent to their refinement using the NLRGS method. Clearly, our proposed method consistently delivers robust and commendable restoration results across a broad spectrum of transformation scenarios. Furthermore, after incorporating the registered images into the NLRGS model, the resulting fused image exhibits pronounced   improvements across multiple evaluation metrics.  Notably,  when subjected to nonlinear transformation conditions, our method achieves restoration outcomes comparable to those in linear transformation scenarios across diverse datasets. This validates the efficacy of the multi-step linearization process in the Generalized Gauss-Newton algorithm. Then we present in Figure \ref{DF1} a visual illustration of the restoration process, showcasing both the original input image and the recovered image from the `Pavia Center' dataset. It can be observed that, our method consistently achieves outstanding registration performance under various transformation scenarios. Furthermore, in contrast to traditional registration techniques, the registered HSI obtained by RAF automatically supplements the scene information to correspond with the MSI. This is an ancillary benefit of our method, enabling greater flexibility in application scenarios.

	\subsubsection{Comparisons with other methods} 
	To further verify the effectiveness of our proposed method, we incorporate a comparative analysis with several benchmark  registration and fusion techniques. These include twol registration methods: SIFT and NTG, along with two well-known fusion algorithms: LTMR and SCOTT. We initially upsampled the HSI and MSI along spatial  and spectral dimensions respectively. Subsequently, we applied band-wise registration using the corresponding algorithms before downsampling the images back to their original resolutions, thereby generating the registered HSI-MSI pairs. Ultimately, the fused image is produced by applying the fusion algorithm. Additionally, our method is compared with the joint registration and fusion algorithm SHR, NED as well as DFMF for a comprehensive evaluation. In our experiments, we deliberately simulate a scenario wherein each band of HSI is subjected to a 5-pixel translation in both spatial axes, followed by resizing back to the initial dimensions. The corresponding experimental results are  presented in Table \ref{DT5}.
	
		\begin{table*}[hpbt]
		\renewcommand\arraystretch{1.35}
		\caption{ The experimental results of compared methods on three HSI datasets.} 
		\vspace{-0.05cm}      
		\centering
		\label{DT5}               
		\begin{tabular}{p{1.98cm}|p{0.7cm}p{0.7cm}p{0.7cm}p{0.7cm}|p{0.7cm}p{0.7cm}p{0.7cm}p{0.7cm}|p{0.7cm}p{0.7cm}p{0.7cm}p{0.8cm}}
			\hline
			\textbf{Datasets}  & PSNR &SSIM & ERGAS& SAM  & PSNR &SSIM & ERGAS& SAM &  PSNR &SSIM & ERGAS& SAM 
			\\ \hline
			\multicolumn{13}{c}{\textbf{\qquad\quad Pavia University (sf = 4)}\qquad\qquad \textbf{Pavia Center (sf = 4)} \qquad\qquad\quad\textbf{Face (sf = 8)}} \\
			\hline
			SIFT-LTMR &  31.47& 0.955 & 3.894 & 4.032 &  27.54 & 0.836 & 7.400 & 9.038 &  35.70 & 0.941 & 2.737 & 19.443 \\ 
			\cline{2-13} SIFT-SCOTT & 30.65 & 0.923 & 4.285 & 4.441& 26.92 & 0.795 & 7.476 & 7.793 &  32.86 &0.920 & 4.005 & 23.196  \\
			\cline{2-13}
			NTG-LTMR &28.35 & 0.896 & 8.356 & 7.477& 28.32& 0.943& 6.505& 5.963&  34.13 & 0.923 & 3.623 & 20.820  \\ 
			\cline{2-13} NTG-SCOTT &29.33 & 0.899 & 8.220 &7.349&27.95  &0.879  & 6.632  &6.766 & 34.20 & 0.926 & 3.610 & 19.869  \\  
			\hline
			SHR& 40.36 & 0.988 & 1.424 & 2.476 &  43.21 & 0.991 &1.197 & 3.182 &  40.02 & 0.959 & 1.836 & 15.873  \\ 
			\cline{2-13} NED& 41.91 & 0.989 & 1.265 & 2.252  & 41.48 & 0.989 & 1.366 & 3.562 &  41.15 & 0.971 & 1.696 & 16.107  \\ 
			\cline{2-13} RAF-NLRGS &\textbf{43.57} & \textbf{0.991} & \textbf{1.063}  & \textbf{2.075}& \textbf{45.79} & \textbf{0.994} & \textbf{0.953}  & \textbf{2.652} &  \textbf{42.46} & \textbf{0.977} & \textbf{1.570}  & \textbf{15.413}  \\ 
			\hline
		\end{tabular}
	\end{table*}

\begin{figure*}[htbp]
		\centering
		\vspace{-7pt}
    
		\subfloat{\includegraphics[width=2cm]{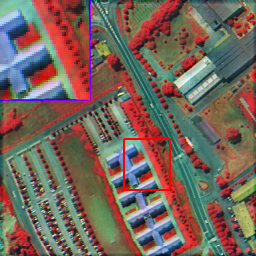}}\!
		\subfloat{\includegraphics[width=2cm]{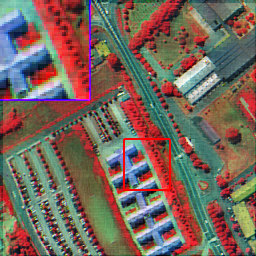}}\!
		\subfloat{\includegraphics[width=2cm]{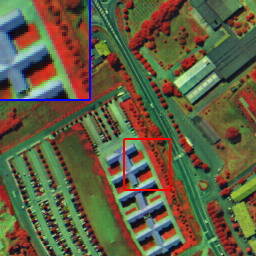}}\!
		\subfloat{\includegraphics[width=2cm]{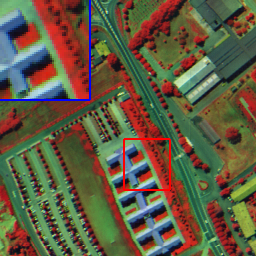}}
		\subfloat{\includegraphics[width=2cm]{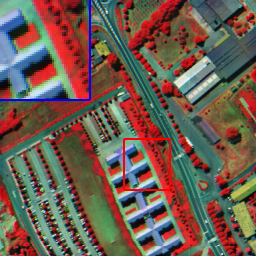}}\!
		\subfloat{\includegraphics[width=2cm]{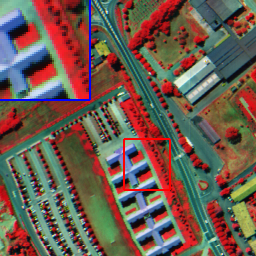}}\!
		\subfloat{\includegraphics[width=2cm]{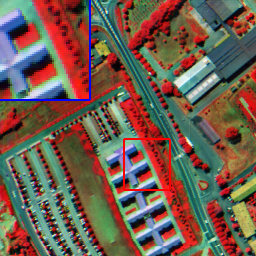}}\!
		\subfloat{\includegraphics[width=2cm]{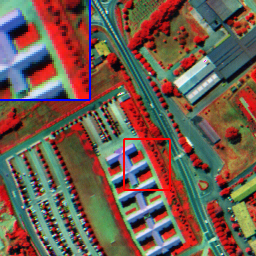}}

		\vspace{-9pt}
		\subfloat[SIFTLTMR]{\includegraphics[width=2cm]{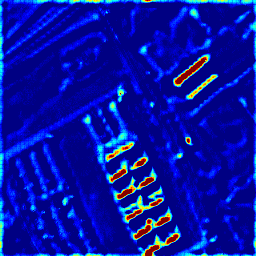}}\!
		\subfloat[SIFTSCOTT]{\includegraphics[width=2cm]{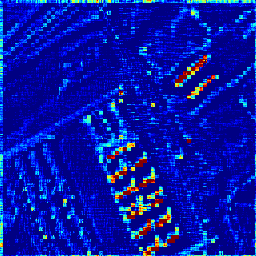}}\!
		\subfloat[NTGLTMR]{\includegraphics[width=2cm]{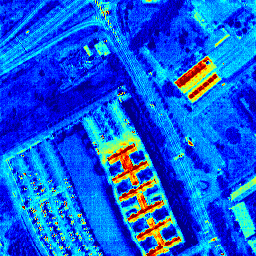}}\!
		\subfloat[NTGSCOTT]{\includegraphics[width=2cm]{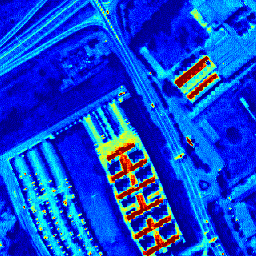}}\!
		\subfloat[SHR]{\includegraphics[width=2cm]{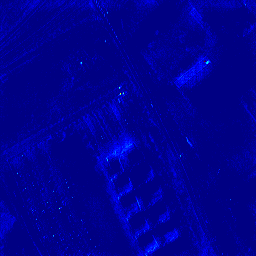}}\!
		\subfloat[NED]{\includegraphics[width=2cm]{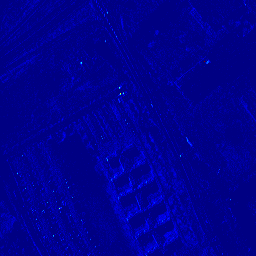}}\!
		\subfloat[Ours]{\includegraphics[width=2cm]{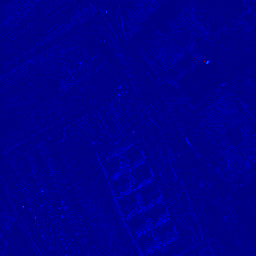}}\!
		\subfloat[GT]{\includegraphics[width=2cm]{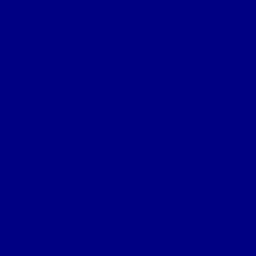}}

        \vspace{4pt}
		\includegraphics[width=9cm]{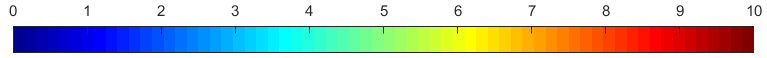}
		\caption{The first row presents the reconstructed images of the compared methods for Pavia University. The second row shows corresponding error images.\tiny\label{DF0}}
	\end{figure*}
	
	The presented tabular data illustrates that our proposed methodology persistently outperforms comparative methods across all evaluation metrics. The NED approach yielded suboptimal outcomes when applied to the `Pavia University' and `Face' datasets. Of particular interest is the fact that, the registration-prior-to-fusion strategy does not yield satisfactory outcomes. One plausible explanation for this might lie in the unregistered HSI-MSI pairs exhibit significant disparities in both spatial and spectral resolutions. Although resizing by upsampling makes their dimensions match, there still exists considerable divergence in their internal features. This inherent difference hampers the ability of registration algorithms to attain higher levels of precision. 
	Since DFMF  focuses on the registration and fusion of remote sensing images for large-scale scenes, we found its performance on simulated datasets to be unsatisfactory in our experiments. Consequently, we do not include a comparison with it here. To be convincing, we display its experimental  results on the `Face' dataset within the Appendix E. The corresponding restored images from the `Pavia University' dataset are visually presented in Figure \ref{DF0}.
	
	Observing the figures, the restoration outcomes based on SIFT primarily exhibit more flaws in edge details, whereas those based on NTG present clear spatial details but suffer from color discrepancies, which indicates imperfections in the recovery of spectral information. In comparison, the restoration outcomes showcased by SHR and NED illustrate improved restorative attributes. Notably, our proposed methodology demonstrates the minimal presence of imperfections. These experimental results furnish compelling evidence in support of the efficacy of our  approach.
	
	\subsubsection{Robustness evaluation}
\begin{figure}[htbp]
		\vspace{-15pt}
		\centering
		\subfloat[Translation]{\includegraphics[width=4cm]{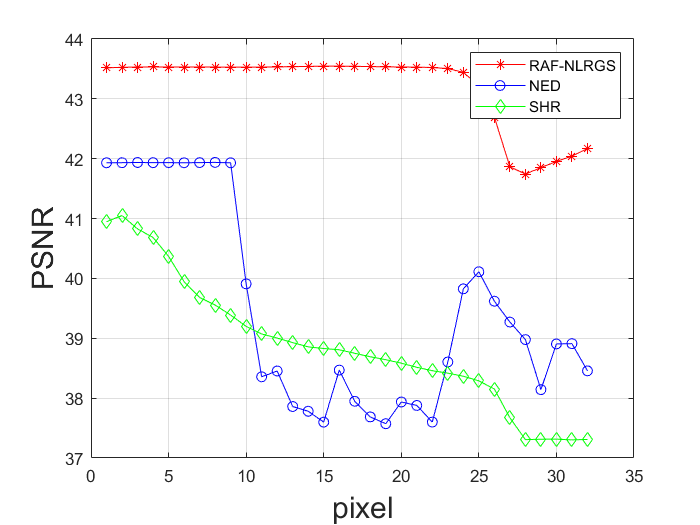}}
		\subfloat[Rotation]{\includegraphics[width=4cm]{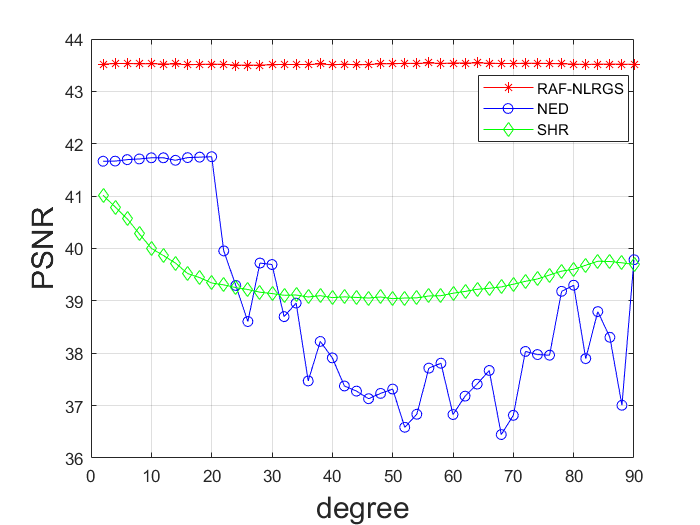}}
		\subfloat[Barrel]{\includegraphics[width=4cm]{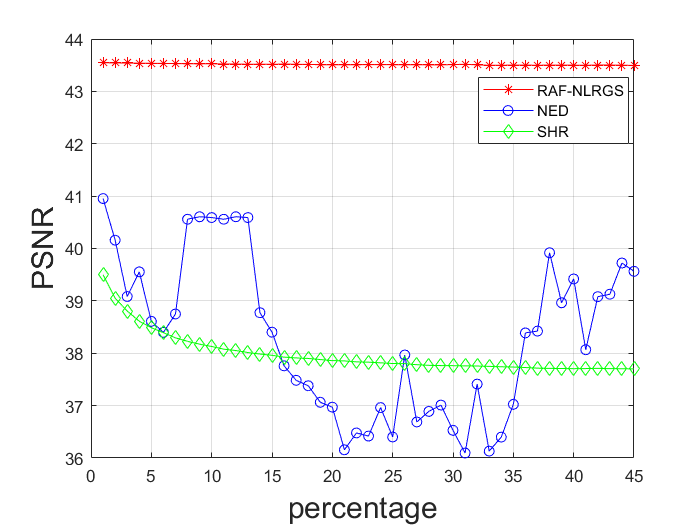}}
		\subfloat[Pincushion]{\includegraphics[width=4cm]{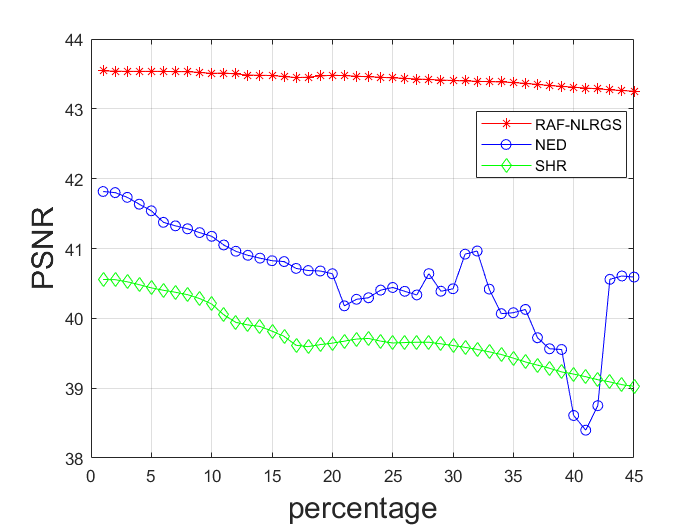}}
		\caption{The correlation between the image restoration quality of different methodologies and the scale of transformation on `Pavia University'.\tiny\label{DF3}}
		\vspace{-5pt}
	\end{figure}

	It is noteworthy that many HSI-MSI registration methodologies frequently demonstrate a conspicuous decline in performance when confronted with substantial image scaling transformations. To substantiate the robustness of our proposed approach against significant scale variations, we undertake a thorough examination of the correlation between the restoration quality attained by various methodologies and the extent of image scaling manipulations imposed. In light of the initial subpar performance observed in the first four methods during the preliminary experimental phase, our subsequent comparative analysis has been refined to concentrate specifically on SHR, NED, and RAF-NLRGS. The experimental results on `Pavia University' are graphically illustrated in Figure \ref{DF3}. Given that flipped scenarios rarely occur, we do not present the stability results here.

	The figures reveal that the RAF-NLRGS model exhibits optimal performance even under substantial scale modifications, all the while maintaining a notably consistent level of output quality.  The PSNR values of images restored via SHR exhibit a gradual diminution as the degree of spatial transformation escalates. In contrast, while NED performs admirably under modest scale variations, it encounters a decline in the consistency of image restoration as the intensity of transformation escalates. The experiments  conclusively attest to the robustness and superior performance of our proposed methodology.
	
	\subsubsection{Experiments on real-world datasets}
	Finally, we validate the effectiveness of our method on real-world datasets. The GF1-GF5 image pair are not pre-registered and the degradation operators remain unidentified. Our initial step necessitates the estimation of  degradation operators.  To achieve this, we adopt the strategy of further downsampling the spatial dimensions of both HSI and MSI by a factor of 4 to highlight spectral information while mitigating the effects caused by misregistration. Then, we employ the methodology described in \cite{Hysure} to infer the degradation operators, effectively addressing the following optimization model\footnotetext[1]{Since the kernel \textbf{B} often approximates a Gaussian distribution, we can also manually determine the  kernel \textbf{B} directly by parameter tuning, without needing to obtain it through an algorithm.}:
	$$
	\min\limits_{\textbf{B},\textbf{R}}\Vert \textbf{R}\mathcal{Y}_{(3)}-\mathcal{Z}_{(3)}\textbf{BS} \Vert_F^2+\lambda_r\Phi_r(\textbf{R})+\lambda_b\Phi_b(\textbf{B}),
	$$
	where $\Phi_r(\textbf{R})$ and $\Phi_b(\textbf{B})$ are the regularizer of $\textbf{R}$ and $\textbf{B}$, respectively$^1$. $\lambda_r$ and $\lambda_b$ are positive regularization parameters. Subsequently, the proposed  RAF-NLRGS can be  utilized to accomplish the registration and fusion tasks. The experimental outcomes of the comparative methods are depicted in Figure \ref{DF5} for visual comparison.

\begin{figure}[htbp]
		\centering

		\subfloat[Input HSI]{\includegraphics[width=4.3cm]{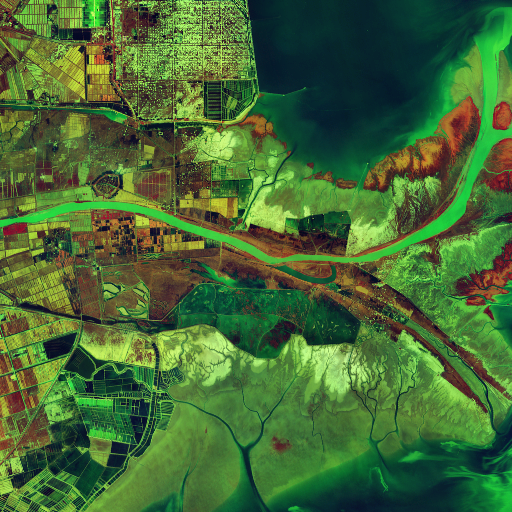}}\qquad
		\subfloat[Input MSI]{\includegraphics[width=4.3cm]{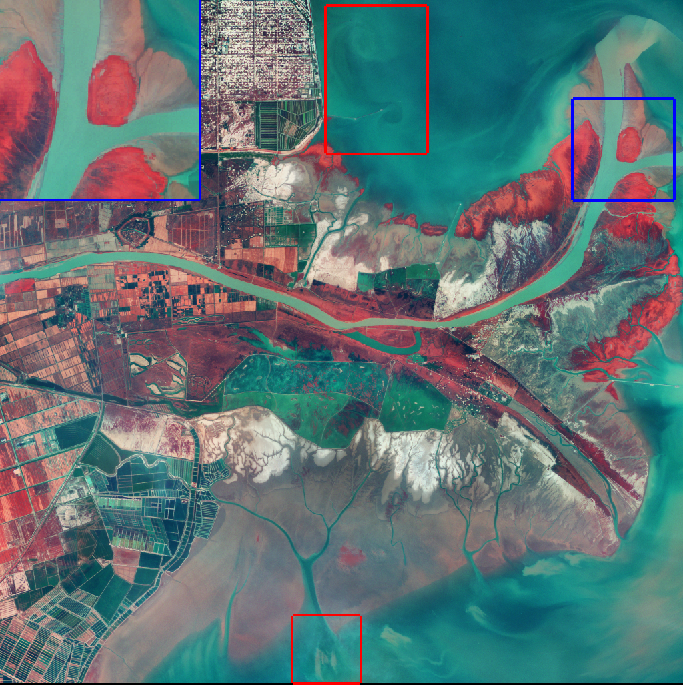}}

        \vspace{-5pt}
		\subfloat[NTG-Hysure]{\includegraphics[width=4.3cm]{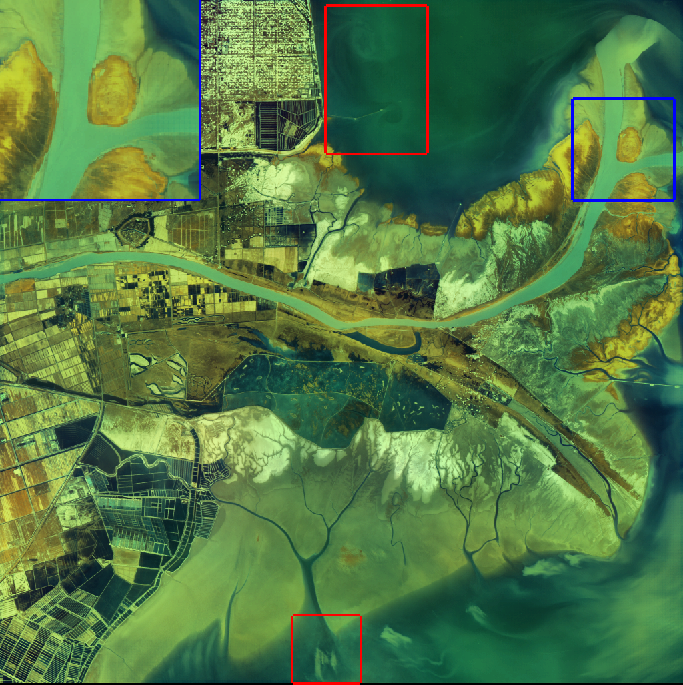}}\qquad
		\subfloat[NED]{\includegraphics[width=4.3cm]{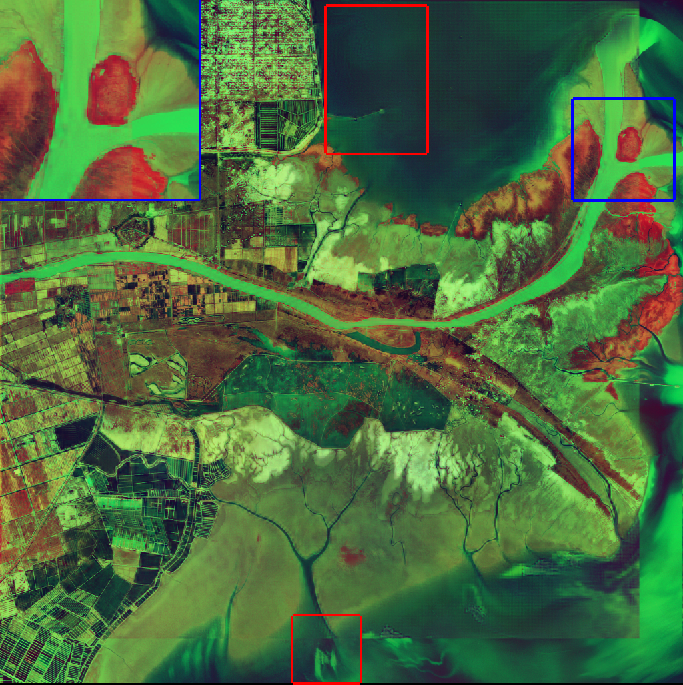}}

        \vspace{-5pt}
		\subfloat[DFMF]{\includegraphics[width=4.3cm]{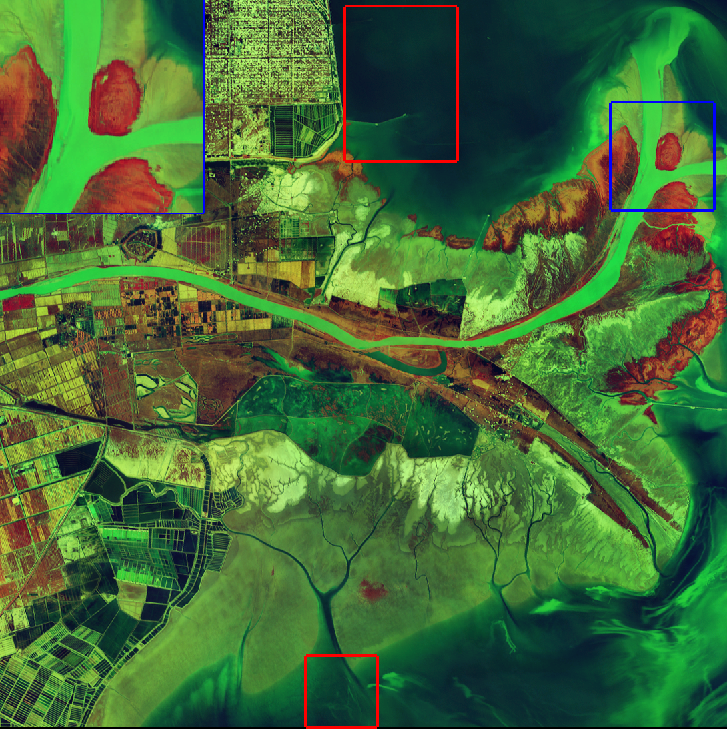}}\qquad
		\subfloat[RAF-NLRGS]{\includegraphics[width=4.3cm]{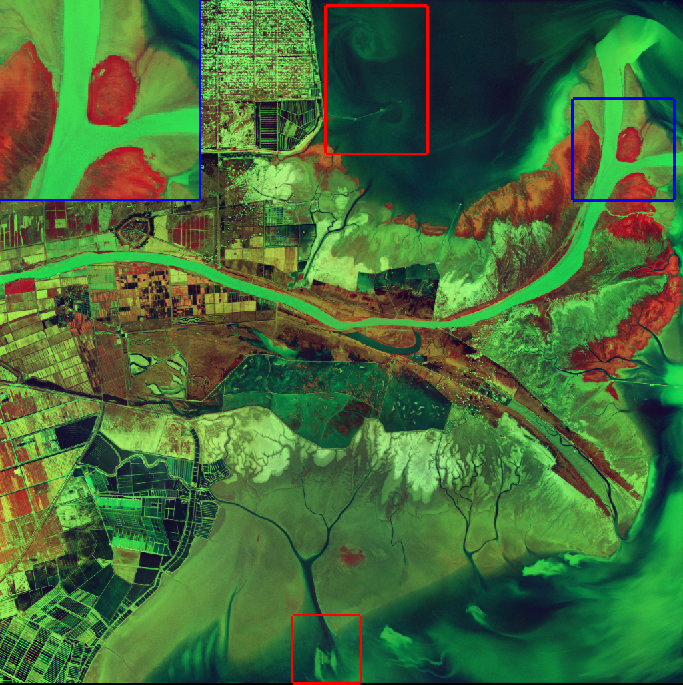}}
		
		\caption{The input  original GF1-GF5 image pair, along with the fused images (bands 30, 60, and 90) resulting from compared methodologies.\tiny\label{DF5}}
		\vspace{-3pt}
	\end{figure}
	In the figure, the red squares mark the regions of interest for detail comparison of the restored image, while the blue-highlighted sections are magnified and displayed in the upper left corner. Observation reveals that the fused image produced by NTG-Hysure exhibits pronounced color inconsistencies, aligning with observations made in preceding experiments. We conjecture that the NTG might unintentionally distort spectral information during the registration process. On the other hand, the restored image from NED appears less distinct, with more noticeable imperfections observed in the vicinity of the registration borders. DFMF achieves commendable restoration results. However, it is noticeable from the red-marked areas that the fused image is deficient in certain crucial details. It is possible that during training, the network tends to produce local detail smoothing, leading to a loss of fine detail information. Comparatively, our proposed method delivers superior restoration outcomes, preserving and recovering these important details more effectively.

	\subsection{Experimental results under aligned conditions}
	In this subsection, we concentrate on evaluating the fusion capabilities of the NLRGS model under the condition of perfectly registered HSI-MSI pairs. The numerical results of compared methods are systematically presented in Table \ref{DT4}.

    \begin{table*}[htbp]
		\renewcommand\arraystretch{1.35}
		\caption{ The PSNR,SSIM,ERGAS,SAM of compared methods on four HSI datasets.} 
		\vspace{-0.05cm}      
		\centering
		\label{DT4}               
		\begin{tabular}{ |p{1.4cm}<{\centering}|p{1.2cm}<{\centering}|p{1.2cm}<{\centering}|p{1.2cm}<{\centering}|p{1.2cm}<{\centering}|p{1.2cm}<{\centering}|p{1.3cm}<{\centering}|p{1.2cm}<{\centering}|p{1.2cm}<{\centering}|p{1.2cm}<{\centering}|}
			\hline
			\textbf{Datasets}  & \textbf{Index} & \textbf{Hysure} & \textbf{SCOTT} & \textbf{LTTR} & \textbf{LTMR} & \textbf{S$^4$-LRR} & \textbf{IRTenSR} & \textbf{ZSL}& \textbf{NLRGS}
			\\ \hline
			\multirow{3}{*}{\begin{minipage}[b]{0.5\columnwidth}\vspace{0.05cm}
					\raisebox{-0.1\width}{\includegraphics[width=1.38cm,height=1.38cm]{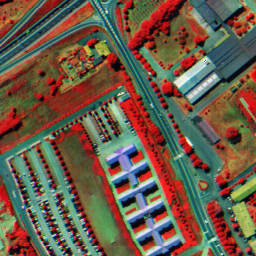}}
			\end{minipage}} & PSNR& 43.92 & 44.41 & 44.20 & 45.76 & 41.19& 44.41  & 45.54 & \textbf{47.18}
			\\ \cline{2-10} 	& SSIM& 0.992 & 0.992 & 0.988 & 0.993 & 0.986& 0.994 & 0.993  & \textbf{0.995} \\ 
			\cline{2-10} 	&ERGAS& 0.97  & 0.93  & 1.02  & 0.79  & 1.52 & 0.93 & 0.91  &  \textbf{0.66} \\ 
			\cline{2-10}  \textbf{PaviaU}	& SAM & 1.918 & 1.746 & 1.923 & 1.517 & 2.071 & 1.834 & 1.601  &  \textbf{1.299}  \\ \hline
			
			\multirow{3}{*}{\begin{minipage}[b]{0.5\columnwidth}\vspace{0.05cm}
					\raisebox{-0.5\width}{\includegraphics[width=1.38cm,height=1.38cm]{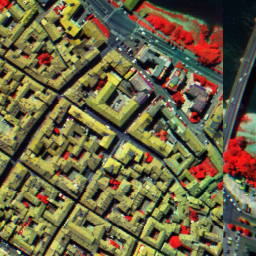}}
			\end{minipage}} & PSNR& 45.36 & 45.69 & 44.16 & 46.79 & 43.46& 44.19   & 46.30 & \textbf{48.25}
			\\ \cline{2-10} 	& SSIM& 0.994 & 0.994 & 0.989 & 0.995 & 0.993& 0.990   & 0.995 & \textbf{0.996} \\ 
			\cline{2-10} 	&ERGAS& 0.98  & 0.93  & 1.22  & 0.85  & 1.14 & 1.77   & 0.92 &  \textbf{0.72} \\ 
			\cline{2-10}  \textbf{PaviaC}	& SAM & 2.643 & 2.609 & 3.119 & 2.336 & 2.695 & 4.048  &  2.446 &  \textbf{2.019}  \\  \hline
			
			\multirow{3}{*}{\begin{minipage}[b]{0.5\columnwidth}\vspace{0.05cm}
					\raisebox{-0.3\width}{\includegraphics[width=1.38cm,height=1.38cm]{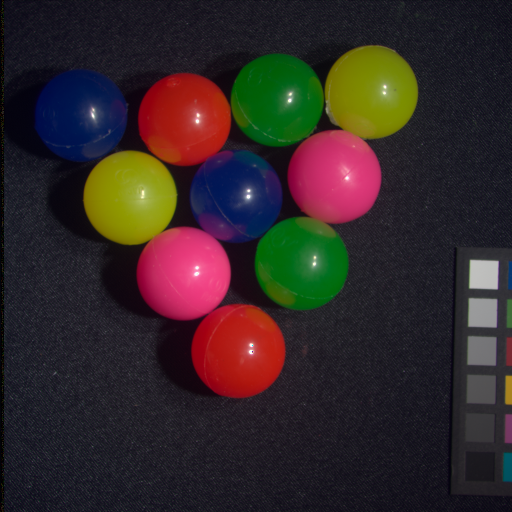}}
			\end{minipage}} & PSNR& 40.01 & 42.08 & 42.59 & 42.45 & 41.14& 43.05   & 44.52 &   \textbf{45.93}
			\\ \cline{2-10} 	& SSIM& 0.929 & 0.954 & 0.979 & 0.964 & 0.981& 0.966   &  0.982 &   \textbf{0.988} \\ 
			\cline{2-10} 	&ERGAS& 0.79  & 0.56  & 0.58 & 0.58  & 0.65 & 0.49   &  0.41 &  \textbf{0.36} \\ 
			\cline{2-10}  \textbf{Superballs}	& SAM & 28.258 & 11.895 & 8.223 & 17.663 & 8.439 & 10.262 &  7.156 &  \textbf{6.469}  \\ \hline
			
			\multirow{3}{*}{\begin{minipage}[b]{0.5\columnwidth}\vspace{0.05cm}
					\raisebox{-0.5\width}{\includegraphics[width=1.38cm,height=1.38cm]{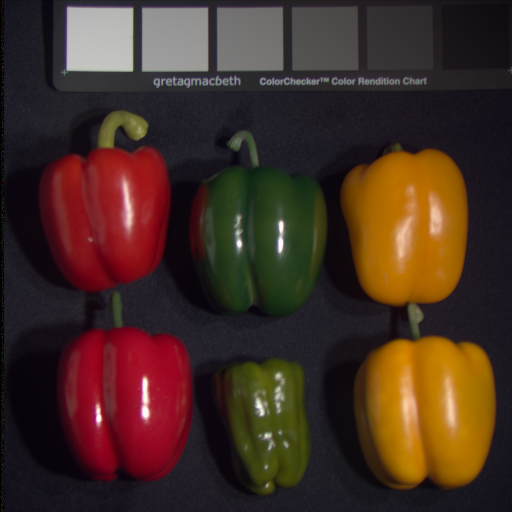}}
			\end{minipage}} & PSNR& 40.14 & 43.86 & 45.02 & 44.18 & 44.73& 44.27   & 46.21  &   \textbf{47.28}
			\\ \cline{2-10} 	& SSIM& 0.968 & 0.974 & 0.992 & 0.989 & 0.990& 0.980   & 0.992 & \textbf{0.994} \\ 
			\cline{2-10} 	&ERGAS& 0.67  & 0.73  & 0.74  & 0.74  & 0.81  & 0.76   & 0.60 &   \textbf{0.57} \\ 
			\cline{2-10}  \textbf{Peppers}	& SAM & 14.632 &10.632 & 5.131 & 6.722 & 6.145& 8.173  &  4.570 &  \textbf{4.362}  \\ \hline
		\end{tabular}
	\end{table*}
 
	From the table data, the outcomes yielded by Hysure and S$^4$-LRR are generally inferior to those of other methods. A plausible explanation behind this lies in the requirement of both algorithms to unfold the data into matrix form, a process that inadvertently disrupts the inherent spatial configuration, consequently undermining their effectiveness. Notably, our proposed NLRGS achieves a 1.4 dB higher PSNR than LTMR, and our ERGAS is significantly lower than the other compared methods. Furthermore, our reconstructed images demonstrate significantly superior performance compared to those of IR-TenSR across all evaluation metrics. This highlights that the integration of appropriate prior knowledge into the residual design is not only more efficacious but also constitutes a more rational approach compared to straightforward iterative repetitions.  Deep learning-based ZSL and other tensor factorization-based methods have all achieved remarkably effective recovery results. Nonetheless,  our approach still maintains discernible advantages  multiple evaluation metrics, which demonstrates the superior performance of NLRGS.  

\begin{figure*}[htbp]
		\centering
		\subfloat{\includegraphics[width=1.81cm]{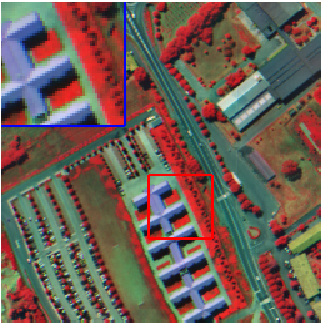}}\!
		\subfloat{\includegraphics[width=1.81cm]{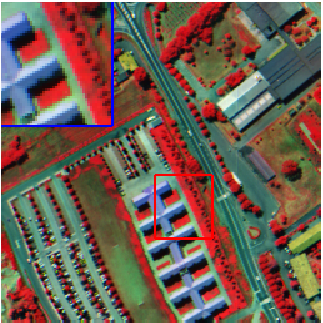}}\!
		\subfloat{\includegraphics[width=1.81cm]{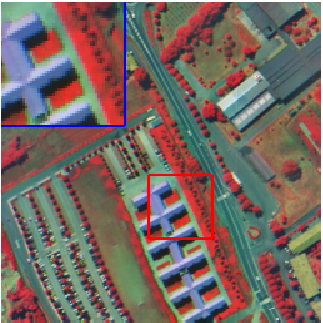}}\!
		\subfloat{\includegraphics[width=1.81cm]{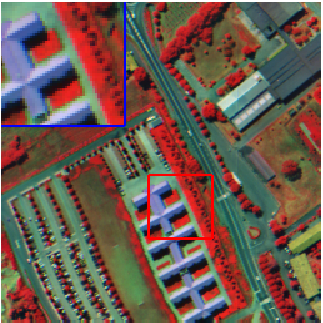}}\!
		\subfloat{\includegraphics[width=1.81cm]{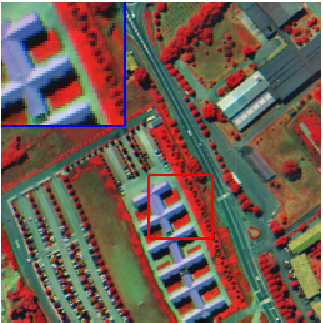}}\!
		\subfloat{\includegraphics[width=1.81cm]{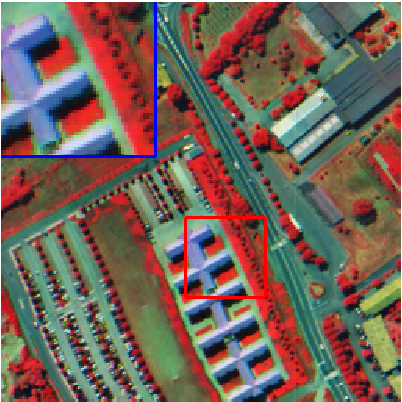}}\!
		\subfloat{\includegraphics[width=1.81cm]{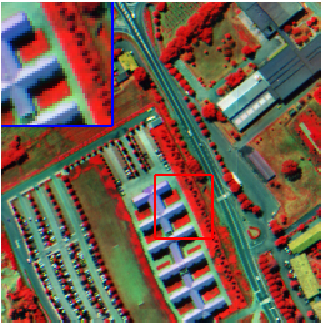}}\!
		\subfloat{\includegraphics[width=1.81cm]{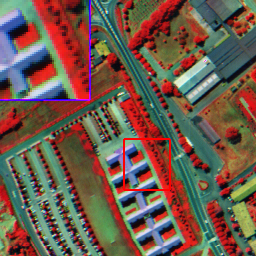}}\!
		\subfloat{\includegraphics[width=1.81cm]{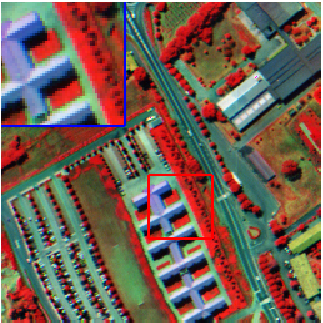}}
		\vspace{-8pt}

		\subfloat{\includegraphics[width=1.81cm]{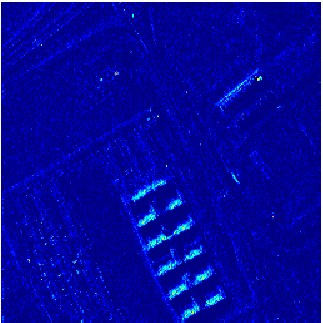}}\!
		\subfloat{\includegraphics[width=1.81cm]{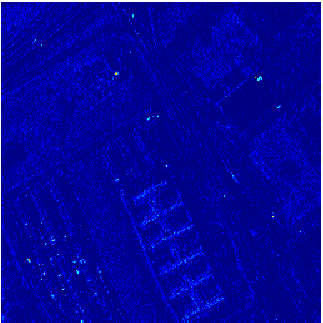}}\!
		\subfloat{\includegraphics[width=1.81cm]{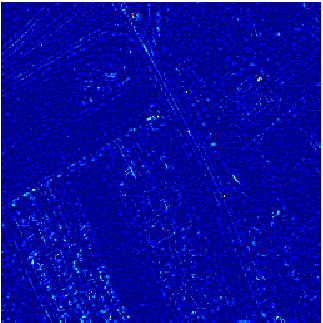}}\!
		\subfloat{\includegraphics[width=1.81cm]{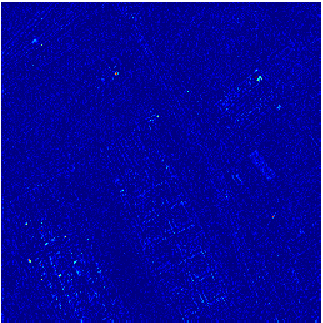}}\!
		\subfloat{\includegraphics[width=1.81cm]{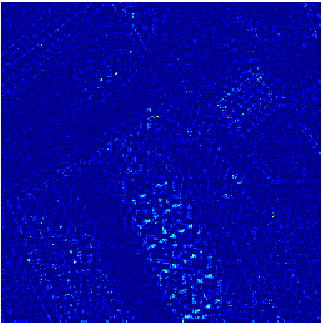}}\!
		\subfloat{\includegraphics[width=1.81cm]{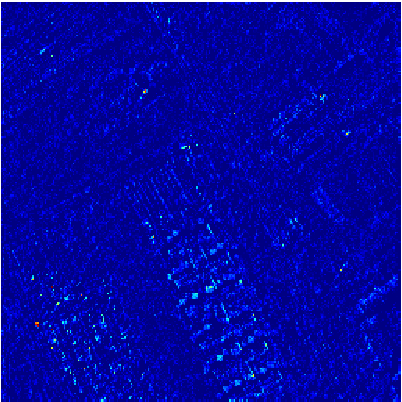}}\!
		\subfloat{\includegraphics[width=1.81cm]{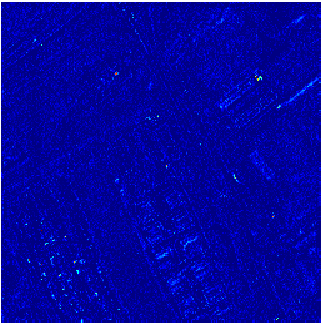}}\!
		\subfloat{\includegraphics[width=1.81cm]{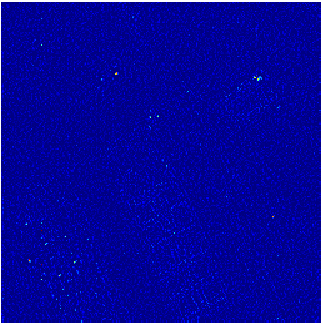}}\!
		\subfloat{\includegraphics[width=1.81cm]{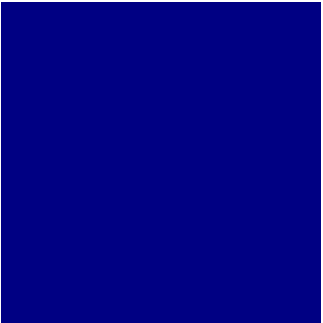}}

		\subfloat{\includegraphics[width=1.81cm]{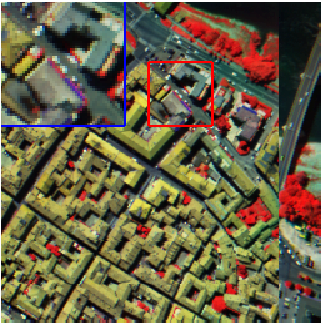}}\!
		\subfloat{\includegraphics[width=1.81cm]{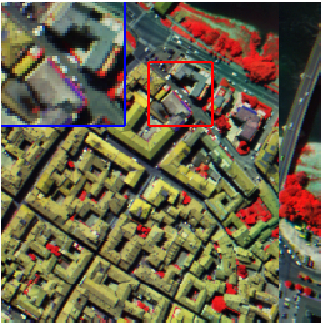}}\!
		\subfloat{\includegraphics[width=1.81cm]{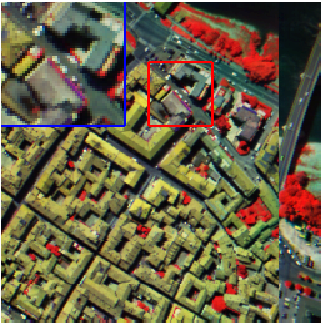}}\!
		\subfloat{\includegraphics[width=1.81cm]{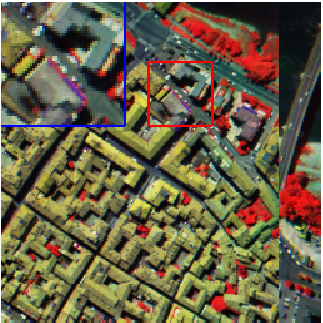}}\!
		\subfloat{\includegraphics[width=1.81cm]{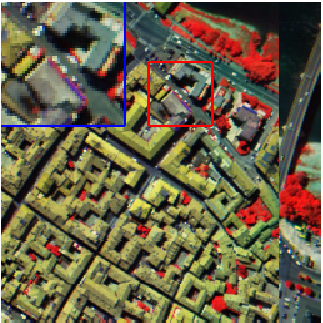}}\!
		\subfloat{\includegraphics[width=1.81cm]{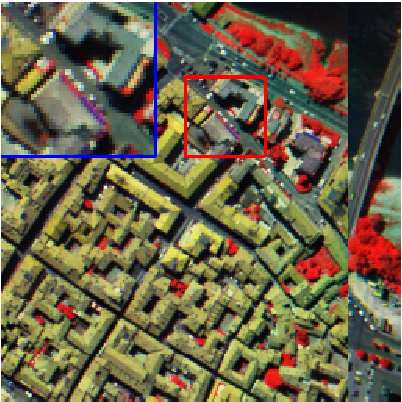}}\!
		\subfloat{\includegraphics[width=1.81cm]{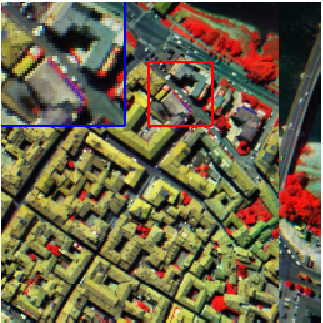}}\!
		\subfloat{\includegraphics[width=1.81cm]{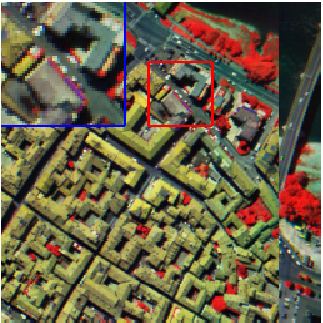}}\!
		\subfloat{\includegraphics[width=1.81cm]{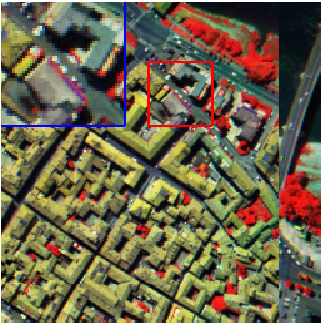}}
		
		\captionsetup[subfloat]{labelsep=none,format=plain,labelformat=empty}
        \vspace{-8pt}
		\subfloat[Hysure]{\includegraphics[width=1.81cm]{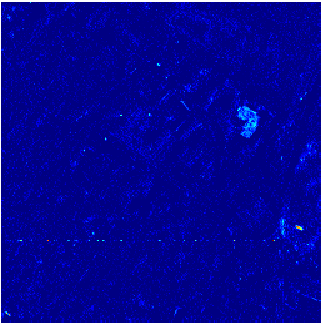}}\!
		\subfloat[SCOTT]{\includegraphics[width=1.81cm]{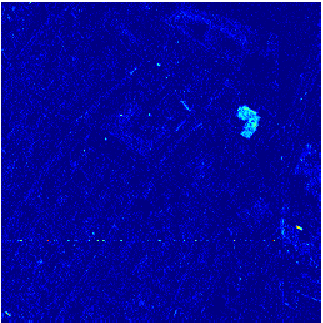}}\!
		\subfloat[LTTR]{\includegraphics[width=1.81cm]{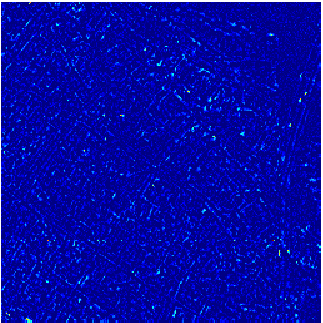}}\!
		\subfloat[LTMR]{\includegraphics[width=1.81cm]{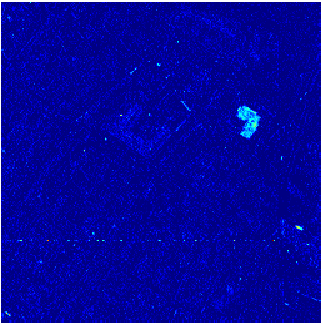}}\!
		\subfloat[S$^4$-LRR]{\includegraphics[width=1.81cm]{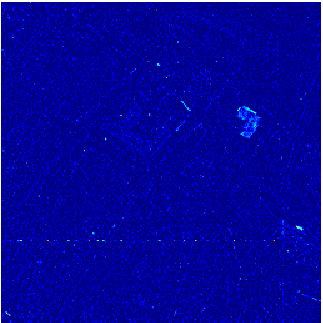}}\!
		\subfloat[IR-TenSR]{\includegraphics[width=1.81cm]{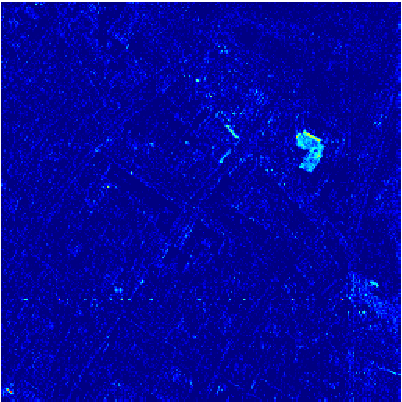}}\!
		\subfloat[ZSL]{\includegraphics[width=1.81cm]{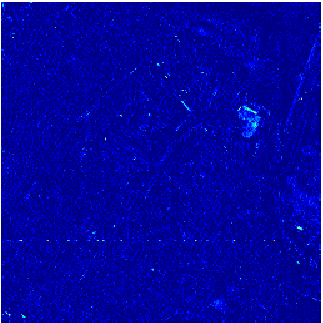}}\!
		\subfloat[NLRGS]{\includegraphics[width=1.81cm]{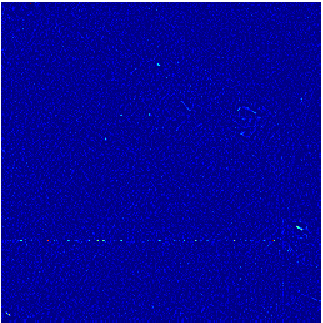}}\!
		\subfloat[GT]{\includegraphics[width=1.81cm]{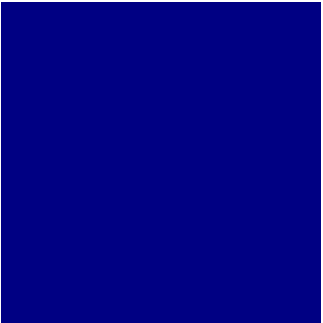}}

	\includegraphics[width=10cm]{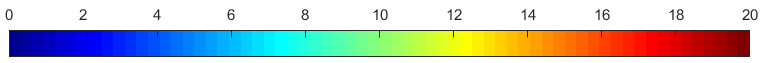}
		\caption{The reconstructed images and corresponding error images of 'Pavia University' and 'Pavia Center' (bands 20, 40, 60) of all the compared methods.\tiny\label{DF2}}
	\end{figure*}

 \begin{figure*}[htbp]
		\centering
		\subfloat{\includegraphics[width=1.81cm]{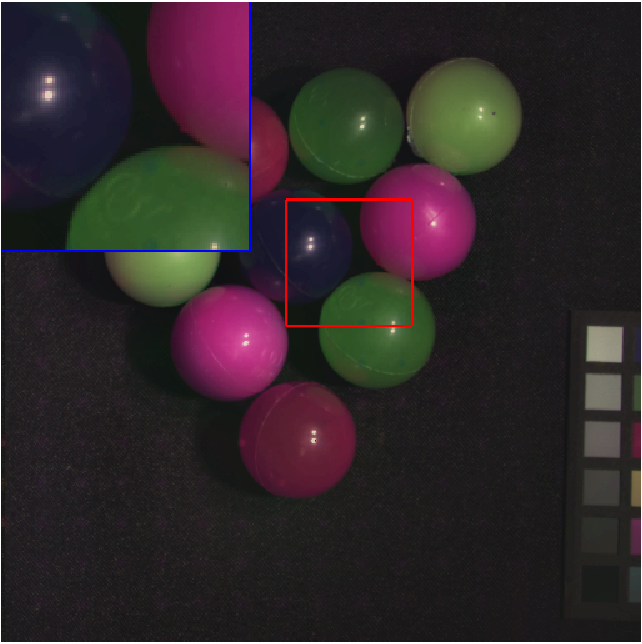}}\!
		\subfloat{\includegraphics[width=1.81cm]{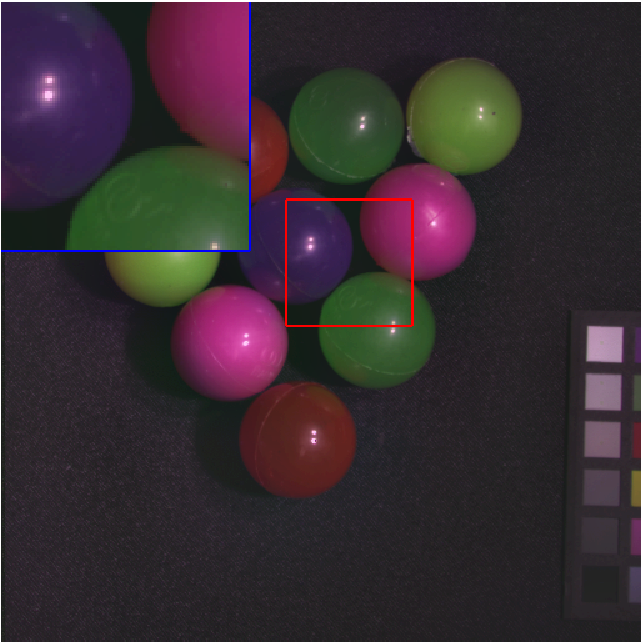}}\!
		\subfloat{\includegraphics[width=1.81cm]{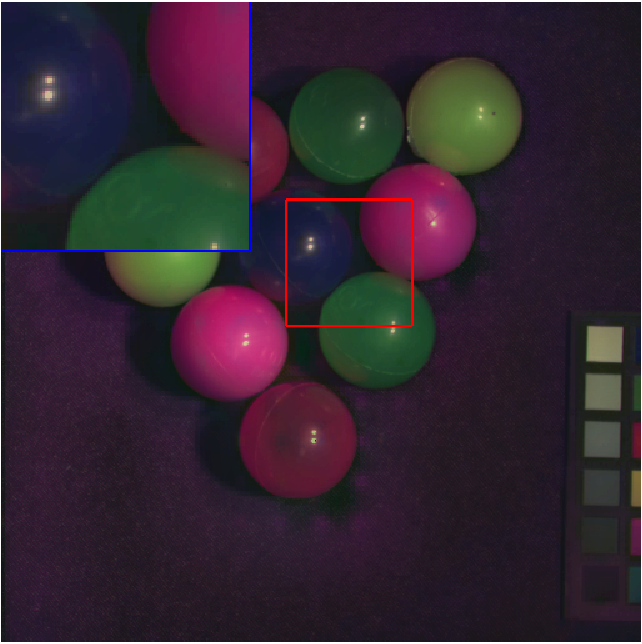}}\!
		\subfloat{\includegraphics[width=1.81cm]{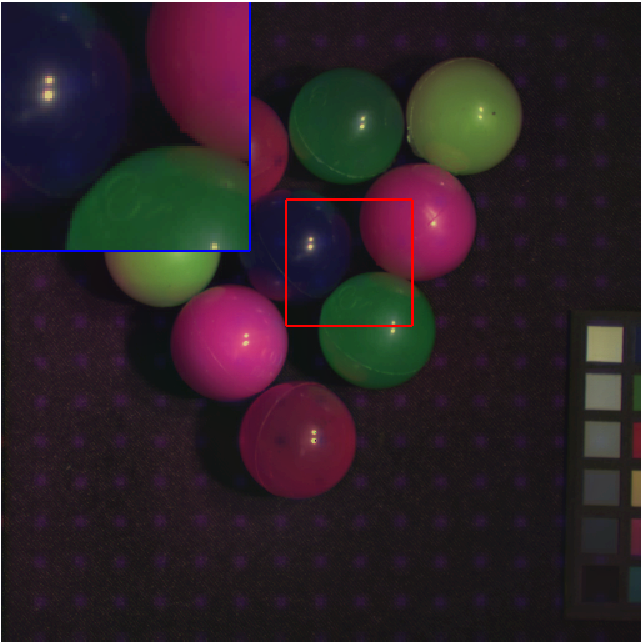}}\!
		\subfloat{\includegraphics[width=1.81cm]{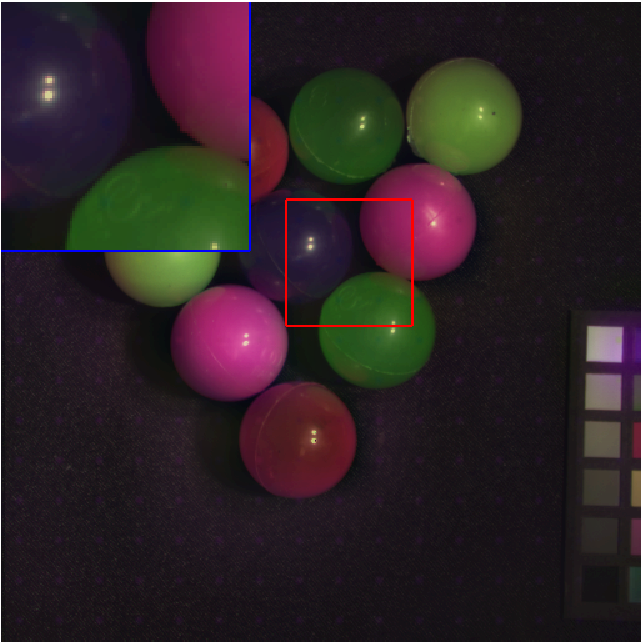}}\!
		\subfloat{\includegraphics[width=1.81cm]{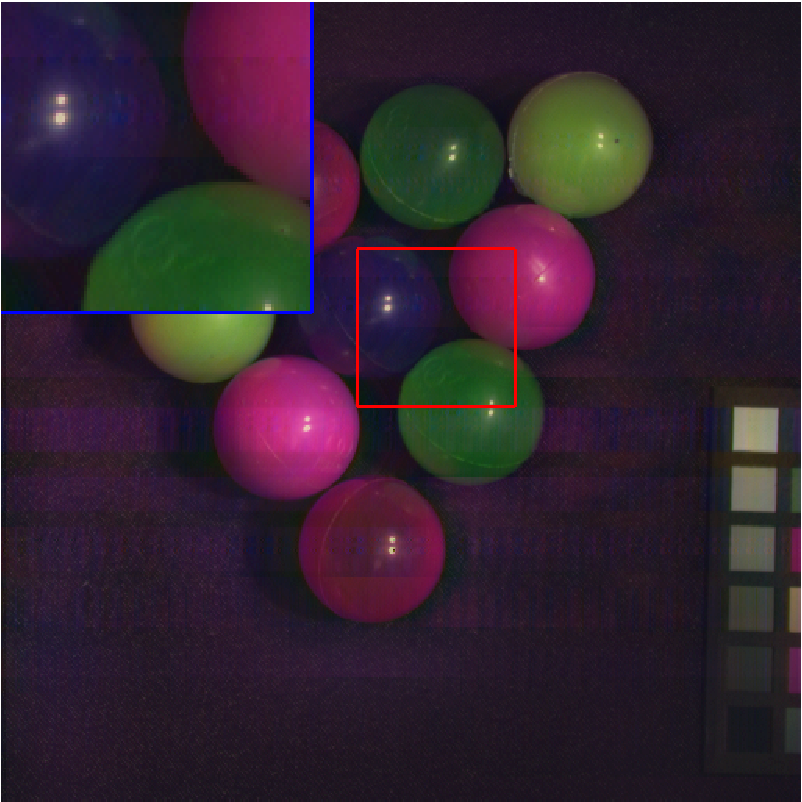}}\!
		\subfloat{\includegraphics[width=1.81cm]{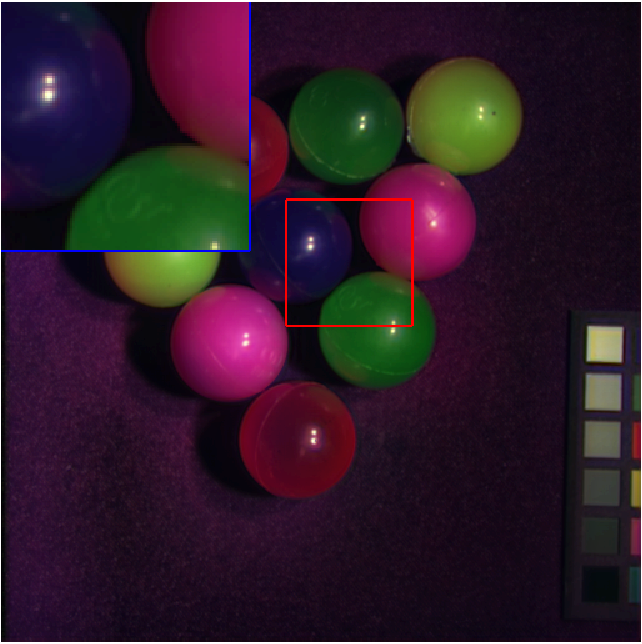}}\!
		\subfloat{\includegraphics[width=1.81cm]{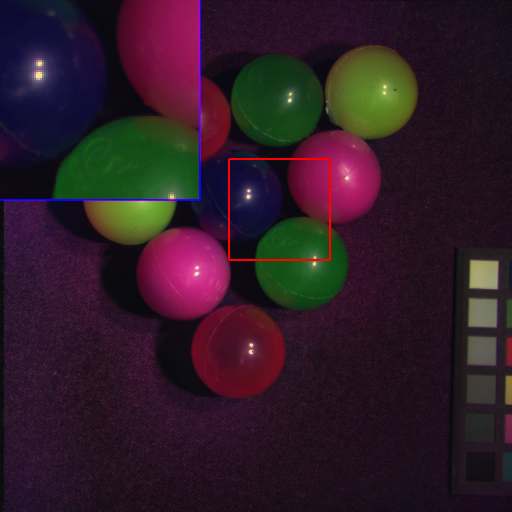}}\!
		\subfloat{\includegraphics[width=1.81cm]{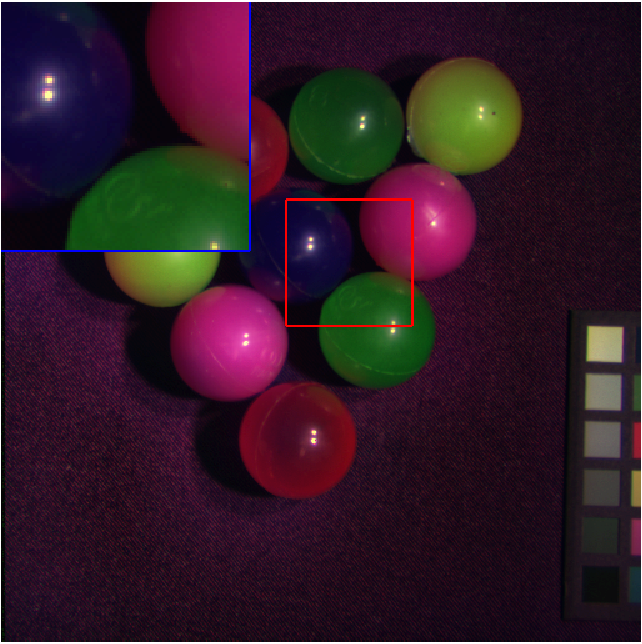}}
		
		\vspace{-8pt}
		\subfloat{\includegraphics[width=1.81cm]{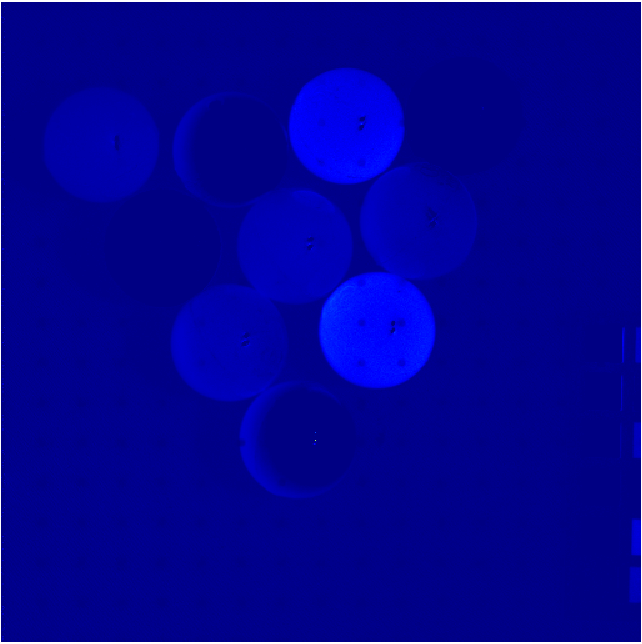}}\!
		\subfloat{\includegraphics[width=1.81cm]{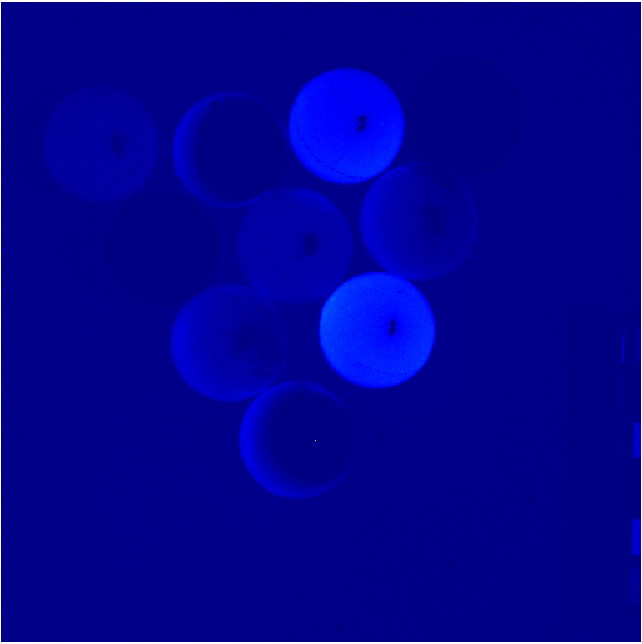}}\!
		\subfloat{\includegraphics[width=1.81cm]{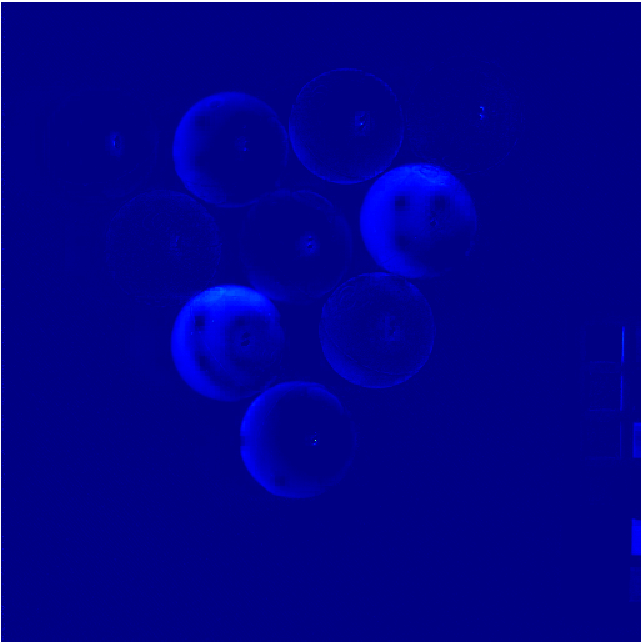}}\!
		\subfloat{\includegraphics[width=1.81cm]{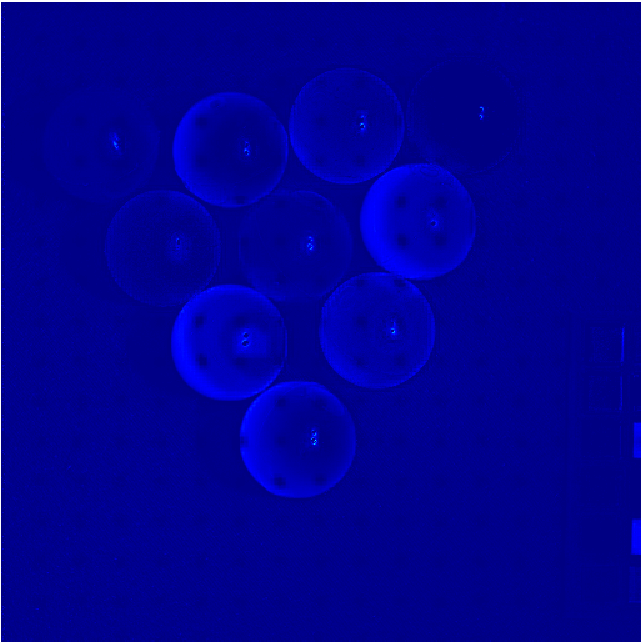}}\!
		\subfloat{\includegraphics[width=1.81cm]{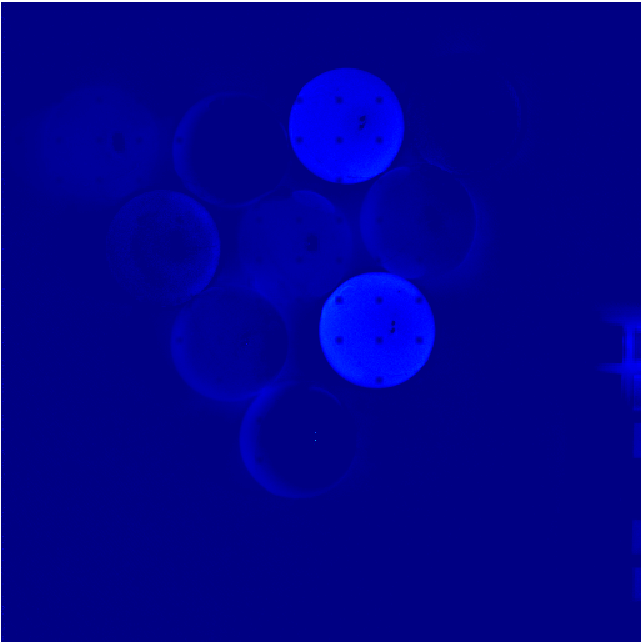}}\!
		\subfloat{\includegraphics[width=1.81cm]{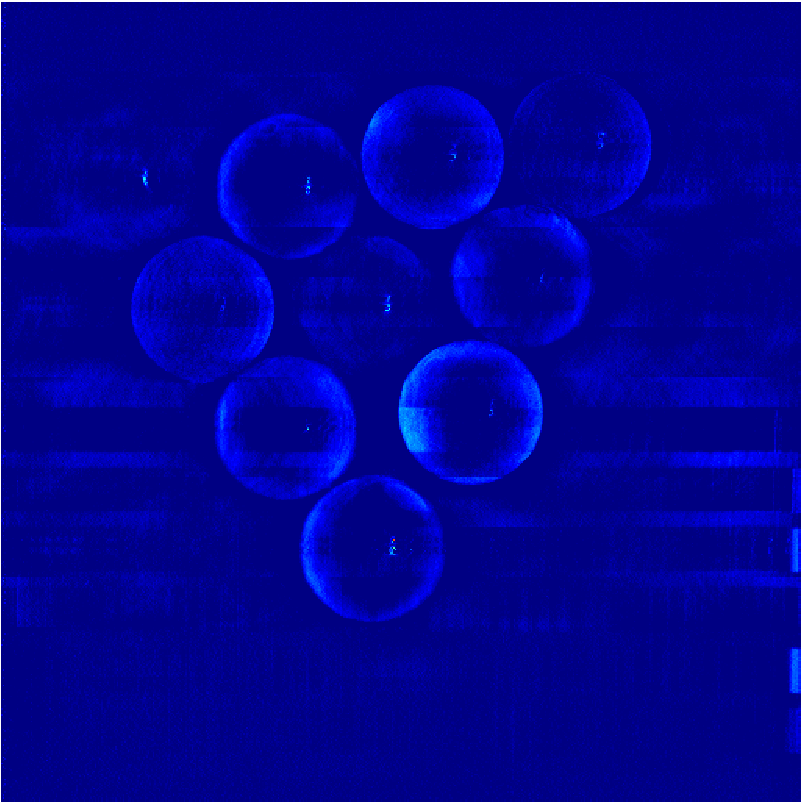}}\!
		\subfloat{\includegraphics[width=1.81cm]{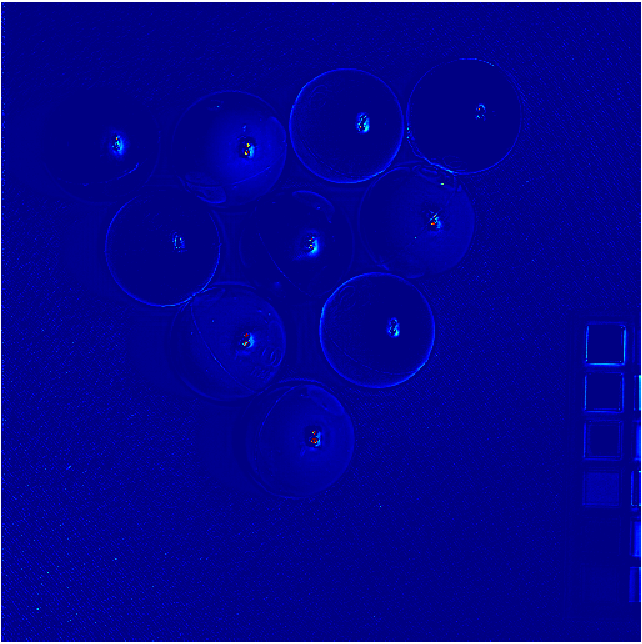}}\!
		\subfloat{\includegraphics[width=1.81cm]{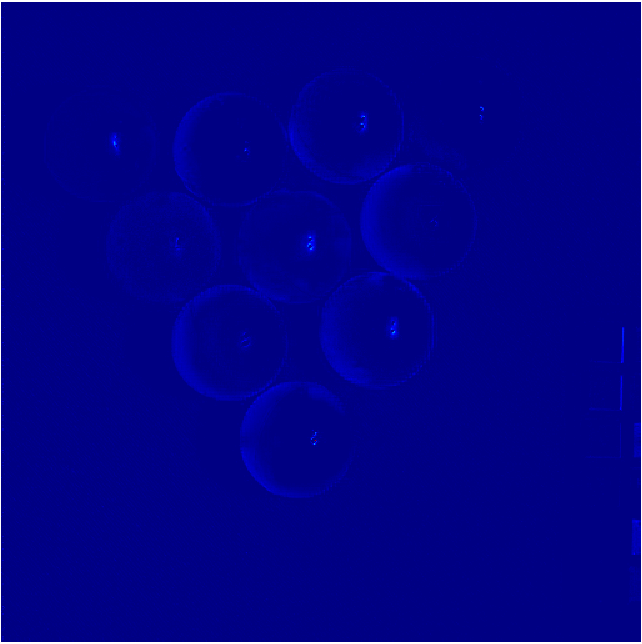}}\!
		\subfloat{\includegraphics[width=1.81cm]{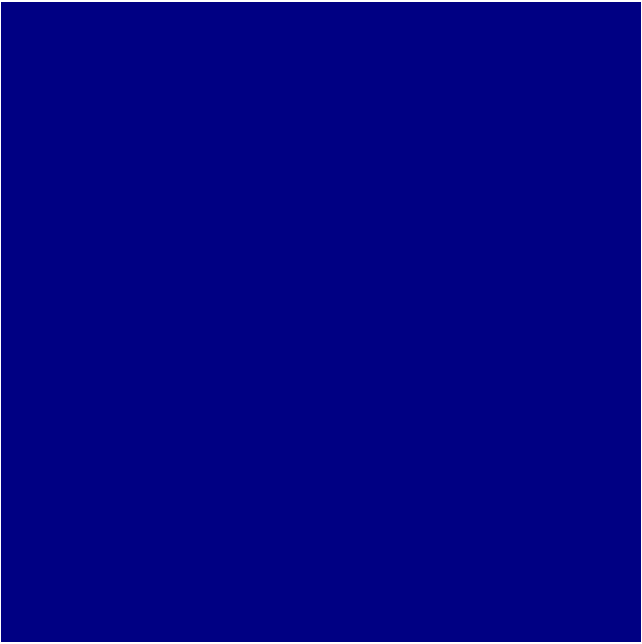}}

		\subfloat{\includegraphics[width=1.81cm]{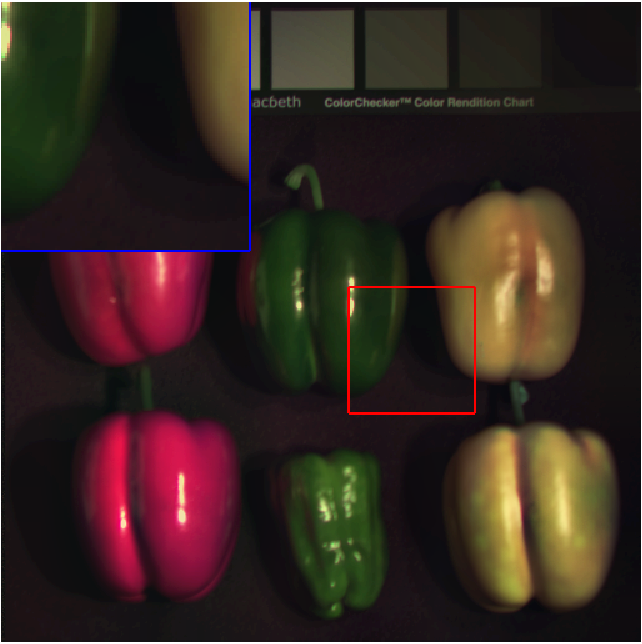}}\!
		\subfloat{\includegraphics[width=1.81cm]{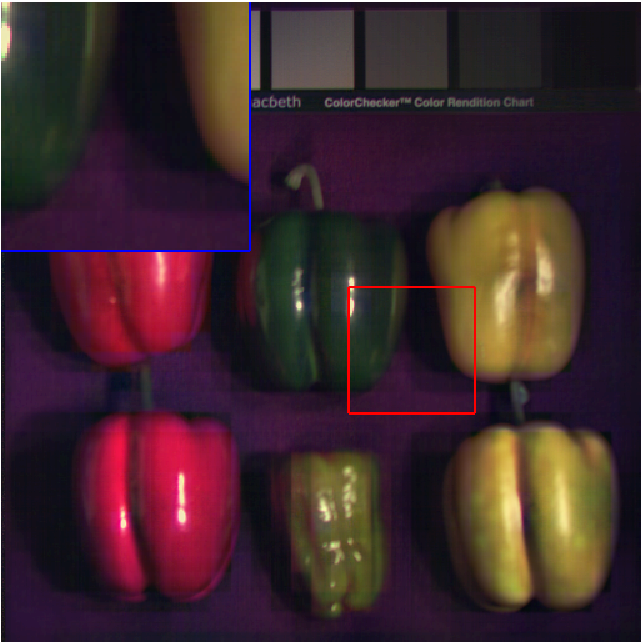}}\!
		\subfloat{\includegraphics[width=1.81cm]{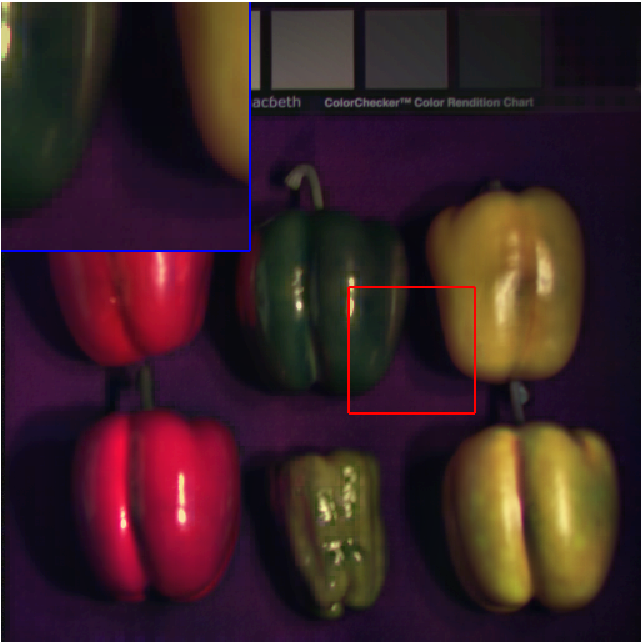}}\!
		\subfloat{\includegraphics[width=1.81cm]{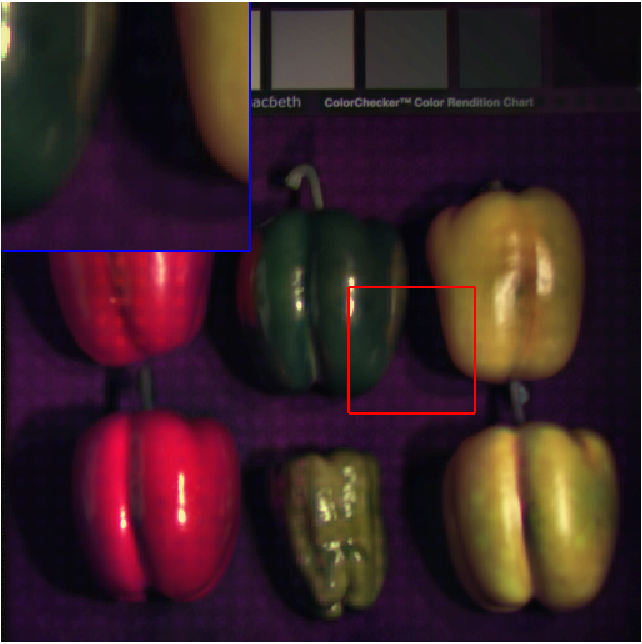}}\!
		\subfloat{\includegraphics[width=1.81cm]{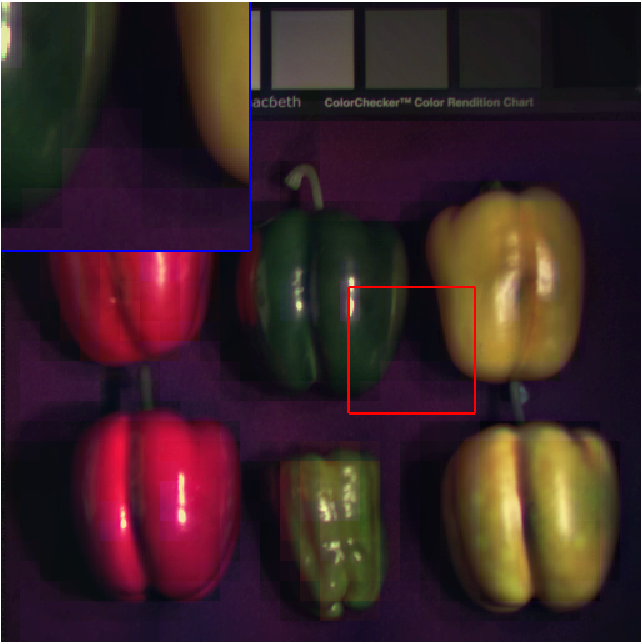}}\!
		\subfloat{\includegraphics[width=1.81cm]{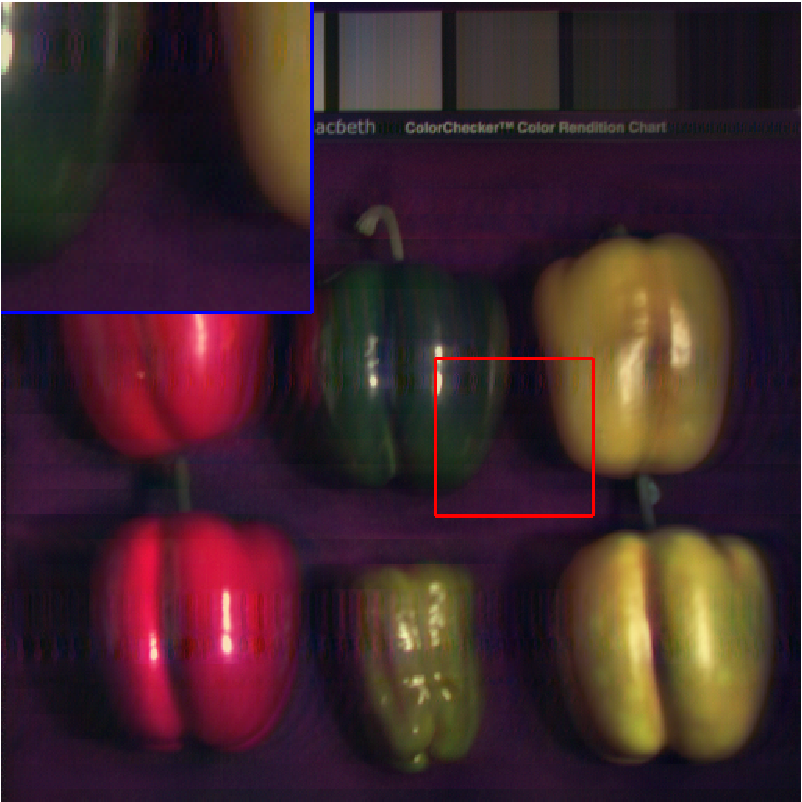}}\!
		\subfloat{\includegraphics[width=1.81cm]{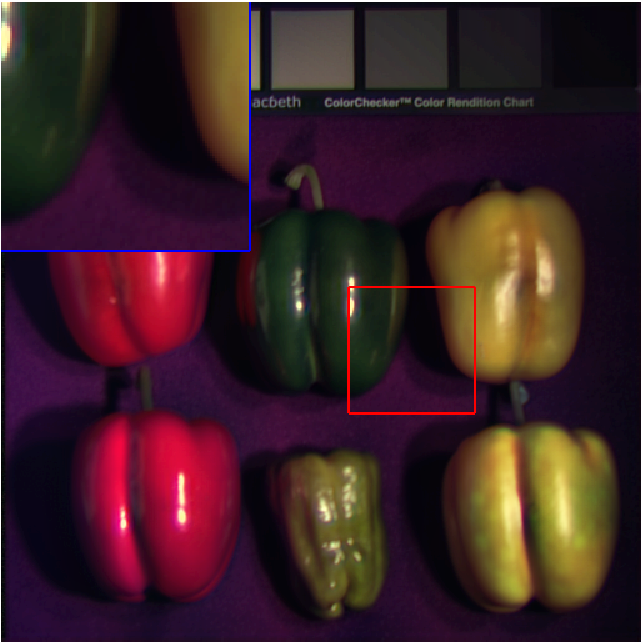}}\!
		\subfloat{\includegraphics[width=1.81cm]{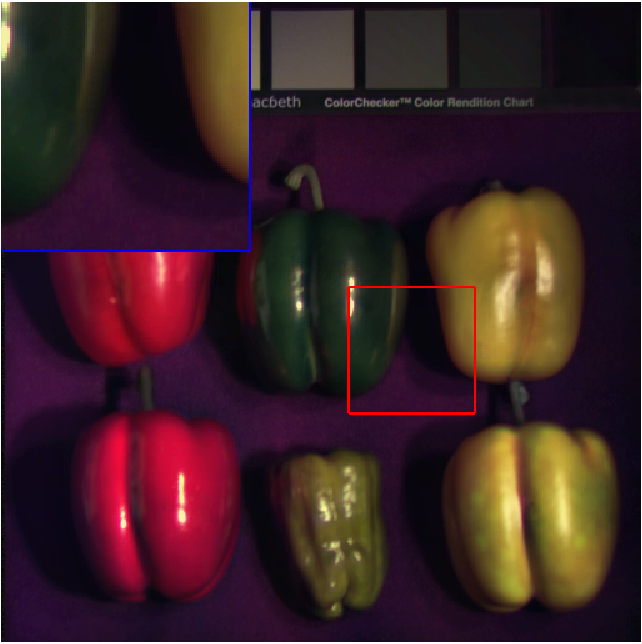}}\!
		\subfloat{\includegraphics[width=1.81cm]{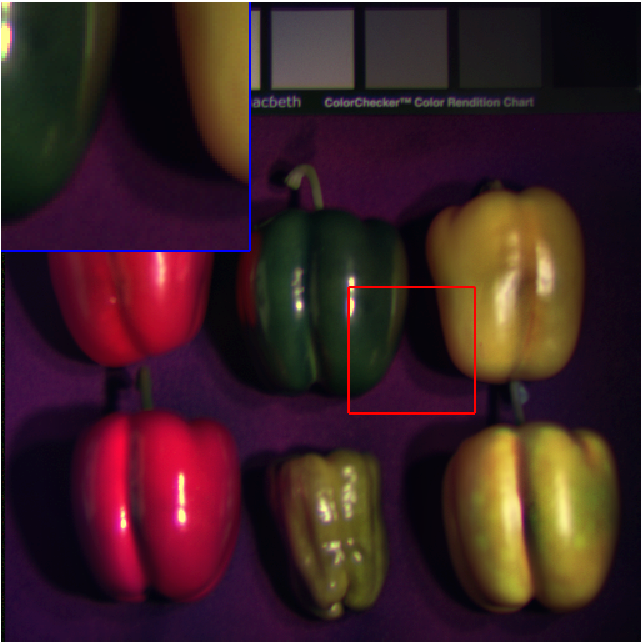}}
  	
  	\captionsetup[subfloat]{labelsep=none,format=plain,labelformat=empty}
	\vspace{-8pt}
		\subfloat[HySure]{\includegraphics[width=1.81cm]{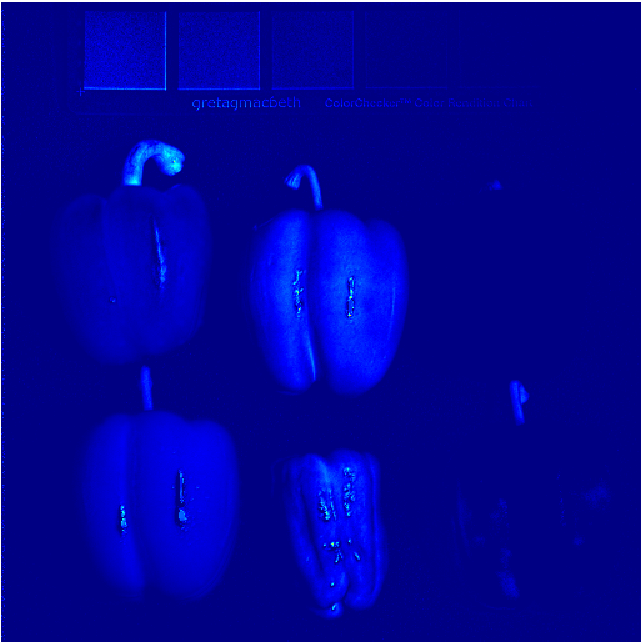}}\!
		\subfloat[SCOTT]{\includegraphics[width=1.81cm]{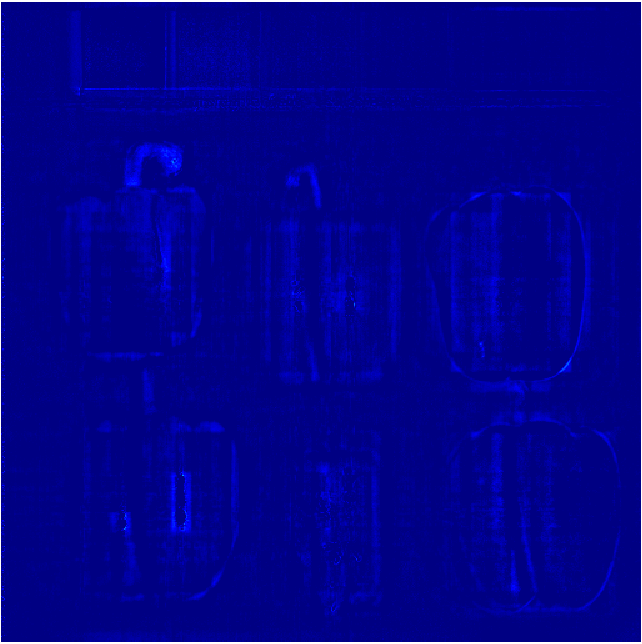}}\!
		\subfloat[LTTR]{\includegraphics[width=1.81cm]{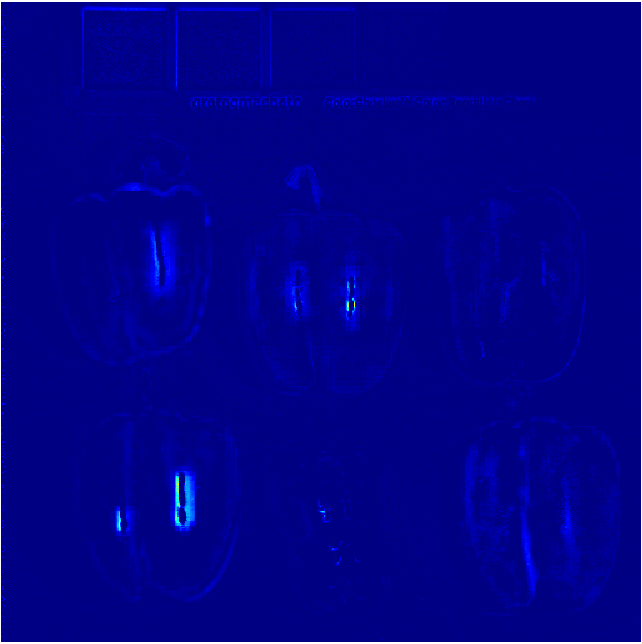}}\!
		\subfloat[LTMR]{\includegraphics[width=1.81cm]{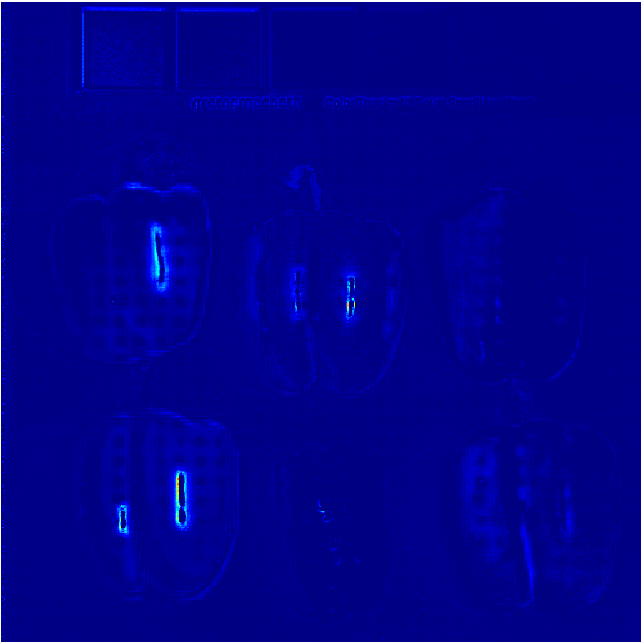}}\!
		\subfloat[S$^4$-LRR]{\includegraphics[width=1.81cm]{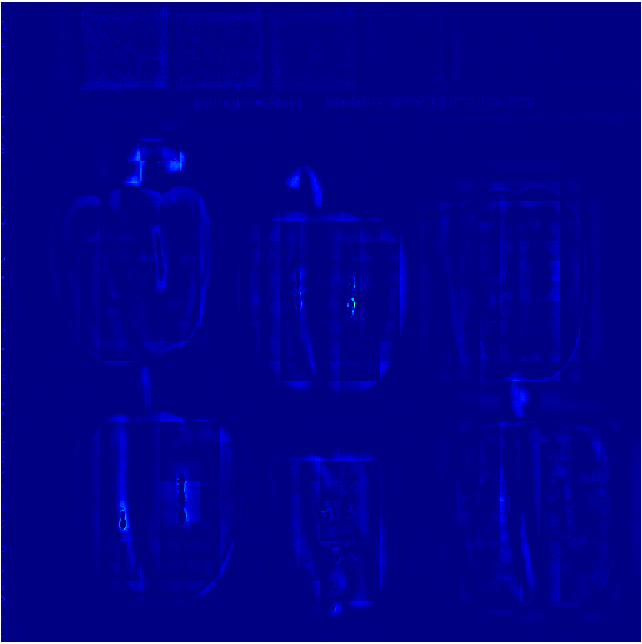}}\!
		\subfloat[IR-TenSR]{\includegraphics[width=1.81cm]{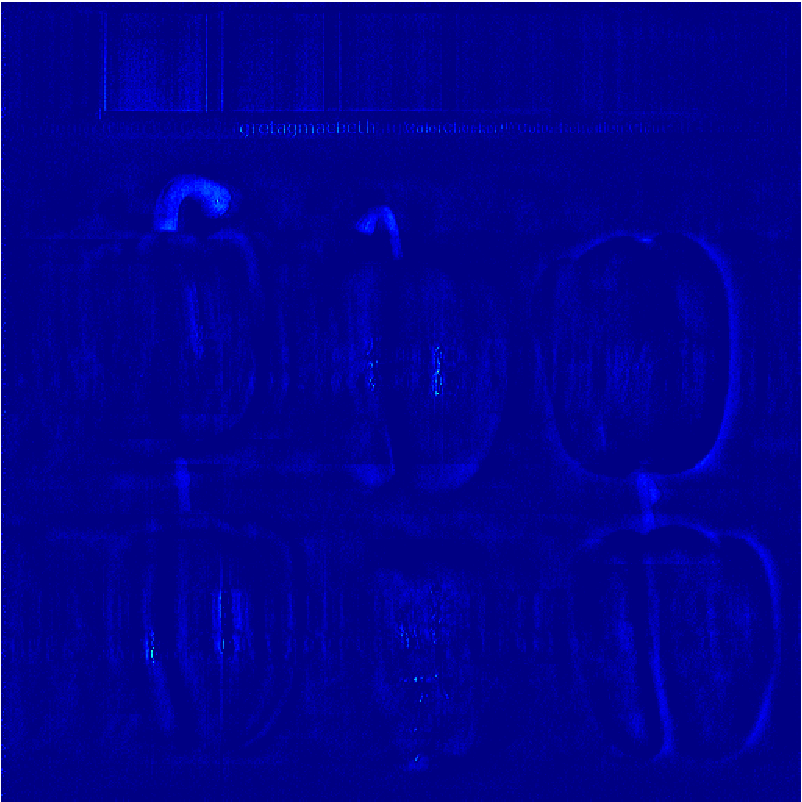}}\!
		\subfloat[ZSL]{\includegraphics[width=1.81cm]{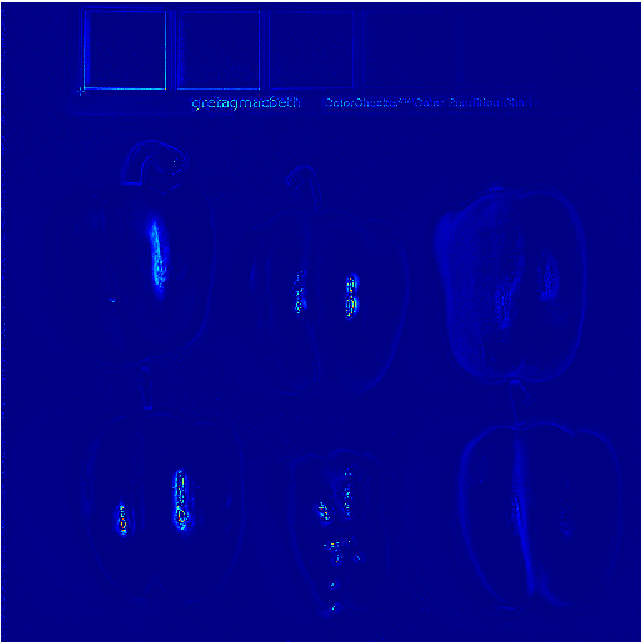}}\!
		\subfloat[NLRGS]{\includegraphics[width=1.81cm]{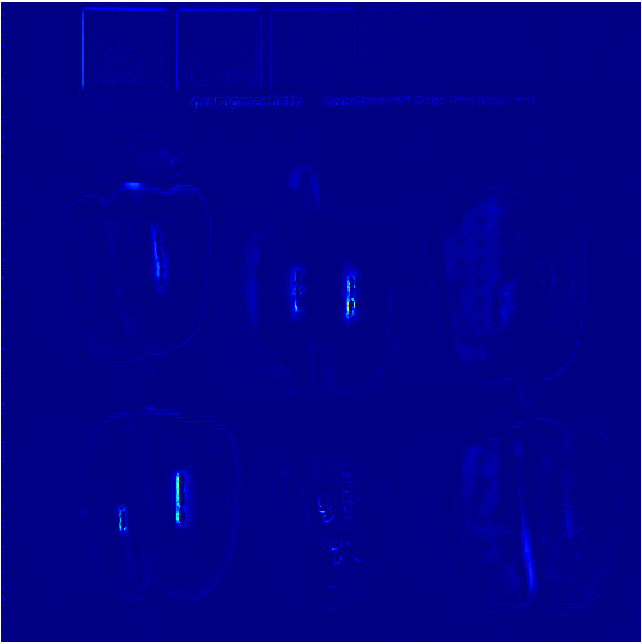}}\!
		\subfloat[GT]{\includegraphics[width=1.81cm]{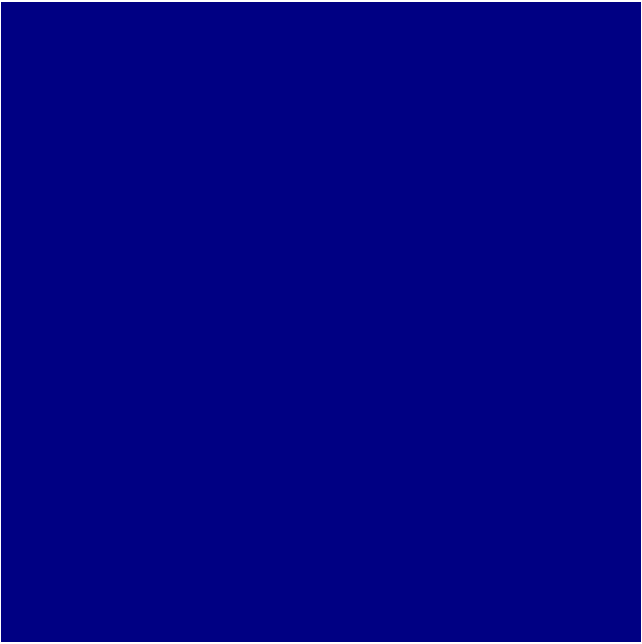}}
		
		\includegraphics[width=10cm]{overleaf_RAF/band_cave.png}
		\caption{The reconstructed images and corresponding error images of 'Superballs' and 'Peppers' (bands 10, 20, 30) of all the compared methods.\tiny\label{DF22}}
		\vspace{-5pt}
	\end{figure*}
 
 Figure \ref{DF2} and \ref{DF22} visually illustrate the reconstructed images obtained from all comparative methods, wherein the red block region within each image is enlarged and situated at the top-left corner for detailed examination. Additionally, the figure encompasses error maps corresponding to the reconstructed HSIs, offering a comprehensive insight into the discrepancies. Hysure employs vector total variation, a choice which, albeit beneficial for certain aspects, inadvertently accentuates defects along the edges of the restored images, making them more conspicuous.  SCOTT has shown superior performance on various datasets, which is attributed to the Tucker decomposition's ability to fully utilize the spatial-spectral correlation of the images. ZSL demonstrates strong competitiveness across various datasets, owing to the powerful feature extraction capabilities of neural networks. Both LTTR and LTMR incorporate nonlocal clustering together with low-rank approximation to achieve high quality reconstruction of the principal content. However, due to the neglect of residual information, their error maps manifest conspicuous deficiencies. Our methodology stands out by yielding notably fewer artifacts in the restored images compared to rival approaches, a feat chiefly accomplished through our strategy of further harnessing the residual information to extract and utilize crucial, previously untapped details effectively.
	
	Subsequently,  to verify the effectiveness of group sparsity, comparative experiments were carried out with the NLRGS model across diverse datasets. Specifically, we juxtaposed scenarios where group sparsity priors were  integrated against instances lack of such priors.
	The experimental results are presented in table \ref{table3}.

	\begin{table}[htbp]\small
		\renewcommand\arraystretch{1.31}
		\begin{center}
			\caption{Contrast experiment of NLRGS to explore the effectiveness of group sparsity prior. \label{table3}}
			\begin{tabular}{@{}*{6}{l}}
				\hline
				Dataset&$\Vert \cdot \Vert_{l_F^\psi}$ & PSNR   &SSIM &ERGAS &SAM  \\
				\hline
				\multirow{2}{*}{Pavia U}  & \checkmark & 47.18  & 0.995 & 0.664 & 1.299   \\
				\cline{2-6}
				& $\times$  & 46.95 &  0.994  & 0.683 & 1.307 \\		
				\hline
				\multirow{2}{*}{Pavia C}  & \checkmark  & 48.25  & 0.996 & 0.718 & 2.019  \\
				\cline{2-6}
				&$\times$ & 47.95  & 0.996 &  0.727  & 2.019  \\	
				\hline
				\multirow{2}{*}{Superballs} &\checkmark  & 45.93  & 0.988 &  0.365  &  6.469  \\			
				\cline{2-6}
				&$\times$ & 43.89  & 0.982 &  0.442 &  7.853  \\			
				\hline
				\multirow{2}{*}{Peppers} & \checkmark  & 47.28  & 0.994 &  0.561  &  4.362  \\
				\cline{2-6}
				&$\times$ & 45.77  & 0.989 & 0.657 &  6.198  \\			
				\hline
			\end{tabular}
		\end{center}
	\end{table}

	From the table, we can clearly discern the impact of incorporating a group sparsity prior. With respect to the two Pavia datasets, while the group sparsity prior does show some effectiveness, its enhancement is not dramatically pronounced. Nonetheless, when examining the CAVE dataset, it strikingly emerges that the adoption of the group sparsity prior yields a performance enhancement of remarkable significance. For instance, when considering the `Superballs' dataset, the implementation of group sparsity results in a notable improvement of 2dB in PSNR compared to scenarios where it is not utilized. This enhancement is corroborated by consistent improvements across other evaluation metrics as well. This convincingly demonstrates the effectiveness of group sparsity priors in extracting residual information.

	\subsection{Convergence  Behavior}
\begin{figure}[htbp]
		\centering
		\subfloat[$\kappa_{raf}$]{\includegraphics[width=5.7cm,height=4.5cm]{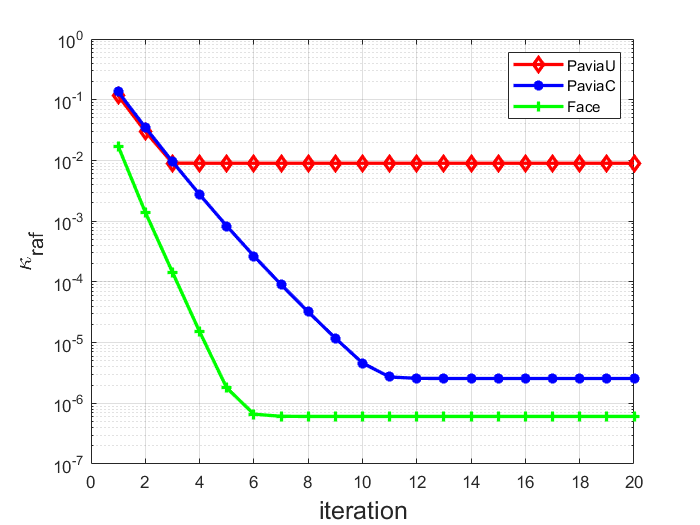}}
        \quad
		\subfloat[$\kappa_{nlrgs}$]{\includegraphics[width=5.7cm,height=4.5cm]{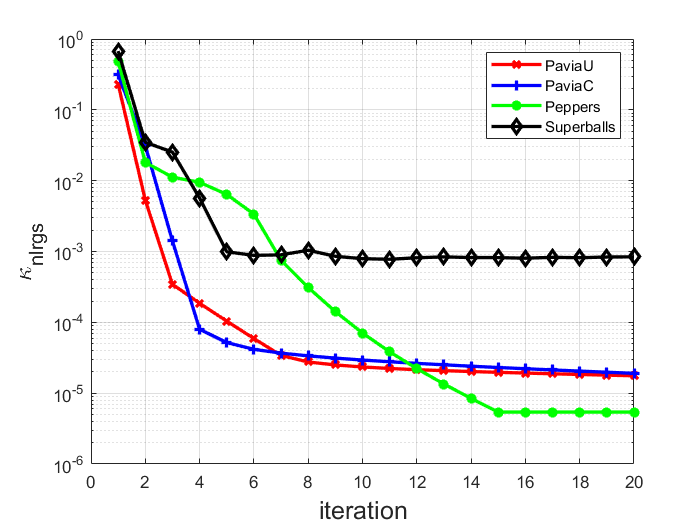}}
		\caption{The termination indicators with respect to the iteration numbers for different HSI datasets.\tiny\label{DF4}}
	\end{figure}

	In this subsection, the numerical convergence is meticulously examined across varying HSI datasets, with a particular emphasis placed on assessing the outer loop convergence. Figure \ref{DF4} offers a rigorous empirical examination of the convergence trends displayed by both the RAF and NLRGS models, offering a deep insight into their convergence behaviors.
	
	As depicted, the termination indicators $\kappa_{raf}$ and $\kappa_{nlrgs}$ demonstrate a consistent decline with increasing iteration counts, ultimately attaining a stable equilibrium point. 
	This observation further substantiates the effectiveness of both our algorithm and the termination criterion employed.

	\section{CONclusion}\label{G}
	In this paper, we introduce batch image alignment into the HSI-MSI fusion task, proposing a constrained optimization model capable of simultaneously performing image registration and fusion. Building upon this foundation,  we develop the RAF-NLRGS framework, which achieves more stable and high-quality registration and fusion performance. The RAF model handles both image registration and fusion concurrently, thoughtfully accommodating spectral dimensional disparities inherent between multi-source remote sensing images. The NLRGS model, nested within RAF, further elevates the quality of fused images. Moreover, we establish error bounds for the NLRGS model. Subsequently, we employ the GGN and PAO algorithms to solve the model, complemented by a thorough convergence analysis. Extensive numerical experiments on HSI datasets are conducted to verify the effectiveness of our proposed method.
	
	In our future work, we aspire to develop even more efficient methodologies  to tackle the challenging scenario characterized by non-aligned images and an uncertain degradation operator.

\bibliographystyle{unsrt}  


\bibliography{CiteTex}

\newpage

\textbf{\Large Appendix}

In the appendix, we will provide supplementary material to some of the content in the text, including proofs of certain theorems and additional experiments.

\noindent\textbf{\large A. Proximity Mapping of the Functions }
	
	\begin{itemize}
		%
		\item $l_1:$ \textbf{Soft-thresholding}:
		
		\qquad\qquad$	prox_{\alpha\psi}(x)=\text{sign}(x)\cdot \text{max}\{|x|-\alpha,0\}$.
		
		\item $logarithm:$
		
		\qquad\qquad$	prox_{\alpha\psi}(x)=\left\{
		\begin{array}{rcl}
			0,       &      & b\le 0,\\
			\dfrac{a+\sqrt{b}}{2},   &      & b>0,
		\end{array} \right.$
		
		where $a=|x|-\theta$ and $b=a^2-4(\alpha-\theta|x|)$.
		
		\item $\text{MCP}:$
		
		$	prox_{\alpha\psi}(x)=\left\{
		\begin{array}{rcl}
			0,       &      & |x|\le \alpha,\\
			\dfrac{\text{sign}(x)(|x|-\alpha)}{1-1/\theta},   &      & \alpha<|x|\le\theta\alpha; \\
			x,       &      & |x|> \theta\alpha.
		\end{array} \right.$
	\end{itemize}
	
	\vspace{10pt}
	\noindent\textbf{\large B. The Proof of Theorem 4.1 }
	
	\textbf{Proof:} We proceed by leveraging the following conclusion to prove the theorem.
	Jittorntrum \textit{et al.} \cite{JK1} considered to solve the following optimal problem:
	\begin{equation}\tag{1.1}\label{P1}
		\min\limits_\textbf{x} \Vert f(\textbf{x}) \Vert_{\diamond},
	\end{equation}
	where $\Vert \cdot \Vert_{\diamond}$ denotes a norm and $f$ is a twice continuously differentiable mapping. Then the generalized Gauss–Newton method is given in the following iteration:
	
	\noindent(i) Compute $\textbf{t}_i$ at $\textbf{x}_i$ by solving the subproblem: $$\min\limits_{\textbf{t}_i}\Vert f(\textbf{x}_i)+\nabla f(\textbf{x}_i)\textbf{t}_i \Vert_{\diamond}$$
	
	\noindent(ii) Set $\textbf{x}_{i+1}=\textbf{x}_i+\textbf{t}_i$.
	
	\noindent Then the sequence $\{\textbf{x}_i\}_{i\in\mathbb{N}}$ converges quadratically in the neighborhood of any strongly unique local minimum of (\ref{P1}).
	
	To demonstrate the convergence of our algorithm, it suffices to  define  
	\begin{equation*}
		\begin{aligned}
			\Vert \mathcal{M} \Vert_{\diamond}:=&\min\limits_{\mathcal{L},\textbf{A},\mathcal{B}}\enspace \Vert \mathcal{L} \Vert_{\text{TNN}}\\
			&s.t. \quad \mathcal{H}_{spa}(\mathcal{L}\times_3\textbf{D})=\mathcal{M},  \\& \qquad\mathcal{H}_{spec} (\mathcal{L}\times_3\textbf{D})=\mathcal{Z}\times_3 \textbf{A}+\mathcal{B},
		\end{aligned}
	\end{equation*}
	and prove it is a norm.	Now we verify that $\Vert \mathcal{M} \Vert_{\diamond}$ satisfies the three properties of a norm.
	
	(i) Nonnegativity. For any $\mathcal{L}\in\mathbb{R}^{M\times N \times L_1}$, we consider two scenarios:
	\begin{itemize}
		\item if $\mathcal{L}\times_3\textbf{D}\notin \mathbb{X}_{s}$, it implies $rank(\textbf{D}\cdot unfold_3(\mathcal{L}))>s$. Then we have $\mathcal{H}_{spa}(\mathcal{L}\times_3\textbf{D})\neq 0$ since  the dimension of the null space of $\mathcal{H}_{spa}$ is $s$. 
		
		\item if $\mathcal{L}\times_3\textbf{D}\in \mathbb{X}_{s}$.  Considering $\mathcal{H}_{spa}$ satisfies the REC(s) in $\mathbb{X}_{s}$,  we have $\mathcal{H}_{spa}(\mathcal{L}\times_3\textbf{D})\neq 0$ for any  $(\mathcal{L}\times_3 \textbf{D})\neq 0$.
	\end{itemize}
	Therefore, $\mathcal{H}_{spa}(\mathcal{L}\times_3 \textbf{D})= 0$ if and only if $\mathcal{L}\times_3 \textbf{D}= 0$, which is also equivalent to $\mathcal{L}=0$ since $\textbf{D}$ is column full rank. Based on this, $\mathcal{M}=0$ can deduce $\mathcal{L}=0$, thus $\Vert \mathcal{M} \Vert_{\diamond}=0$.
	
	  Conversely, $\Vert \mathcal{M} \Vert_{\diamond}$=0 means $\Vert \mathcal{L} \Vert_{\text{TNN}}=0$, which is equivalent to $\mathcal{L}=0$. It is evident that $\mathcal{L}=0$ implies $\mathcal{M}=0$. Thus, the nonnegativity is established since $\Vert \mathcal{M} \Vert_{\diamond}\ge 0$ holds true.
	
	(ii) Positive Homogeneity.  For any constant $a$,
	\begin{equation*}
		\begin{aligned}
			\Vert a\mathcal{M} \Vert_{\diamond}=&\min\limits_{\mathcal{L},\textbf{A},\mathcal{B}}\enspace \Vert \mathcal{L} \Vert_{\text{TNN}}\\
			&s.t. \quad \mathcal{H}_{spa}(\mathcal{L}\times_3\textbf{D})=a\mathcal{M},  \\& \qquad\mathcal{H}_{spec} (\mathcal{L}\times_3\textbf{D})=\mathcal{Z}\times_3 \textbf{A}+\mathcal{B}.
		\end{aligned}
	\end{equation*}
	Let $\mathcal{L}^\ast:=\frac{1}{a}\mathcal{L}$, $\textbf{A}^\ast:=\frac{1}{a}\textbf{A}$ and $\mathcal{B}^\ast:=\frac{1}{a}\mathcal{B}$, then 
	\begin{equation*}
		\begin{aligned}
			\Vert a\mathcal{M} \Vert_{\diamond}=&\min\limits_{\mathcal{L},\textbf{A},\mathcal{B}}\enspace \Vert a\mathcal{L}^\ast \Vert_{\text{TNN}}\\
			&s.t. \quad \mathcal{H}_{spa}(\mathcal{L}^\ast\times_3\textbf{D})=\mathcal{M},  \\& \qquad\mathcal{H}_{spec} (\mathcal{L}^\ast\times_3\textbf{D})=\mathcal{Z}\times_3 \textbf{A}^\ast+\mathcal{B}^\ast.
		\end{aligned}
	\end{equation*}
	According to the definition of TNN, we have $ \Vert a\mathcal{L}^\ast \Vert_{\text{TNN}}=|a| \Vert \mathcal{L}^\ast \Vert_{\text{TNN}}$. Therefore, $\Vert a\mathcal{M} \Vert_{\diamond}=|a|\Vert \mathcal{M} \Vert_{\diamond}$.
	
	(iii) Triangle Inequality. 
	\begin{equation*}
		\begin{aligned}
			\Vert \mathcal{M}_1+\mathcal{M}_2 \Vert_{\diamond}=&\min\limits_{\mathcal{L},\textbf{A},\mathcal{B}}\enspace \Vert \mathcal{L} \Vert_{\text{TNN}}\\
			&s.t. \quad \mathcal{H}_{spa}(\mathcal{L}\times_3\textbf{D})=\mathcal{M}_1+\mathcal{M}_2,  \\& \qquad\mathcal{H}_{spec} (\mathcal{L}\times_3\textbf{D})=\mathcal{Z}\times_3 \textbf{A}+\mathcal{B}.
		\end{aligned}
	\end{equation*}
	Due to the ill-conditioning of the equations in the constraints for $\mathcal{M}_i$, with $i=1,2$, there exist $\mathcal{L}_i$ such that $\mathcal{H}_{spa}(\mathcal{L}_i \times_3 \textbf{D}) = \mathcal{M}_i$ for both $i=1$ and $i=2$. It is evident that there also exist $\textbf{A}_i$ and $\mathcal{B}_i$, for $i=1,2$, satisfying $\mathcal{H}_{spec}(\mathcal{L}_i \times_3 \textbf{D}) = \mathcal{Z} \times_3 \textbf{A}_i + \mathcal{B}_i$ for each $i=1$ and $i=2$. Therefore,  one has 
	\begin{equation*}
		\begin{aligned}
			\Vert \mathcal{M}_1+\mathcal{M}_2 \Vert_{\diamond}\le&\min\limits_{\mathcal{L}_i,\textbf{A}_i,\mathcal{B}_i}\enspace \Vert \mathcal{L}_1+\mathcal{L}_2 \Vert_{\text{TNN}}\\
			&s.t. \quad \mathcal{H}_{spa}(\mathcal{L}_i\times_3\textbf{D})=\mathcal{M}_i,  \\& \qquad\mathcal{H}_{spec} (\mathcal{L}_i\times_3\textbf{D})=\mathcal{Z}\times_3 \textbf{A}_i+\mathcal{B}_i\\&\qquad (i = 1,2),
			\\&\le 	\Vert \mathcal{M}_1 \Vert_{\diamond}+	\Vert \mathcal{M}_2 \Vert_{\diamond}.
		\end{aligned}
	\end{equation*}
	
	In summary, $\|\mathcal{M}\|_{\diamond}$ is indeed a norm. Thus, by defining $f(\textbf{x}) = \mathcal{Y} \circ \tau$ in Equation (\ref{P1}), the theorem's conclusion can be substantiated.
	
	\vspace{10pt}
	\noindent\textbf{\large C. The Proof of Theorem 4.4 }
	
	Prior to embarking on the proofs, we shall leverage the following lemmas.
	\begin{lemma}\cite{GNBLR}
		Concavity of $\psi (x)$ and $\psi (0) = 0$ means its subadditivite in $[0,+\infty)$, i.e., for any $x_1\ge 0$ and $x_2\ge 0$, $\psi(x_1+x_2)\le \psi(x_1)+\psi(x_2)$.
	\end{lemma}
	\begin{lemma}\cite{GNBLR}\label{L4}
		Suppose that $\mathcal{A}$ and $\mathcal{B}$ are two same size arbitrary tensors. Then, the following properties hold:
		\begin{itemize}
			\item $\Vert \mathcal{B}-\mathcal{A} \Vert_{\psi}\ge \Vert \mathcal{B} \Vert_{\psi}-\Vert \mathcal{A} \Vert_{\psi}$;
			\item $\Vert \mathcal{B}-\mathcal{A} \Vert_{l_F^\psi}\le \Vert \mathcal{B} \Vert_{l_F^\psi}+\Vert \mathcal{A} \Vert_{l_F^\psi}$.
		\end{itemize}
	\end{lemma}
	\begin{lemma}\label{Lemma2}(Generalized power-mean inequality\cite{G1})
		Let $w_1,w_2,...,w_n$ be $n$ positive numbers such that $\sum_{i=1}^{n}w_i=1$. Then for any real numbers $k, t$ such that $0<k<t<+\infty$, and for any $a_1,a_2,...,a_n\ge 0$,
		$$
		\left( \sum_{i=1}^{n}w_ia_i^k \right)^{\frac{1}{k}}\le \left( \sum_{i=1}^{n}w_ia_i^t \right)^{\frac{1}{t}}
		$$
		holds with equality if and only if $a_1=a_2=...=a_n$.
	\end{lemma}

	
	Based on the lemma above, we proceed to undertake the proof of \textbf{Theorem 4.4}.
	
	\textbf{Proof:} We begin with the first part of proving the conclusion. For any tensor $\mathcal{A}\in \mathbb{R}^{M\times N\times H}$, one has
	\begin{equation*}
		\begin{aligned}
			\Vert \mathcal{A} \Vert_{l_F^\psi}&=\sum_{i=1}^{M}\sum_{j=1}^{N}\psi(\Vert \mathcal{A}(i,j,:) \Vert_F)\\&\ge \psi\left( \sum_{i=1}^{M}\sum_{j=1}^{N}\Vert\mathcal{A}(i,j,:)\Vert_F \right)\ge \psi(\Vert \mathcal{A} \Vert_F).
		\end{aligned}
	\end{equation*}
	where the first inequality comes from the fact that $\psi$ is a subadditive function and the second inequality is based on the triangle inequality of $\Vert \cdot \Vert_F$.
	By optimality, we have
	\begin{equation*}
		\begin{aligned}
			\alpha\Vert \hat{\mathcal{L}}\Vert_{\psi}+\beta\Vert \hat{\mathcal{E}} \Vert_{l_F^\psi}\le \alpha\Vert \mathcal{L}^\star\Vert_{\psi}+\beta\Vert \mathcal{E}^\star \Vert_{l_F^\psi},
		\end{aligned}
	\end{equation*}
	thereby leading to
	\begin{equation*}
		\begin{aligned}
			\beta\Vert \hat{\mathcal{E}} \Vert_{l_F^\psi}\le \alpha(\Vert \mathcal{L}^\star\Vert_{\psi}-\Vert \hat{\mathcal{L}}\Vert_{\psi})+\beta\Vert \mathcal{E}^\star \Vert_{l_F^\psi}.
		\end{aligned}
	\end{equation*}
	From Lemma \ref{L4}, we obtain
	\begin{equation*}
		\begin{aligned}
			\beta \Vert \hat{\mathcal{E}}-\mathcal{E}^\star \Vert_{l_F^\psi}&\le
			\beta\Vert \hat{\mathcal{E}} \Vert_{l_F^\psi}+\beta\Vert \mathcal{E}^\star \Vert_{l_F^\psi}
			\\&\le \alpha(\Vert \mathcal{L}^\star\Vert_{\psi}-\Vert \hat{\mathcal{L}}\Vert_{\psi})+2\beta\Vert \mathcal{E}^\star \Vert_{l_F^\psi}
			\\&\le\alpha\Vert \mathcal{L}^\star-\hat{\mathcal{L}}\Vert_{\psi}+2\beta\Vert \mathcal{E}^\star \Vert_{l_F^\psi}.
		\end{aligned}
	\end{equation*}
	Let $\mathcal{L}= \mathcal{U}*\mathcal{D}*\mathcal{V}^H$, and $\sigma_k^l:=\sigma_l(\bar{\textbf{D}}^{(k)})$. In addition, we can deduce that for any tensor $\mathcal{L}\in \mathbb{R}^{M\times N\times L_1}$,
	\begin{equation}\tag{4.1}\label{B5}
		\begin{aligned}
			\Vert \mathcal{L}\Vert_{\psi}&=\sum_{k=1}^{L_1}\sum_{l=1}^{r}\frac{1}{L_1}\psi(\sigma_k^l)\le r\psi \left( \sum_{k=1}^{L_1}\sum_{l=1}^{r}\frac{1}{L_1r}\sigma_k^l \right)\\&\le r\psi\left(\sqrt{\frac{1}{L_1r}\sum_{k=1}^{L_1}\sum_{l=1}^{r}(\sigma_k^l)^2}\right)\\&\le r\psi(r^{-\frac{1}{2}}\Vert \mathcal{D} \Vert_F)=r\psi(r^{-\frac{1}{2}}\Vert \mathcal{L} \Vert_F).
		\end{aligned}
	\end{equation}
	where the first inequality utilizes the concavity of function $\psi$, while the second inequality makes use of the non-decreasing property of function $\psi$ and Lemma \ref{Lemma2} with $k=1$ and $t=2$.
	According to inequality (\ref{B5}), we have
	\begin{equation} \tag{4.2}\label{B22}
		\begin{aligned}
			\Vert \mathcal{L}^\star-\hat{\mathcal{L}}\Vert_{\psi}\le r\psi(r^{-\frac{1}{2}}\Vert \mathcal{L}^\star-\hat{\mathcal{L}} \Vert_F)=s.
		\end{aligned}
	\end{equation}
    Considering the constraint conditions of the NLRGS model, one has
	\begin{equation*}
		\begin{aligned}
			&\mathcal{H}_{spec} (\hat{\mathcal{L}}\times_3\textbf{D}_{\mathcal{L}}+\hat{\mathcal{E}}\times_3\textbf{D}_{\mathcal{E}})=\mathcal{Z},\\
			&\mathcal{H}_{spec} (\mathcal{L}^\star\times_3\textbf{D}_{\mathcal{L}}+\mathcal{E}^\star\times_3\textbf{D}_{\mathcal{E}})=\mathcal{Z}.
		\end{aligned}
	\end{equation*}
	This implies that
	\begin{equation}\tag{4.3}\label{B21}
		\begin{aligned}
			\Vert\mathcal{H}_{spec} ((\mathcal{E}^\star-\hat{\mathcal{E}})\times_3\textbf{D}_{\mathcal{E}})\Vert_F&=\Vert\mathcal{H}_{spec} ((\hat{\mathcal{L}}-\mathcal{L}^\star)\times_3\textbf{D}_{\mathcal{L}})\Vert_F\\&\ge \xi\Vert \hat{\mathcal{L}}-\mathcal{L}^\star\Vert_F
		\end{aligned}
	\end{equation}
	where the inequality comes from $\mathcal{H}_{spec}$ satisfies the REC($s$) with $\Vert (\mathcal{L}^\star-\hat{\mathcal{L}})\times_3\textbf{D}_{\mathcal{L}} \Vert_{\psi}\le r\psi(r^{-\frac{1}{2}}\Vert (\mathcal{L}^\star-\hat{\mathcal{L}})\times_3\textbf{D}_{\mathcal{L}} \Vert_F)=s$, which means $(\mathcal{L}^\star-\hat{\mathcal{L}})\times_3\textbf{D}_{\mathcal{L}}\in\mathbb{X}_{s}$. 
	Then we get
	\begin{equation}\tag{4.4}\label{B33}
		\begin{aligned}
			\Vert \mathcal{L}^\star-\hat{\mathcal{L}} \Vert_F\le\frac{1}{\xi}\Vert\mathcal{H}_{spec} ((\mathcal{E}^\star-\hat{\mathcal{E}})\times_3\textbf{D}_{\mathcal{E}})\Vert_F.
		\end{aligned}
	\end{equation}
	Combining inequality (\ref{B33}) with (\ref{B22}), one has
	\begin{equation*}
		\begin{aligned}
			\Vert \mathcal{L}^\star&-\hat{\mathcal{L}}\Vert_{\psi}	\le r\psi(r^{-\frac{1}{2}}\Vert \mathcal{L}^\star-\hat{\mathcal{L}} \Vert_F)\\&\le  r\psi(\frac{1}{\sqrt{r}\xi}(\Vert\mathcal{H}_{spec} ((\mathcal{E}^\star-\hat{\mathcal{E}})\times_3\textbf{D}_{\mathcal{E}})\Vert_F) \\&\le r\psi\left(\frac{\Vert\textbf{RD}_{\mathcal{E}}\Vert_F}{\sqrt{r}\xi}\Vert\hat{\mathcal{E}}-\mathcal{E}^\star\Vert_F\right)\\&\le
			rv\psi(\Vert \hat{\mathcal{E}}-\mathcal{E}^\star\Vert_F),
		\end{aligned}
	\end{equation*}
	where the last inequality holds since $\psi$ is taken to be the MCP  function, which is bounded by the $l_1$ norm.  $v$ is a positive constant related to $r,s, \psi$ and $\textbf{R}$.
	Therefore, we obtain
	\begin{equation*}
		\begin{aligned}
			\beta \Vert \hat{\mathcal{E}}-\mathcal{E}^\star \Vert_{l_F^\psi}
			&\le\alpha\Vert \mathcal{L}^\star-\hat{\mathcal{L}}\Vert_{\psi}+2\beta\Vert \mathcal{E}^\star \Vert_{l_F^\psi}\\&\le \alpha rv\psi(\Vert \hat{\mathcal{E}}-\mathcal{E}^\star\Vert_F)+2\beta\Vert \mathcal{E}^\star \Vert_{l_F^\psi}\\&\le \alpha rv\Vert \hat{\mathcal{E}}-\mathcal{E}^\star \Vert_{l_F^\psi}+2\beta\Vert \mathcal{E}^\star \Vert_{l_F^\psi}.
		\end{aligned}
	\end{equation*}
	Finally, provided that $\beta>\alpha r v$, we get
	\begin{equation}\tag{4.5}
		\begin{aligned}
			\psi(\Vert \hat{\mathcal{E}}-\mathcal{E}^\star\Vert_F)\le\Vert \hat{\mathcal{E}}-\mathcal{E}^\star \Vert_{l_F^\psi}
			\le\frac{2\beta\Vert \mathcal{E}^\star \Vert_{l_F^\psi}}{\beta-\alpha rv}.
		\end{aligned}
	\end{equation}
	Thus far, the proof of the first theorem has been completed. Next, we proceed to establish the proof for the second part of the conclusion.
	Let $\mathcal{F}:=\mathcal{L}^\star-\hat{\mathcal{L}}$ and $h:=MN$, according to the definition of $\Vert\cdot\Vert_{l_F^\psi}$, we have
	\begin{equation*}
		\begin{aligned}
			\Vert \mathcal{L}^\star - \hat{\mathcal{L}} \Vert_{l_F^\psi}&=\sum_{i=1}^{M}\sum_{j=1}^{N}\psi(\Vert \mathcal{F}(i,j,:) \Vert_F)
			\\&= h\sum_{i=1}^{M}\sum_{j=1}^{N}\frac{1}{h}\psi(\Vert \mathcal{F}(i,j,:) \Vert_F)
			\\&\le h\psi\left( \sum_{i=1}^{M}\sum_{j=1}^{N}\frac{1}{h}\Vert \mathcal{F}(i,j,:) \Vert_F\right)
			\\&\le h\psi\left(\sqrt{\dfrac{\sum_{i=1}^{M}\sum_{j=1}^{N}\Vert \mathcal{F}(i,j,:) \Vert_F^2}{h} }\right)
			\\&=h\psi(h^{-\frac{1}{2}}\Vert \mathcal{L}^\star - \hat{\mathcal{L}} \Vert_F).
		\end{aligned}
	\end{equation*}
	where the first inequality utilizes the concavity and the second inequality follows from Lemma \ref{Lemma2} with $k=1,t=2$. 
	Then from the inequality \ref{B21}, we obtain
	\begin{equation*}
		\begin{aligned}
			\Vert \mathcal{L}^\star - \hat{\mathcal{L}} \Vert_{l_F^\psi}&\le h\psi(h^{-\frac{1}{2}}\Vert \mathcal{L}^\star - \hat{\mathcal{L}} \Vert_F)\\&\le h\psi\left(h^{-\frac{1}{2}}\dfrac{\Vert\mathcal{H}_{spec} ((\hat{\mathcal{E}}-\mathcal{E}^\star)\times_3\textbf{D}_{\mathcal{E}})\Vert_F}{\xi}\right)
			\\&\le h\psi\left( \dfrac{\Vert \textbf{RD}_{\mathcal{E}} \Vert_F}{\sqrt{h}\xi}\Vert  \mathcal{E}^\star - \hat{\mathcal{E}} \Vert_F \right)
			\\&\le hw\psi(\Vert  \mathcal{E}^\star - \hat{\mathcal{E}} \Vert_F)
			\\&\le \frac{2hw\beta\Vert \mathcal{E}^\star \Vert_{l_F^\psi}}{\beta-\alpha rv}.
		\end{aligned}
	\end{equation*}
	where  $w$ is a positive constant related to $h,s, \psi$ and $\textbf{R}$. Similar to our initial conclusion, given the condition $\beta>\alpha r v$, we arrive at
	\begin{equation*}
		\begin{aligned}
			\Vert \mathcal{L}^\star - \hat{\mathcal{L}} \Vert_{l_F^\psi}\le hw\psi(\Vert  \mathcal{E}^\star - \hat{\mathcal{E}} \Vert_F)
			\le \frac{2hw\beta\Vert \mathcal{E}^\star \Vert_{l_F^\psi}}{\beta-\alpha rv}.
		\end{aligned}
	\end{equation*}
	We conclude the proof by $\psi(\Vert \mathcal{L}^\star - \hat{\mathcal{L}} \Vert_F)\le \Vert \mathcal{L}^\star - \hat{\mathcal{L}} \Vert_{l_F^\psi}$.

    \vspace{10pt}
	\noindent\textbf{\large D. The Proof of Theorem 4.5 }
	
	In this section, we establish the global convergence of Algorithm 4 by utilizing the KL property \cite{Attouch1,KL31}. 

 \textbf{Proof:} Considering that we concatenate the individual blocks back together after solving the subproblem $\mathcal{L}$, we do not take the clustering process into account in our convergence analysis. We now utilize the KL property and establish the fulfillment of the H1 condition of this algorithm. To simplify matters, we  denote:
	\begin{equation*}\label{E4}
		\begin{aligned}
			q(\mathcal{L},\mathcal{E})=\Vert \mathcal{H}_{spa}(\mathcal{L}\times_3\textbf{D}_{\mathcal{L}}+\mathcal{E}\times_3\textbf{D}_{\mathcal{E}})-\mathcal{Y}_R \Vert_F^2+\Vert \mathcal{H}_{spec} (\mathcal{L}\times_3\textbf{D}_{\mathcal{L}}+\mathcal{E}\times_3\textbf{D}_{\mathcal{E}})-\mathcal{Z} \Vert_F^2,
		\end{aligned}
	\end{equation*}
	and then 
	\begin{equation*}\label{E5}
		\begin{aligned}
			g(\mathcal{L},\mathcal{E})=q(\mathcal{L},\mathcal{E})+\alpha\Vert \mathcal{L} \Vert_{\psi}+\beta\Vert \mathcal{E} \Vert_{l_F^\psi},
		\end{aligned}
	\end{equation*}
	which is the objective function of model (28). From (29) and (30), we have
	\begin{equation}\tag{5.1}\label{E6}
		\begin{aligned}
			q(\mathcal{L}^{k+1},\mathcal{E}^k)+\alpha\Vert \mathcal{L}^{k+1} \Vert_{\psi}+\dfrac{\mu}{2}\Vert \mathcal{L}^{k+1}-\mathcal{L}^k \Vert_F^2\leq q(\mathcal{L}^{k},\mathcal{E}^k)+\alpha\Vert \mathcal{L}^{k} \Vert_{\psi},
		\end{aligned}
	\end{equation}
	\begin{equation}\tag{5.2}\label{E7}
		\begin{aligned}
			q(\mathcal{L}^{k+1},\mathcal{E}^{k+1})
			+\beta\Vert \mathcal{E}^{k+1} \Vert_{l_F^\psi}+\dfrac{\mu}{2}\Vert \mathcal{E}^{k+1}-\mathcal{E}^k \Vert_F^2\le q(\mathcal{L}^{k+1},\mathcal{E}^{k})+\beta\Vert \mathcal{E}^{k} \Vert_{l_F^\psi}.
		\end{aligned}
	\end{equation}
	By combining equation (\ref{E6}) and equation (\ref{E7}), we obtain
	\begin{equation}\tag{5.3}\label{E8}
		\begin{aligned}
			g(\mathcal{L}^{k+1},\mathcal{E}^{k+1})+\dfrac{\mu}{2}(\Vert\mathcal{L}^{k+1}-\mathcal{L}^k \Vert_F^2+\Vert \mathcal{E}^{k+1}-\mathcal{E}^k \Vert_F^2)\le g(\mathcal{L}^{k},\mathcal{E}^k).
		\end{aligned}
	\end{equation}
	Henceforth, we have effectively showcased the sufficient descent condition, namely, the H1 condition.
	Then, let $N$ be a positive integer. Summing (\ref{E8}) from $k=0$ to $N-1$, we obtain
	\begin{equation*}
		\begin{aligned}
			\sum\limits_{k=0}^{N-1}(\Vert \mathcal{L}^{k+1}-\mathcal{L}^k \Vert^2+\Vert \mathcal{E}^{k+1}-\mathcal{E}^k \Vert^2)=\sum\limits_{k=0}^{N-1} \Vert \mathcal{V}^{k+1}-\mathcal{V}^k \Vert^2\le \frac{2}{\mu}(g(\mathcal{V}^{0})-g(\mathcal{V}^{N})).
		\end{aligned}
	\end{equation*}
	By the inequality (\ref{E8}), we  get the sequence $g(\mathcal{V}^{k})$ is non-increasing. Since $g(\mathcal{V})$ is bound below, the sequence $\{g(\mathcal{V}^{k})\}$ converges. Taking the limits as $N\to +\infty$, we get $\sum\limits_{k=0}^{+\infty} \Vert \mathcal{V}^{k+1}-\mathcal{V}^k \Vert^2\textless +\infty $ and  ${\rm lim}_{k\to +\infty}\Vert \mathcal{V}^{k+1}-\mathcal{V}^k \Vert=0$.
	
	Advancing in our discourse, we now pivot to demonstrating the fulfillment of the H2 condition.  Considering the optimal conditions of (29) and (30), we can deduce that
	\begin{equation*}\label{E9}
		\begin{aligned}
			&\nabla_\mathcal{L}q(\mathcal{L}^{k+1},\mathcal{E}^k)+\mathcal{U}^{k+1}+\mu(\mathcal{L}^{k+1}-\mathcal{L}^k)=0,\\
			&\nabla_\mathcal{E}q(\mathcal{L}^{k+1},\mathcal{E}^{k+1})+\mathcal{V}^{k+1}+\mu(\mathcal{E}^{k+1}-\mathcal{E}^k)=0.
		\end{aligned}
	\end{equation*}	
	for some $\mathcal{U}^{k+1}\in \partial( \alpha\Vert \mathcal{L} \Vert_{\psi})$ and $\mathcal{V}^{k+1}\in \partial(\beta\Vert \mathcal{E}^{k+1} \Vert_{l_F^\psi})$, where `$\partial$' represents the subgradient of a function. On the other hand, from [Exercise 8.8 (c), \cite{VA1}], we obtain
	\begin{equation*}\label{E10}
		\begin{aligned}
			&\partial_\mathcal{L}f(\mathcal{L},\mathcal{E})=\nabla_\mathcal{L}q(\mathcal{L},\mathcal{E})+\partial(\alpha\Vert \mathcal{L} \Vert_{\psi}),\\
			&\partial_\mathcal{E} f(\mathcal{L},\mathcal{E})=\nabla_\mathcal{E}q(\mathcal{L},\mathcal{E})+\partial(\beta\Vert \mathcal{E}^{k+1} \Vert_{l_F^\psi}).
		\end{aligned}
	\end{equation*}
	Therefore, we have the following equation: 
	\begin{equation*}\label{E11}
		\begin{aligned}
			\nabla_\mathcal{L}q&(\mathcal{L}^{k+1},\mathcal{E}^{k+1})-\nabla_\mathcal{L}q(\mathcal{L}^{k+1},\mathcal{E}^{k})+\mu(\mathcal{L}^{k}-\mathcal{L}^{k+1})\\&=\nabla_\mathcal{L}q(\mathcal{L}^{k+1},\mathcal{E}^{k+1})+\mathcal{U}^{k+1}\in\partial_\mathcal{L}f(\mathcal{L}^{k+1},\mathcal{E}^{k+1}),\\
			\mu(\mathcal{E}^k&-\mathcal{E}^{k+1})=\nabla_\mathcal{E}q(\mathcal{L}^{k+1},\mathcal{E}^{k+1})+\mathcal{V}^{k+1}\in\partial_\mathcal{E} f(\mathcal{L}^{k+1},\mathcal{E}^{k+1}),
		\end{aligned}
	\end{equation*}
	where
	\begin{equation*}\label{E12}
		\begin{aligned}
			&\nabla_\mathcal{L}q(\mathcal{L}^{k+1},\mathcal{E}^{k+1})-\nabla_\mathcal{L}q(\mathcal{L}^{k+1},\mathcal{E}^{k})\\&=2\textbf{D}_\mathcal{L}^\top\textbf{D}_\mathcal{E}(\textbf{E}_{(3)}^{k+1}-\textbf{E}_{(3)}^k)(\textbf{BS})^\top+2\tau(\textbf{RD}_\mathcal{L})^\top \textbf{D}_\mathcal{E}(\textbf{E}_{(3)}^{k+1}-\textbf{E}_{(3)}^k).
		\end{aligned}
	\end{equation*}
	Note that $\textbf{D}_\mathcal{L},\textbf{D}_\mathcal{E},\textbf{B},\textbf{S},\textbf{R}$ are all predetermined, this leads to the conclusion that
	\begin{equation*}\label{E13}
		\begin{aligned}
			&\Vert\nabla_\mathcal{L}q(\mathcal{L}^{k+1},\mathcal{E}^{k+1})-\nabla_\mathcal{L}q(\mathcal{L}^{k+1},\mathcal{E}^k)\Vert\\\le&
			2\Vert \textbf{D}_\mathcal{L}^\top \textbf{D}_\mathcal{E}\Vert\Vert \textbf{E}_{(3)}^{k+1}-\textbf{E}_{(3)}^k\Vert\Vert \textbf{BS}\Vert+2\tau \Vert (\textbf{RD}_\mathcal{L})^\top \textbf{D}_\mathcal{E} \Vert \Vert \textbf{E}_{(3)}^{k+1}-\textbf{E}_{(3)}^k\Vert\\\le& m\Vert \mathcal{E}^{k+1}-\mathcal{E}^k\Vert,
		\end{aligned}
	\end{equation*}
	for some constant $m\geq2\Vert \textbf{D}_\mathcal{L}^\top \textbf{D}_\mathcal{E}\Vert\Vert \textbf{BS}\Vert+2\tau \Vert (\textbf{RD}_\mathcal{L})^\top \textbf{D}_\mathcal{E} \Vert$. Now we define $$
	\mathcal{B}_\mathcal{L}:=\nabla_\mathcal{L}q(\mathcal{L}^{k+1},\mathcal{E}^{k+1})-\nabla_\mathcal{L}q(\mathcal{L}^{k+1},\mathcal{E}^{k})+\mu(\mathcal{L}^{k}-\mathcal{L}^{k+1})
	$$ 
	and 
	$$
	\mathcal{B}_\mathcal{E}:=\mu(\mathcal{E}^k-\mathcal{E}^{k+1}),
	$$
	then from [Proposition 2.1, \cite{KL31}], one has $\textbf{d}:=(\mathcal{B}_\mathcal{L},\mathcal{B}_\mathcal{E})\in \partial f(\mathcal{L}^{k+1},\mathcal{E}^{k+1})$
	and $\textbf{d}$ satisfies $$\Vert \textbf{d} \Vert\le (m+2\mu)(\Vert \mathcal{L}^{k+1}-\mathcal{L}^k\Vert+\Vert \mathcal{E}^{k+1}-\mathcal{E}^k\Vert).$$ Consequently, we have successfully established the validity of the H2 condition.
	
	Next, we will prove that the function $g(\mathcal{L},\mathcal{E})$ is a KL function. To begin, we observe that the Frobenius norm $\|\cdot\|_{F}$ and the MCP function $\psi$ have been shown to be semi-algebraic in \cite{Bochnak1} and \cite{YangL1}, respectively. Additionally, it has been established in [Lemma 4, \cite{Zhao11}] that the tensor nuclear norm $\|\cdot\|_{\text{TNN}}$ is also semi-algebraic. 
	$\Vert \cdot \Vert_{\psi}$ is the composite function of $\psi$ and $\|\cdot\|_{\text{TNN}}$. Therefore, $\Vert \cdot \Vert_{\psi}$ is also  semi-algebraic.  In a similar way, $\Vert \cdot \Vert_{l_F^\psi}$ is semi-algebraic.
	Then, we can conclude that the function $g(\mathcal{L},\mathcal{E})$ is semi-algebraic as it can be expressed as a finite sum of semi-algebraic functions.	
	Furthermore, since $g(\mathcal{L},\mathcal{E})$ is proper and lower semi-continuous, we can apply [Theorem 3, \cite{Bolte}] to show that $g$ is a KL function. Additionally, since the bounded sequence $\{\mathcal{V}^{k}\}_{k\in\mathbb{N}}$ admits a converging subsequence, we  conclude that continuity condition is trivially fulfilled because of the continuity of $g$.  Consequently, by [Theorem 2.9, \cite{Attouch1}], the sequence $\{\mathcal{V}^k\}_{k\in\mathbb{N}}$ produced by Algorithm 4 converges to a critical point of function $g$,
	which completes the proof. \qed

 \newpage
	\noindent\textbf{\large E. Experimental results of DFMF on 'Face' }
	
	\vspace{-10pt}
	\begin{table}[htbp]\small
		\renewcommand\arraystretch{1.3}
		\begin{center}
			\caption{ The experimental results  on 'Face'.}
			\begin{tabular}{@{}*{6}{l}}
				\hline
				Dataset& Method & PSNR   &SSIM &ERGAS &SAM  \\
				\hline
				\multirow{2}{*}{Face}  & DFMF \cite{DFMF1} & 36.44  & 0.934 & 2.472 & 16.635   \\
				\cline{2-6}
				& RAF-NLRGS  & 42.46 &  0.977  & 1.570 & 15.413 \\		
				\hline
			\end{tabular}
		\end{center}
	\end{table} 
	\vspace{-20pt}
	\begin{figure}[htbp]
		\centering
		\subfloat{\includegraphics[width=2.8cm]{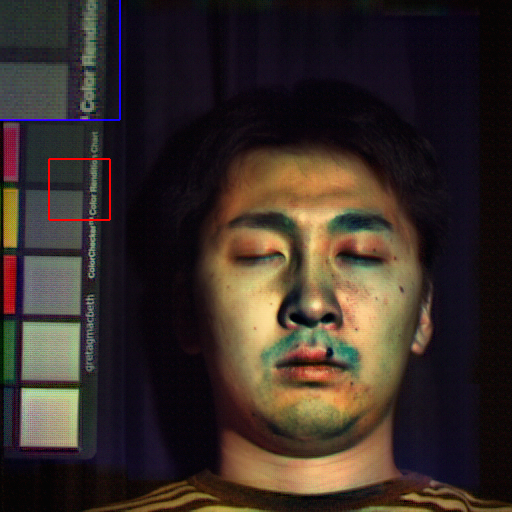}}~\!
		\subfloat{\includegraphics[width=2.8cm]{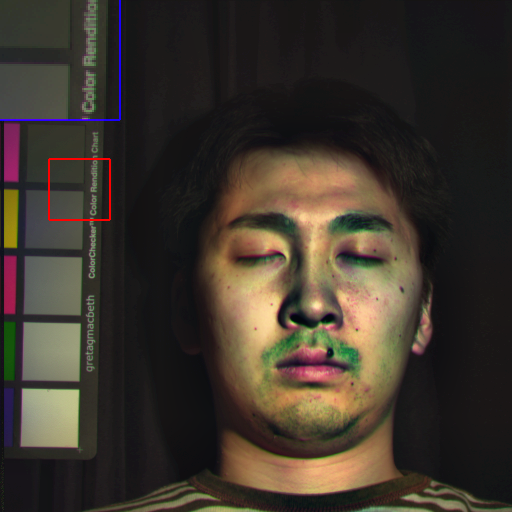}}~\!
		\subfloat{\includegraphics[width=2.8cm]{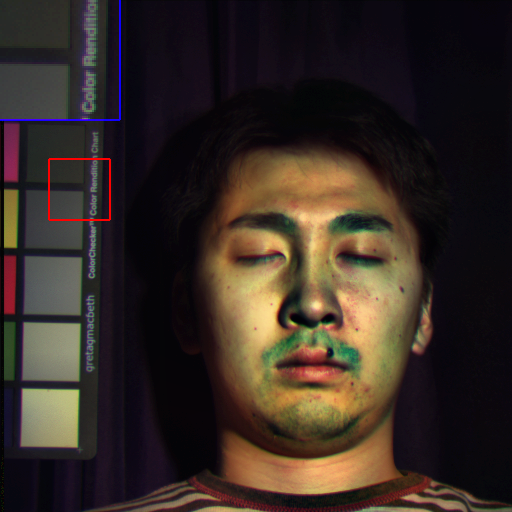}}
		
		\vspace{-5pt}
		\subfloat[DFMF]{\includegraphics[width=2.8cm]{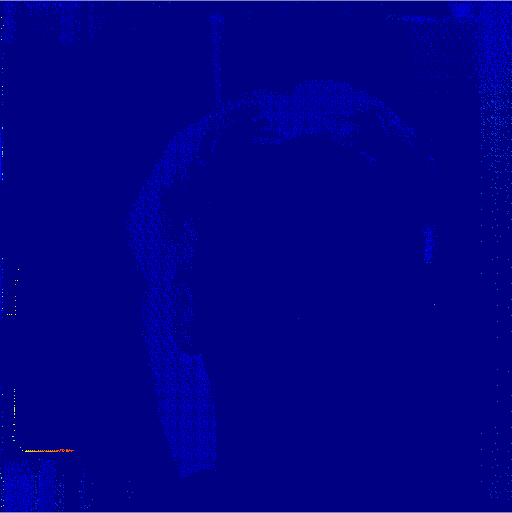}}~\!
		\subfloat[RAF-NLRGS]{\includegraphics[width=2.8cm]{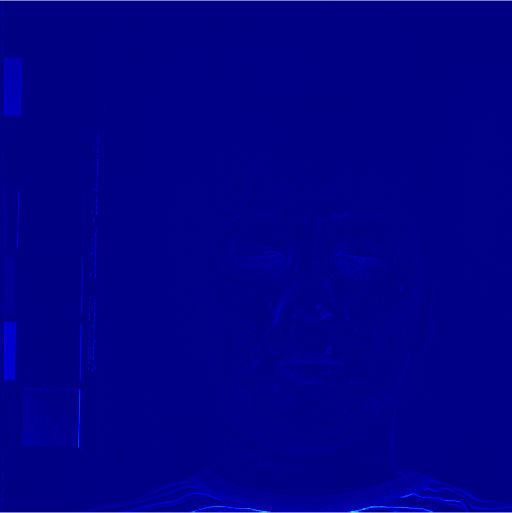}}~\!
		\subfloat[GT]{\includegraphics[width=2.8cm]{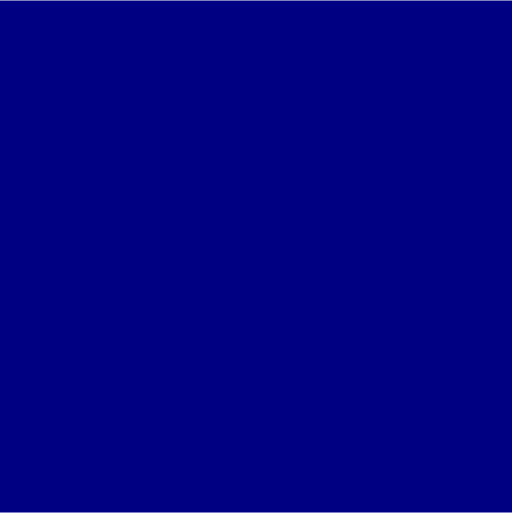}}
		
		\vspace{-5pt}
		\subfloat{\includegraphics[width=7.5cm]{overleaf_RAF/band_cave.png}}
		
		\vspace{-3pt}
		\caption{The reconstructed `Face' (bands 10, 20, and 30) and corresponding error images resulting from compared methods.\tiny\label{DF5}}
	\end{figure}

\end{document}